\documentclass[times,twocolumn,final]{elsarticle}  
\DeclareUnicodeCharacter{FF0C}{,}  

\usepackage{medima}
\usepackage{framed,multirow}
\usepackage{enumitem}
\usepackage{longtable}
\usepackage{tabularx}

\usepackage{amssymb}
\usepackage{latexsym}

\usepackage{url}
\usepackage{xcolor}
\usepackage{adjustbox}

\usepackage{cite}
\usepackage{amsmath,amssymb,amsfonts}
\usepackage{graphicx}
\usepackage{textcomp}

\usepackage{algorithmicx}               
\usepackage{algorithm} 
\usepackage{listings}
\usepackage{algpseudocode}  

\usepackage{array}
\usepackage{booktabs}
\usepackage{colortbl}
\definecolor{dark-gray}{gray}{0.7} 

\usepackage{graphicx, caption}
\usepackage{color,soul}
\usepackage{rotating}

\usepackage{pdfpages}  
\usepackage{hyperref}  
\newcommand{\gain}[1]{\textcolor{green!50!black}{#1}}
\newcommand{\dropval}[1]{\textcolor{red!70!black}{#1}}

\definecolor{newcolor}{rgb}{.8,.349,.1}

\journal{Medical Image Analysis}

\renewcommand{\arraystretch}{0.9} 

\begin{document}
\verso{Cheng \textit{et~al.}}

\begin{frontmatter}
\title{PathLang: A Language-Centered Benchmark for Vision--Language Models in Computational Pathology}%

\author[1]{Fanqi \snm{Cheng}}
\author[2]{Kuo \snm{Gong}}
\author[3,9]{Shangke \snm{Liu}}
\author[4]{Beidi \snm{Zhao}}
\author[6]{Junchao \snm{Zhu}}
\author[7]{Zheyu \snm{Zhu}}
\author[8]{Leiyue \snm{Zhao}}
\author[9]{Fengbei \snm{Liu}}
\author[10]{John \snm{Cannon}}
\author[4,5]{Gang \snm{Wang}}
\author[4,5]{Zu-hua \snm{Gao}}
\author[3]{Kenji \snm{Ikemura}}
\author[3]{Yihe \snm{Yang}}
\author[11]{Yaohong \snm{Wang}}
\author[6]{Yuankai \snm{Huo}}
\author[4]{Xiaoxiao \snm{Li}}
\author[3,12]{Mert R. \snm{Sabuncu}}
\author[3]{Ruining \snm{Deng}\corref{cor1}}
\ead{rud4004@med.cornell.edu}
\cortext[cor1]{Corresponding author.}

\address[1]{New York University, New York, NY, USA}
\address[2]{Columbia University, New York, NY, USA}
\address[3]{Weill Cornell Medicine, New York, NY, USA}
\address[4]{University of British Columbia, Vancouver, BC, Canada}
\address[5]{BC Cancer Agency, Vancouver, BC, Canada}
\address[6]{Vanderbilt University, Nashville, TN, USA}
\address[7]{University of Pennsylvania, Philadelphia, PA, USA}
\address[8]{Johns Hopkins University, Baltimore, MD, USA}
\address[9]{Cornell University, Ithaca, NY, USA}
\address[10]{New York Medical College, New York, NY, USA}
\address[11]{The University of Texas MD Anderson Cancer Center, Houston, TX, USA}
\address[12]{Cornell Tech, New York, NY, USA}

\received{xxxxxx}
\finalform{xxxxxx}
\accepted{xxxxxx}
\availableonline{xxxxxx}
\communicated{xxxxxx}

\begin{abstract}

Pathology vision–language models (VLMs) have shown strong visual perception ability, but their robustness in the language domain remains poorly characterized. Existing pathology VLM benchmarks largely rely on canonical closed-set prompts or perturb only generic templates, treating language as a fixed evaluation component rather than a variable axis of model behavior. In real-world clinical use, however, diagnostic language varies across reports, institutions, and candidate diagnoses, especially in fine-grained subtyping, cross-organ differentiation, and differential diagnosis. We introduce \textbf{PathLang, a language-centered and clinically grounded zero-shot benchmark.} PathLang is language-centered: it holds the underlying slides, ground-truth labels, and image–text evaluation direction fixed while systematically varying only the diagnostic language, so that performance differences reflect how a diagnosis is phrased rather than what is imaged. It is clinically grounded: the language variation follows how pathologists actually rephrase diagnoses (terminology, specificity, and reporting style), and all prompts and candidate pools are validated by six board-certified pathologists. PathLang evaluates models across four task families: (1) zero-shot image classification with image–text alignment analysis, (2) cross-modal retrieval, (3) paraphrase robustness under semantic-equivalence paraphrases, length and reporting-style variation, and prompt ensembling, and (4) open-vocabulary diagnosis retrieval over four candidate pools that impose distinct forms of semantic competition. Across nine VLMs and five public pathology datasets spanning four organs, we find that performance is highly sensitive to clinically equivalent paraphrases, varies substantially across forms of semantic competition, and that image–text alignment quality does not necessarily translate into better inter-class separability, failure modes that canonical-prompt evaluation does not reveal. We release the prompt corpus, candidate pools, pre-computed text embeddings for all nine VLMs, and evaluation code to support reproducible evaluation and the development of more semantically robust pathology VLMs. Code and benchmark resources are available at: https://anonymous.4open.science/r/PathLang, and will be released publicly upon acceptance.

\vspace{-1em}
\end{abstract}

\end{frontmatter}


\section{Introduction}

Pathology vision--language models (VLMs) align histopathology images with paired textual descriptions during pretraining and have rapidly become a foundation for zero-shot diagnosis, cross-modal retrieval, and prompt-driven querying in computational pathology~\citep{conch, plip, musk, biomedclip, li2025multi, dai2025pathologyvlm, imran2026multimodal}. Because they obviate the cost of task-specific labels, these models are increasingly proposed for clinical workflows including fine-grained disease subtyping, cross-organ differentiation, and report-level differential diagnosis~\citep{conch, musk, pathgenclip}. However, evaluation has not kept pace with capability along the language axis on two levels. First, most existing pathology VLM benchmarks treat the language modality as fixed: they measure visual robustness, slide-level aggregation, and closed-set classification accuracy under a small set of hand-curated prompts~\citep{thunder, mizero}, effectively reducing the VLM to an image classifier. Second, the few benchmarks that probe textual variation, such as Histo-VL~\citep{histovl}, focus on coarse template-level changes (e.g., adding ``a histopathology image of'' before the class name) rather than the diagnostic language pathologists actually vary in clinical practice. As a result, the behavior of these models under changes to the \emph{diagnostic language itself} remains largely uncharacterized, even when the underlying image and label do not change.

This second-level gap is clinically consequential. Pathologists routinely describe the same finding along three axes of variation: \textbf{terminology} (\emph{invasive ductal carcinoma} vs.\ \emph{infiltrating ductal carcinoma} vs.\ \emph{ductal carcinoma, not otherwise specified}), \textbf{specificity} (\emph{Gleason 4} vs.\ \emph{Gleason pattern 4 prostate cancer} vs.\ \emph{moderately differentiated adenocarcinoma with cribriform glands}), and \textbf{reporting style} (keyword-style sign-outs vs.\ extended narrative descriptions). Unlike template-level prompt engineering studied in general-domain VLMs~\citep{clip, coop, cocoop}, here the diagnostic concept itself is rephrased, abbreviated, or reformulated. A model that succeeds under one canonical prompt but fails under a clinically equivalent paraphrase is unreliable for any deployment setting in which clinicians query in natural language.

\begin{figure*}[!t]
    \centering
    \centerline{\includegraphics[width=1.0\textwidth]{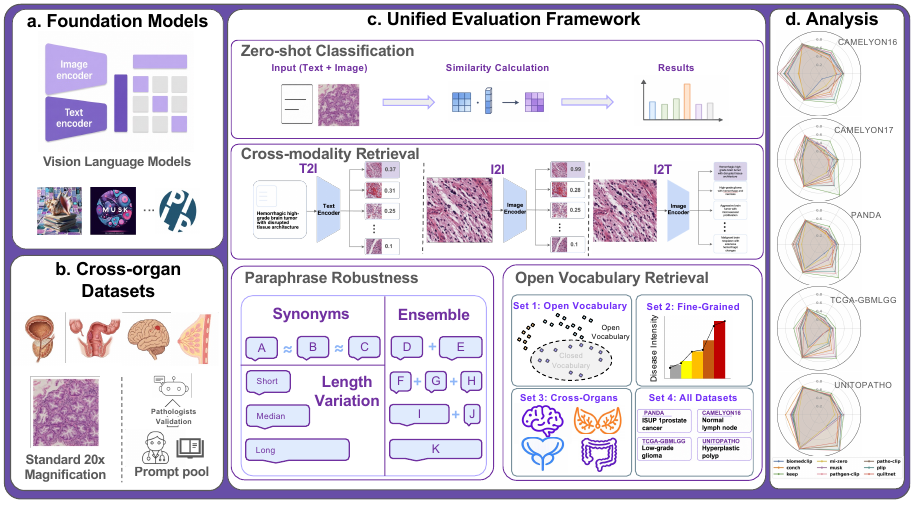}}
    \caption{\textbf{Overall framework of PathLang.} a. We focus on dual-encoder pathology foundation models; b. We use five datasets including WSIs (20$\times$) from four organs and the pathologist validated prompt pool; c. Our evaluation contains four tasks: zero-shot classification, cross-modal retrieval, paraphrase robustness and open vocabulary retrieval. d. We present a variety of results.
    }
    \label{fig:overview}
\end{figure*}
 

We propose \textbf{PathLang}, a language-centered, clinically grounded zero-shot benchmark that treats clinical diagnostic language as a controlled experimental variable. We use these two terms in a specific sense. Language-centered refers to the experimental design. Existing benchmarks vary images, datasets, or aggregation while holding the prompt fixed. PathLang inverts this: it holds the slides, labels, and image–text evaluation direction fixed and varies only the diagnostic text. Performance differences across conditions can therefore be attributed to language-side behavior rather than to slide selection, label composition, or the modality gap that separates retrieval directions. Clinically grounded refers to the content of that variation. Rather than editing generic carrier templates (e.g., prepending "a histopathology image of"), PathLang rephrases the diagnostic concept itself along the three axes pathologists vary in practice (terminology, specificity, and reporting style) while preserving its diagnostic meaning. All prompts and candidate pools were drafted and validated by six board-certified pathologists from multiple institutions across two countries, using WHO, AJCC, and CAP reference standards (Section \ref{sec:prompt_curation}). A query-independent mean-pooling analysis further confirms that the observed instability persists after removing text-conditioned patch selection (\ref{app:query_independent}).
The benchmark operationalizes this principle through four complementary task families, each isolating a distinct aspect of language sensitivity: zero-shot classification with image--text alignment analysis, cross-modal retrieval, paraphrase robustness under semantic-equivalence paraphrases, length and reporting-style variation, and prompt ensembling, and open-vocabulary diagnosis retrieval over four candidate pools (Section~\ref{sec:benchmark}). Table~\ref{tab:benchmark_comparison} positions PathLang against existing pathology VLM evaluation efforts.

The value of this design is that it answers a question existing benchmarks cannot: \emph{will a model that succeeds under a benchmark designer's chosen prompt continue to succeed when a clinician phrases the same diagnosis differently?} As Table 1 shows, prior benchmarks fall into two groups. Some evaluate classification and retrieval under fixed prompts. Others measure textual sensitivity at the template level without clinical validation. Among the studies we survey, PathLang is the only one that combines diagnostic-level language variation, clinical validation of its text inputs, and all classification and retrieval directions. This combination exposes failure modes that remain hidden under canonical-prompt evaluation: collapse under synonym substitution, near-chance retrieval with organ-agnostic vocabulary, and strong image–text alignment without inter-class separability. PathLang thus complements, rather than replaces, visual-axis benchmarks.
 
Our contributions are threefold. \textbf{(1) A controlled-experiment design for pathology VLM evaluation:} PathLang treats clinical diagnostic language as the manipulated variable, comparing alternative diagnostic-language conditions against identical underlying slides and labels within a fixed image--text evaluation direction. \textbf{(2) A clinically validated prompt corpus and four candidate pools instantiating four complementary tasks:} we curate per-class diagnostic-language variants reviewed by domain experts and 
four candidate pools that isolate distinct forms of semantic competition, supporting zero-shot classification, cross-modal retrieval, paraphrase robustness, and open-vocabulary diagnosis retrieval. \textbf{(3) A nine-model, five-dataset empirical study:} across nine pathology and biomedical VLMs and five public datasets spanning four organs, we find heterogeneous failure modes: instability under clinically equivalent paraphrases, degradation under cross-organ distractors, and no consistent winner across language-robustness settings. We release the prompt corpus, candidate pools, pre-computed text embeddings, and evaluation code. 

\begin{table*}[t]
\centering
\caption{\textbf{Positioning against representative pathology foundation-model and VLM studies.}
Rows are ordered roughly chronologically. A check mark indicates that the work
explicitly evaluates the corresponding axis. \textbf{Cls.}: classification or
downstream diagnostic performance; \textbf{I2I}: image-to-image retrieval;
\textbf{I2T/T2I}: image-to-text and text-to-image retrieval;
\textbf{Lang. Sens.}: explicit evaluation of sensitivity to textual variation;
\textbf{Clin. Valid.}: clinical validation of diagnostic text prompts or
candidate descriptions.}
\label{tab:benchmark_comparison}
\scriptsize
\setlength{\tabcolsep}{3.2pt}
\renewcommand{\arraystretch}{1.08}
\resizebox{0.6\textwidth}{!}{
\begin{tabular}{@{}lccccccc@{}}
\toprule
\textbf{Study}
& \textbf{Cls.}
& \textbf{I2I}
& \textbf{I2T}
& \textbf{T2I}
& \textbf{Slide}
& \textbf{Lang. Sens.}
& \textbf{Clin. Valid.} \\
\midrule

MI-Zero~\citep{mizero}
& \checkmark & -- & -- & -- & \checkmark & -- & -- \\

PLIP~\citep{plip}
& \checkmark & \checkmark & -- & \checkmark & -- & -- & -- \\

Quilt-1M~\citep{quilt1m}
& \checkmark & -- & \checkmark & \checkmark & -- & -- & -- \\

CONCH~\citep{conch}
& \checkmark & -- & \checkmark & \checkmark & \checkmark & -- & -- \\

MUSK~\citep{musk}
& \checkmark & \checkmark & \checkmark & \checkmark & \checkmark & -- & -- \\

Histo-VL~\citep{histovl}
& \checkmark & -- & -- & -- & -- & \checkmark & -- \\

PathBench~\citep{ma2025pathbench}
& \checkmark & -- & -- & -- & \checkmark & -- & -- \\

Clinical FM benchmark~\citep{neidlinger2025benchmarking}
& \checkmark & -- & -- & -- & \checkmark & -- & -- \\

THUNDER~\citep{thunder}
& \checkmark & -- & -- & -- & -- & -- & -- \\

\midrule
\textbf{PathLang (ours)}
& \checkmark & \checkmark & \checkmark & \checkmark
& \checkmark
& \checkmark
& \checkmark \\

\bottomrule
\end{tabular}
}
\end{table*}

\section{Related Works}
\subsection{Computational Pathology VLM Benchmarks}
    
Recent pathology VLM benchmarks primarily evaluate dual-encoder models pretrained on histopathology image--text pairs~\citep{plip,quilt1m,conch,musk,keep,titan,pathgenclip,pathoclip,ahmed2024pathalign,albastaki2025multi} on zero-shot classification, retrieval, and visual robustness~\citep{mizero,thunder,ma2025pathbench,neidlinger2025benchmarking,pathclip_benchmark}; a parallel line probes generative reasoning in decoding-based assistants~\citep{he2020pathvqa,sun2024pathmmu,chen2024wsi,gilal2025pathvlm,chen2025slidechat}, which is outside our scope. Across the dual-encoder line, language remains weakly controlled at two levels: most benchmarks hold prompts canonical and emphasize visual robustness or cross-dataset generalization, while the few that vary text, most notably Histo-VL~\citep{histovl}, limit perturbations to template-level edits (e.g., adding domain-specific carrier phrases) rather than the diagnostic phrasing pathologists actually vary in clinical practice. \emph{Prior pathology benchmarks generally treat language variation as one robustness factor among several, whereas PathLang makes diagnostic-language variation the primary axis of evaluation}: by comparing diagnostic-language conditions against identical slides and labels within a fixed image--text direction, PathLang separates performance variation attributable to diagnostic language from dataset composition and slide-level aggregation effects. 

A recent benchmark in the generative line merits explicit contrast, as it also manipulates the language side of evaluation. OmniPathoVQA~\citep{omnipathovqa2026} curates pathology images from educational materials through a fine-grained extraction pipeline, spanning all human anatomical systems and thousands of disease entities, and decomposes evaluation into pathologic disease identification and microanatomic feature understanding. Its central mechanism is taxonomy-guided difficulty control: hard questions are constructed from the microanatomic features that distinguish similar diseases, yielding a coarse-to-fine protocol that improves benchmark difficulty and clinical granularity. The two designs are orthogonal. OmniPathoVQA increases difficulty in the \emph{answer space}, holding question phrasing canonical while making candidate answers semantically adjacent, and therefore measures knowledge-intensive discrimination among taxonomically neighboring diseases. PathLang holds the diagnostic referent fixed and varies only its surface expression, and therefore measures whether image--text alignment remains stable under clinically equivalent rephrasings. A model that separates taxonomically adjacent entities may still collapse under a synonym substitution, and neither failure mode is observable in the other's setting. Table~\ref{tab:benchmark_comparison} positions PathLang against representative pathology VLM benchmarks across seven evaluation axes.

\subsection{Paraphrase Robustness in Vision--Language Models}

Prompt sensitivity is well-documented in dual-encoder VLMs~\citep{clip}: wording, description style, and prompt context alter behavior across classification, retrieval, and grounding tasks~\citep{yun2021does,deka2026prompt}, and prompt-learning methods such as CoOp~\citep{coop} and CoCoOp~\citep{cocoop} confirm that prompt context substantially affects recognition. In medical domains, the same diagnostic entity is routinely expressed with different synonyms, morphological descriptions, and levels of clinical specificity~\citep{wang2024prompt,hassanein2026prompt}, making prompt variability a clinical reliability concern rather than only a Natural Language Processing (NLP) artifact. Conventional prompt-robustness studies often vary generic carrier templates while preserving the diagnostic term. PathLang instead evaluates changes in diagnostic terminology, clinical specificity, morphological description, and reporting style while preserving the underlying diagnostic meaning. In pathology specifically, Histo-VL~\citep{histovl} reports sensitivity to textual changes but evaluates language as one factor within a broader robustness benchmark; PathLang instead formalizes paraphrase robustness as a controlled evaluation problem, separately measuring stability under clinically equivalent paraphrases, synonym variation, prompt length, and reporting style to distinguish models that succeed under a favorable canonical prompt from those that remain stable under clinically realistic diagnostic-language variation. We do not include a controlled comparison between template-level and diagnostic-level perturbations; PathLang varies diagnostic language only.

\begin{figure*}[t]
    \centering
    \centerline{\includegraphics[width=0.8\textwidth]{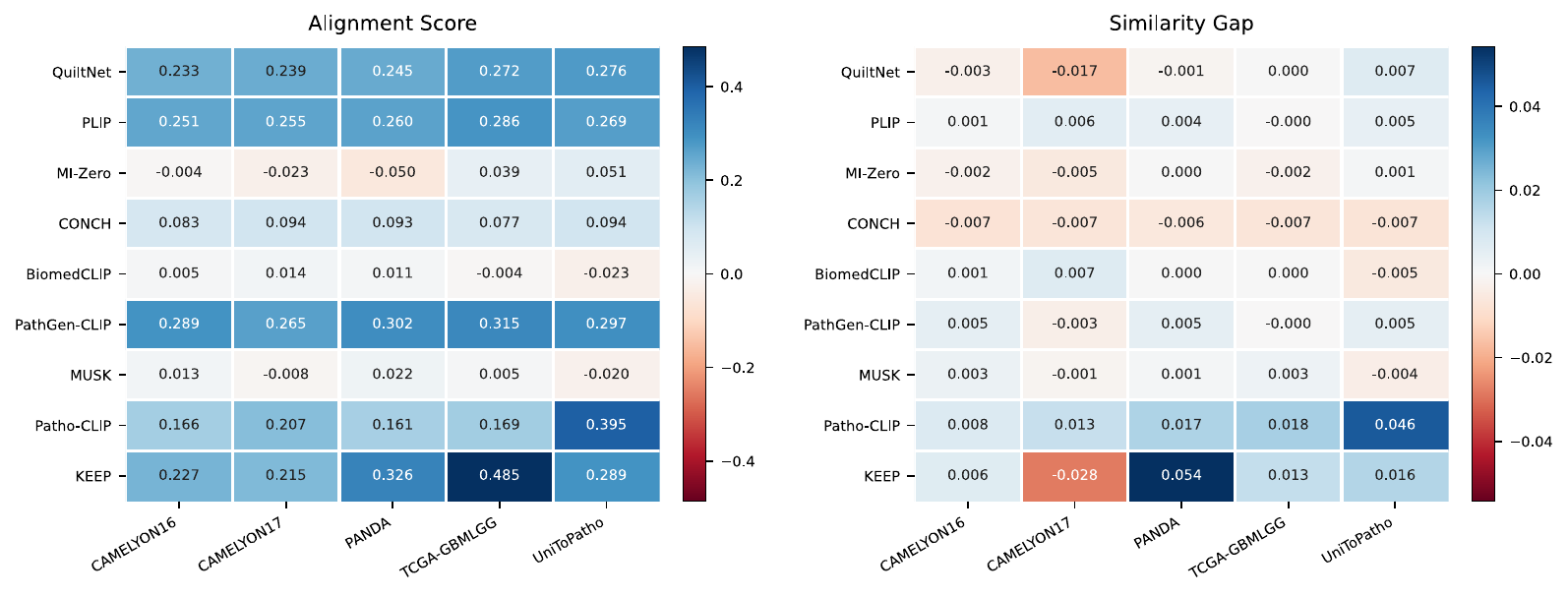}}
    \caption{Zero-shot feature space alignment (left) and discriminative capacity (right) across nine pathology VLMs.  Blue indicates positive alignment and strong class separability, while red denotes negative values and poor alignment or semantic confusion. Results are reported under the upper-bound aggregation protocol (Section~\ref{sec:topk_method}); per-setting results and rank-stability analysis are in \ref{app:k-sensitive}.}
    \label{fig:alignment}
\end{figure*}

\begin{figure*}[t]
    \centering
    \centerline{\includegraphics[width=0.9\textwidth]{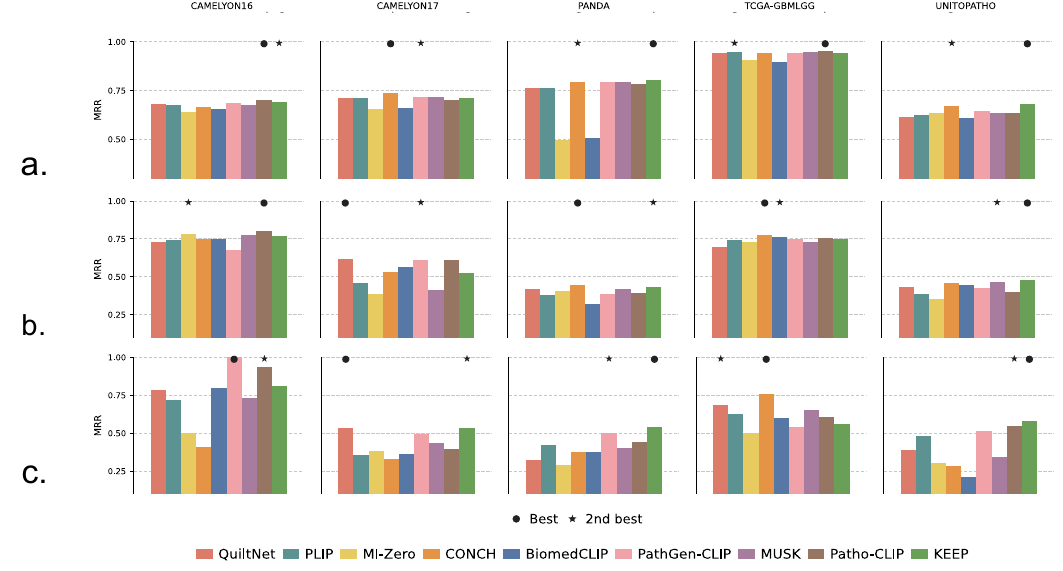}}
    \caption{MRR of cross-modal retrieval (T2). a. I2I, b. I2T, and c. T2I retrieval. Detailed quantitative results are shown in \ref{app:cross_modal_results} (I2I: Table~\ref{tab:crossmodal_I2I}, I2T: Table~\ref{tab:crossmodal_I2T}, T2I: Table~\ref{tab:crossmodal_T2I}). Results are reported under the upper-bound aggregation protocol (Section~\ref{sec:topk_method}); per-setting results and rank-stability analysis are in \ref{app:k-sensitive}.}
    \label{fig:crossmodality}
\end{figure*}

\begin{figure*}[t]
    \centering
    \centerline{\includegraphics[width=0.9\textwidth]{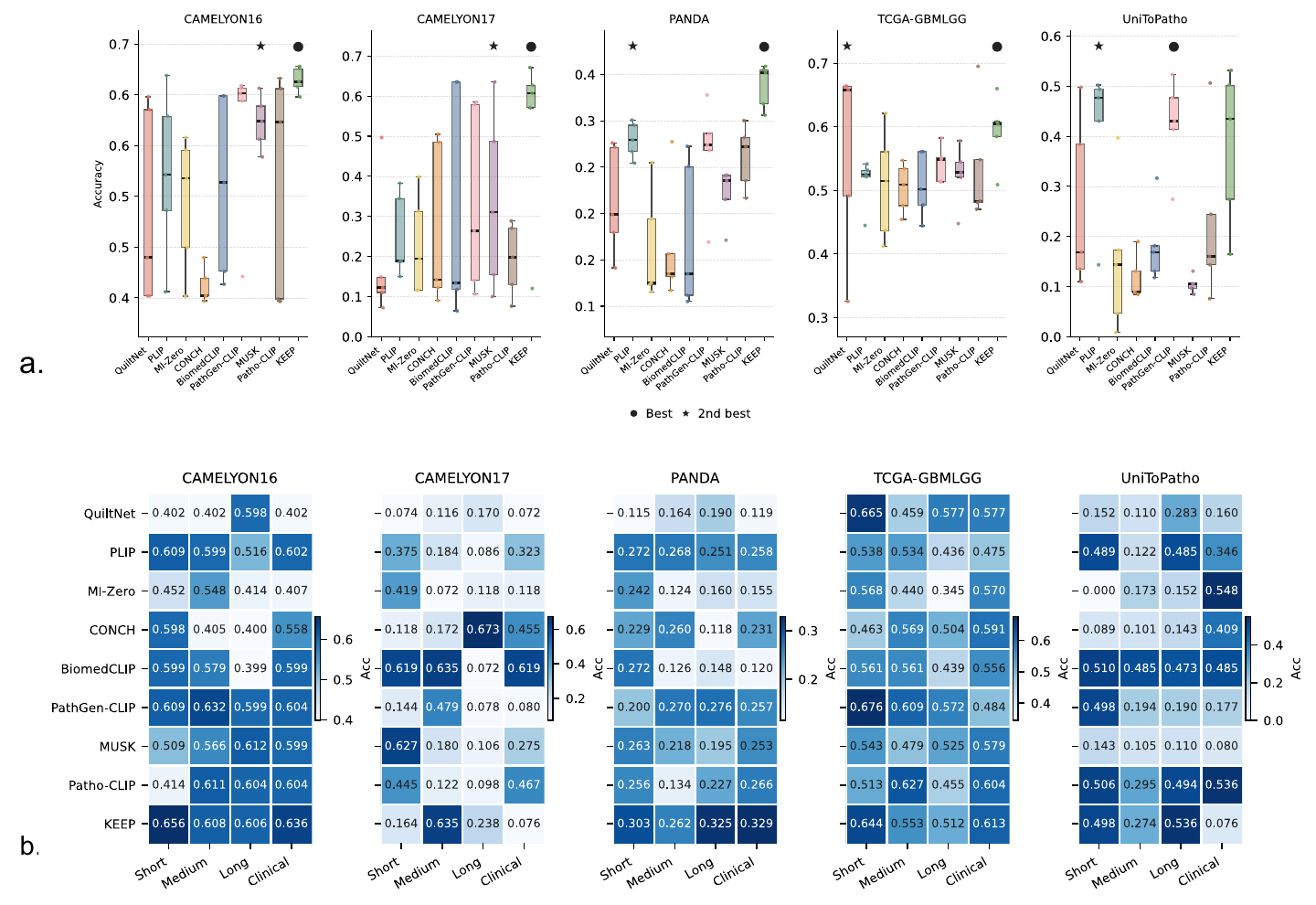}}
    \caption{Results of paraphrase robustness (T3). 
a. Variant~1: accuracy across models and datasets under semantic-equivalence paraphrases. Detailed quantitative results are reported in \ref{app:appendix_paraphrase}, Table~\ref{tab:appendix_paraphrase_synonyms}. b. Variant~2: accuracy stratified by prompt length and reporting style (short, medium, long, and clinical). Detailed quantitative results are reported in \ref{app:appendix_paraphrase}, Table~\ref{tab:appendix_prompt_length}. }
    \label{fig:paraphrase_robustness}
\end{figure*}

\begin{figure*}[htbp]
    \centering
    \includegraphics[width=0.9\textwidth]{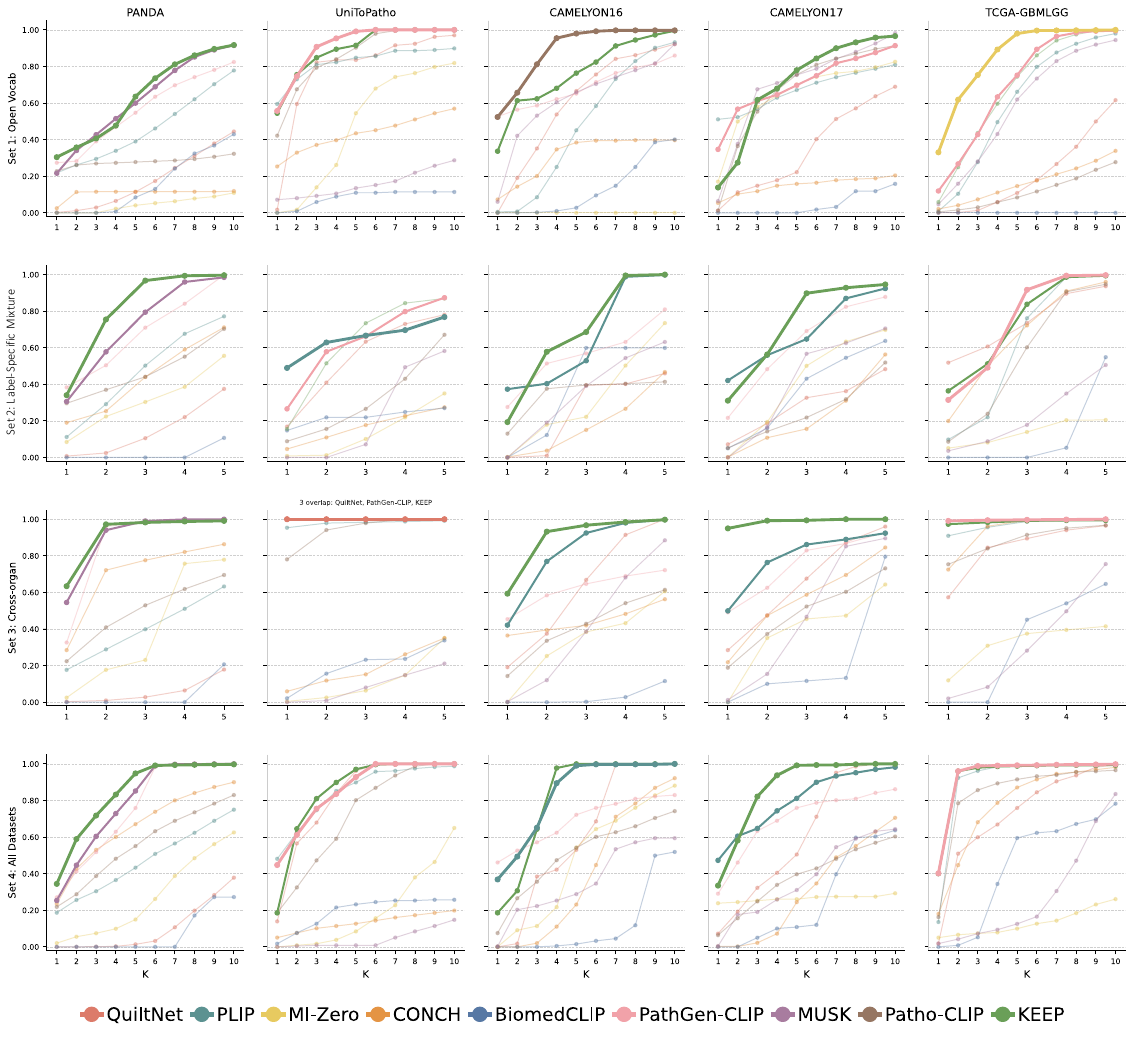}
    \caption{Recall@K of open-vocabulary diagnosis retrieval (T4). Each subplot shows retrieval performance of nine VLMs across five pathology datasets (columns) under four candidate pool settings (rows) that impose distinct forms of semantic competition: open-vocabulary, label-specific mixed, cross-organ, and a global all-datasets pool. Each line represents an individual model; the best and second-best models are highlighted with the thickest and second-thickest curves, respectively. For each dataset and candidate set, models are plotted using their best-performing aggregation setting. Full per-setting results and rank-stability analysis are given in \ref{app:k-sensitive}, where model rankings are stable across settings (Spearman $\rho \geq 0.967$ for all four candidate sets). Detailed quantitative results are reported in \ref{app:appendix_open_vocab} (Set 1: Table~\ref{tab:recall_set1_openvocab}, Set 2: Table~\ref{tab:recall_set2_finegrained}, Set 3: Table~\ref{tab:recall_set3_crossorgan}, Set 4: Table~\ref{tab:recall_set4_alldatasets}).}
    \label{fig:openvocab}
\end{figure*}

\section{PathLang: Language-Centered Benchmarking}\label{sec:benchmark}
\subsection{Overview}

PathLang is organized around four complementary task families that operationalize the controlled-experiment principle (Section~\ref{sec:four_tasks}), supported by a clinically validated prompt and candidate-pool corpus (Section~\ref{sec:prompt_curation}) and a unified patch-to-slide aggregation strategy for fair slide-level comparison across patch-based VLMs (Section~\ref{sec:topk_method}). Experimental protocols and implementation details are provided in Section~\ref{sec:experiment}. Figure~\ref{fig:overview} summarizes PathLang: the left panel shows the evaluated pathology VLMs and five multi-organ datasets; the center panel illustrates the four task families, including zero-shot classification, cross-modal retrieval, paraphrase robustness, and open-vocabulary diagnosis retrieval; and the right panel summarizes language-robustness performance across models and datasets.

\subsection{Four Task Families}\label{sec:four_tasks}
PathLang operationalizes the controlled-experiment principle through four complementary task families. The four families are intentionally orthogonal, with each task isolating a distinct aspect of language robustness while controlling for visual content. The clinically validated prompt and candidate pools used by these tasks are described in Section~\ref{sec:prompt_curation}. 
\textbf{Task 1 (T1) -- Zero-shot Classification.} T1 uses a single canonical diagnostic prompt per class as the baseline reference against which all language-variation experiments are compared, measuring how well each model performs under a canonical diagnostic description. \emph{Example.} For CAMELYON16 ``Metastasis'', the canonical prompt is \textit{``A lymph node containing metastatic carcinoma that forms disruptive nests or clusters effacing normal nodal architecture.''} 
\textbf{Task 2 (T2) -- Cross-modal Retrieval.} T1 uses one canonical diagnostic prompt per class for zero-shot classification. T2 uses a task-specific bank of four expert-reviewed positive descriptions per class, together with the hard-negative pools described in \ref{app:implementation}. The T2 descriptions refer to the corresponding diagnostic classes but are not required to reproduce the T1 canonical prompts verbatim. T2 therefore evaluates retrieval using its own positive-description bank rather than constituting a prompt-matched comparison with T1. This task characterizes cross-modal alignment along each retrieval direction; because the directions differ in candidate-set size, we report them descriptively rather than comparing absolute scores across directions (Section~\ref{sec:experiment_task2}).
\textbf{Task 3 (T3) -- Paraphrase Robustness.} T3 is the core language-sensitivity probe. Holding the underlying slide and label fixed, T3 measures stability under semantic-equivalence paraphrases (Variant~1) and length or reporting-style variation (Variant~2), with prompt ensembling evaluated as Variant~3, drawing from a paraphrase robustness prompt pool. Prompt ensembling is additionally evaluated through an \emph{ensemble gain curve} obtained by averaging multiple prompt embeddings, testing whether each model produces consistent predictions regardless of how a diagnosis is phrased. 
\textbf{Task 4 (T4) -- Open-vocabulary Diagnosis Retrieval.} T4 evaluates retrieval under four candidate pools that impose different forms of semantic competition among clinically plausible diagnoses: organ-agnostic abstract concepts, label-specific mixed candidates, cross-organ candidates, and a global multi-organ candidate set drawn from the union of all five datasets. Together they probe whether models maintain semantic specificity across different sources of diagnostic ambiguity. Full prompt and candidate lists are in \ref{app:appendix_zeroshot_prompt} (T1), \ref{app:appendix_crossmodal_prompt} (T2), \ref{app:appendix_paraphrase_prompt} (T3), and \ref{app:appendix_openvocab_prompt} (T4).

\subsection{Clinically Validated Prompt and Candidate Pool Construction}\label{sec:prompt_curation}

PathLang's controlled-experiment design hinges on the textual inputs that perturb the language modality. We construct two clinically validated resources: a paraphrase robustness prompt pool $\mathbf{P}$ used by T3 to evaluate paraphrase robustness, and four open-vocabulary candidate pools used by T4 for open-vocabulary retrieval. T1 uses a single canonical diagnostic prompt per class as the zero-shot
baseline. T2 uses a separate bank of four expert-validated positive descriptions per class, together with the hard-negative pools described in \ref{app:implementation}.
All prompts and candidate descriptions are reviewed and validated by domain experts to ensure that language perturbations remain medically meaningful while preserving diagnostic content.

\noindent\textbf{Clinical validation protocol.} Prompts and candidate descriptions were drafted, reviewed, and refined by six board-certified pathologists from multiple institutions across two countries, each contributing within their organ-specific clinical expertise. Each candidate text was checked for diagnostic accuracy, terminological correctness, and clinical plausibility using established pathology reference standards and current diagnostic guidelines ~\citep{who_bluebook,ajcc8,cap_protocols,molavi2018,rosai2018}. The final prompt corpus is released alongside the benchmark (\ref{app:promptlist}), and the stages of prompt development and clinical review are documented in \ref{app:clinical_review}, with the review timeline in Table~\ref{tab:prompt_timeline}.
 
\noindent\textbf{Prompt paraphrase robustness variants.} For each (dataset, class) we construct variants along two clinically grounded axes that perturb the diagnostic language while preserving its meaning: \emph{semantic equivalence} (five rephrasings that preserve diagnostic meaning while varying lexical choice and syntax) and \emph{stylistic variation} (four levels of length and formality, from keyword-style to one-sentence to extended morphological to clinical pathology-report). Task 3 combines both axes into a joint pool of all nine variants per class, from which we randomly draw $n \in \{1, \ldots, 9\}$ prompts to form an ensemble; results are reported as a function of ensemble size $n$. \emph{Example.} For CAMELYON16 ``Metastasis'', semantic-equivalence variants include \{\textit{``Metastatic carcinoma involving lymph node''}, \textit{``Nodal metastatic carcinoma''}\}; stylistic variants span \textit{``Metastasis''} (keyword) to \textit{``A lymph node containing metastatic carcinoma effacing normal nodal architecture''} (extended).

\noindent\textbf{Open-vocabulary candidate pools.} For T4 we curate four candidate pools that impose different forms of semantic competition among clinically plausible diagnoses. Set~1 contains abstract organ-agnostic disease concepts; Set~2 uses label-specific candidate pools that combine same-organ diagnostic alternatives with cross-organ and broader pathological distractors, so that organ identity alone is insufficient to isolate the correct answer; Set~3 adds cross-organ candidates that require discrimination across tissue and organ systems; and Set~4 uses the union of ground-truth labels across all five datasets to simulate a global multi-organ retrieval setting. 
Set 2 uses candidate lists specified separately for each target label, whereas Set 3 combines a dataset-level shared cross-organ distractor pool with the label-specific ground-truth candidate. Both sets contain cross-organ competition. Their comparison therefore reflects differences in candidate-pool construction and composition, rather than a controlled same-organ versus cross-organ contrast.
Together these pools present different forms of semantic competition; they differ in construction rather than varying along a single controlled axis. \emph{Example.} For CAMELYON16 ``Metastasis'', Set~1 uses abstract candidates \{\textit{``metastatic carcinoma''}, \textit{``benign tissue''}, \textit{``high-grade malignancy''}\}; Set~2 replaces these with a label-specific pool that mixes nodal alternatives such as \textit{``lymph node micrometastasis''} and \textit{``lymph node isolated tumor cells''} with candidates from other organ systems. 

\subsection{Unified Patch-to-Slide Aggregation}\label{sec:topk_method}
All nine VLMs evaluated in PathLang are patch-level dual encoders: their image encoders take fixed-size tiles rather than whole slides, so a slide-level representation must be constructed externally. Slide-level pathology encoders such as TITAN~\citep{titan} exist but fall outside our scope, since they do not expose the patch-level image--text similarity that PathLang perturbs along the language axis. To evaluate all nine models under a consistent whole-slide protocol, we adopt a single fixed patch-to-slide aggregation rule throughout the benchmark: for each Whole Slide Image (WSI), patch-level image embeddings extracted by each VLM's frozen image encoder are ranked by cosine similarity to a candidate text embedding, and the top-$K$ embeddings are averaged to form the slide-level representation. This aggregation strategy serves as a unified evaluation interface rather than a learned slide-level model. It requires no task-specific training and is applied identically across all VLMs, datasets, prompts, and candidate pools, enabling consistent comparison of language robustness across all tasks. Related settings (patch extraction, embedding normalization, $K$ selection, and task-specific inference) are included in Section~\ref{sec:experiment}; Details are documented in the \ref{app:topk}. To verify that our conclusions are not artifacts of a specific aggregation setting, we repeat all experiments across five settings: four query-dependent selection ratios $r \in \{0.01, 0.05, 0.1, 0.5\}$, denoted Top 1\%, Top 5\%, Top 10\%, and Top 50\%, and query-independent mean pooling over all tiles, denoted Mean (\ref{app:k-sensitive}). Throughout the main text we report, for each model and each evaluation unit, the aggregation setting that maximizes that model's performance on that unit under the task's primary metric; we refer to this as the upper-bound aggregation protocol. The evaluation unit is the (model, dataset) pair for zero-shot classification and paraphrase robustness, and the (model, dataset, candidate pool) triple for open-vocabulary retrieval, since the four pools impose distinct difficulty conditions; within each unit all reported metrics come from a single configuration. For T3, the selected configuration is held fixed across all prompt variants and ensemble sizes within a (model, dataset) pair, so differences across variants reflect the prompt rather than the aggregation setting. The primary metric is accuracy for T1 and T3 and Recall@1 for T4; T2 is reported descriptively (Section 4.3), and its aggregation-sensitivity analysis in Appendix A.1.1 uses nDCG. 
Main-text results use an oracle aggregation setting selected to maximize the task's primary metric among the evaluated configurations within each specified evaluation unit. Because selection uses evaluation labels rather than held-out validation data, these results provide an oracle-selected performance reference, not an estimate of performance under a validation-selected deployment configuration. Maximizing the primary metric does not necessarily minimize prompt-induced variability. The reported instability should therefore not be interpreted as a lower bound across aggregation protocols. Complete per-setting results, a query-independent mean-pooling baseline, and rank-stability analysis across settings are provided in \ref{app:k-sensitive} and \ref{app:query_independent}.

\section{Experiments}\label{sec:experiment}
 
\subsection{Experimental Setup}
\noindent\textbf{Datasets.} Experiments are conducted on five public WSI datasets spanning four organs and multiple classification granularities: \emph{PANDA} (prostate, 6-class ISUP grading, $N=10{,}609$), \emph{UniToPatho} (colon, 6-class colorectal polyp subtyping, $N=237$), \emph{CAMELYON16} (breast, binary lymph-node metastasis detection, $N=399$), \emph{CAMELYON17} (breast, 4-class lymph-node staging, $N=499$), and \emph{TCGA-GBMLGG} (brain, binary GBM/LGG classification, $N=1{,}335$). Slide counts refer to the slides retained after matching whole-slide images to available ground-truth annotations; slides without a resolvable label in the released metadata are excluded. This selection spans binary, multi-class, and fine-grained settings.



\noindent\textbf{Models.}
We evaluate nine pathology and biomedical VLMs that span diverse training corpora, visual backbones, and pretraining objectives:
\emph{CONCH}~\citep{conch},
\emph{PLIP}~\citep{plip},
\emph{KEEP}~\citep{keep},
\emph{PathGen-CLIP}~\citep{pathgenclip},
\emph{Patho-CLIP}~\citep{pathoclip},
\emph{MI-Zero}~\citep{mizero},
\emph{QuiltNet}~\citep{quilt1m},
\emph{MUSK}~\citep{musk}, and
\emph{BiomedCLIP}~\citep{biomedclip}.
All models adopt a dual-encoder architecture that projects histopathology images and natural-language descriptions into a shared embedding space of dimension~$D$, enabling zero-shot inference via cosine similarity without any task-specific fine-tuning.
This selection spans publicly available pathology VLMs at the time of evaluation and includes both domain-specific models (e.g., CONCH, PLIP, MUSK) and broader biomedical models (BiomedCLIP), allowing us to assess whether domain specialization confers language-robustness advantages.

\noindent\textbf{Implementation Details.}
Patches are cropped with 20$\times$ magnification \citep{zhang2025accelerating} consistently across all datasets. Then, we use VLM models to extract patch embeddings, which are aggregated into slide-level representations using a Hybrid Top-$K$ scheme, after which both image and text embeddings are L2-normalized.
For zero-shot classification (T1), each slide is assigned the label of the class with highest cosine similarity.
For cross-modal retrieval (T2), image and text embeddings are ranked by cosine similarity in I2T, T2I, and I2I settings.
Paraphrase robustness experiments (T3) evaluate multiple clinically validated prompt variants independently or via embedding averaging for prompt ensembles.
For open-vocabulary diagnosis retrieval (T4), each slide embedding is matched against its label-specific candidate pool, and candidates are ranked by cosine similarity.
All experiments are conducted on NVIDIA A100 GPUs. 
Detailed descriptions of the Hybrid Top-$K$ aggregation and the implementation of tasks 1--4 are provided in \ref{app:topk}, \ref{app:implementation} respectively.

\noindent\textbf{Metrics.} For zero-shot image classification (T1), we report \emph{accuracy (Acc)}, \emph{macro-F1 (mF1)}, \emph{Area Under the ROC Curve (AUC)}, \emph{alignment score} and \emph{similarity gap}.
For cross-modal and visual retrieval (T2), we report \emph{Recall@1, Mean Reciprocal Rank (MRR)}, and \emph{Normalized Discounted Cumulative Gain (nDCG)}, with \emph{Recall@5} additionally reported for the image-to-image and text-to-image directions.
Paraphrase robustness experiments (T3) are summarized using \emph{accuracy} with \emph{standard deviation, prediction consistency (P.C.)} and \emph{Prompt Stability Score (PSS)} across prompt variants. The 95\% prompt-sampling interval (2.5th--97.5th percentile) and interquartile range of mean accuracy across 100 prompt draws are also reported. For open-vocabulary diagnosis retrieval (T4), 
we evaluate retrieval performance within each candidate pool using \emph{Recall@K} (where $K \in \{1,\ldots,10\}$ for Set 1 and Set 4,  and $K \in \{1,\ldots,5\}$ for Set 2 and Set 3).
Formal definitions of all metrics and additional evaluation details are provided in \ref{app:metrics}.
 
\subsection{T1 -- Zero-shot Classification}\label{sec:experiment_task1}
We evaluate baseline zero-shot performance of nine VLMs using a single canonical text prompt per class (Table~\ref{tab:task1}). Two patterns emerge. First, no single model consistently dominates: model leadership shifts substantially across all datasets (e.g., KEEP leads on PANDA and TCGA-GBMLGG; Patho-CLIP on UniToPatho and CAMELYON16) and metrics (e.g., on CAMELYON17, CONCH leads on accuracy, PathGen-CLIP on
AUC, and Patho-CLIP on macro-F1). Second, alignment quality and inter-class separability are decoupled (Fig.~\ref{fig:alignment}): KEEP and PathGen-CLIP achieve high image-to-text alignment scores yet exhibit negative similarity gaps on certain datasets (notably KEEP on CAMELYON17), indicating that strong image--text matching does not necessarily translate to reliable semantic separation between clinically related classes. Patho-CLIP, in contrast, maintains positive similarity gaps across all datasets, suggesting a feature space with stronger inter-class separability. Because each cell reports a single aggregation configuration, the disagreement across metrics reflects genuine differences in what each metric rewards rather than differences in aggregation.

\noindent\textbf{Takeaway.} Strong canonical-prompt performance does not imply robust semantic discrimination, as alignment quality and inter-class separability are often decoupled across datasets and models.

\begin{table*}[h]
\centering
\caption{Accuracy, AUC and macro-F1 of zero-shot classification (T1).
\textbf{Bold} = best; \underline{underline} = second best; ties at the reported
precision are both marked. For each (model, dataset) pair we report the aggregation
setting that maximizes that model's accuracy on that dataset, so all three metrics
within a cell come from a single configuration (Section~\ref{sec:topk_method}).
Per-setting results and rank-stability analysis are in \ref{app:k-sensitive}.}
\label{tab:task1}
\setlength{\tabcolsep}{4pt}
\resizebox{\textwidth}{!}{%
\begin{tabular}{lccccccccccccccc}
\toprule
\multirow{2}{*}{Model} & \multicolumn{3}{c}{CAMELYON16} & \multicolumn{3}{c}{CAMELYON17} & \multicolumn{3}{c}{PANDA} & \multicolumn{3}{c}{TCGA-GBMLGG} & \multicolumn{3}{c}{UniToPatho} \\
\cmidrule(lr){2-4}\cmidrule(lr){5-7}\cmidrule(lr){8-10}\cmidrule(lr){11-13}\cmidrule(lr){14-16}
 & Acc & AUC & mF1 & Acc & AUC & mF1 & Acc & AUC & mF1 & Acc & AUC & mF1 & Acc & AUC & mF1 \\
\midrule
QuiltNet~\citep{quilt1m} & 0.402 & 0.561 & 0.287 & 0.072 & \underline{0.528} & 0.034 & 0.127 & 0.542 & 0.109 & 0.577 & 0.546 & 0.389 & 0.169 & 0.534 & 0.106 \\
PLIP~\citep{plip} & 0.561 & 0.565 & \textbf{0.550} & 0.323 & \underline{0.528} & \underline{0.212} & 0.250 & 0.506 & 0.088 & 0.603 & 0.390 & 0.505 & \underline{0.376} & 0.529 & \underline{0.199} \\
MI-Zero~\citep{mizero} & 0.407 & 0.461 & 0.292 & 0.176 & 0.494 & 0.084 & 0.160 & 0.493 & 0.090 & 0.444 & 0.331 & 0.325 & 0.152 & 0.478 & 0.145 \\
CONCH~\citep{conch} & 0.578 & 0.527 & 0.423 & \textbf{0.629} & 0.502 & 0.199 & 0.140 & 0.496 & 0.108 & 0.458 & 0.487 & 0.399 & 0.097 & 0.495 & 0.080 \\
BiomedCLIP~\citep{biomedclip} & 0.612 & 0.505 & 0.451 & 0.557 & 0.499 & 0.198 & 0.271 & 0.498 & 0.071 & 0.577 & 0.546 & 0.389 & 0.038 & 0.470 & 0.017 \\
PathGen-CLIP~\citep{pathgenclip} & 0.604 & 0.556 & 0.388 & 0.086 & \textbf{0.533} & 0.065 & \underline{0.283} & 0.539 & 0.149 & 0.489 & \underline{0.555} & 0.486 & 0.181 & \underline{0.568} & 0.157 \\
MUSK~\citep{musk} & 0.607 & \textbf{0.624} & 0.430 & 0.293 & 0.474 & 0.191 & 0.261 & \underline{0.571} & 0.152 & 0.580 & \textbf{0.596} & \underline{0.580} & 0.156 & 0.479 & 0.063 \\
Patho-CLIP~\citep{pathoclip} & \textbf{0.639} & 0.545 & \underline{0.543} & 0.477 & 0.524 & \textbf{0.260} & 0.267 & 0.541 & \underline{0.162} & \underline{0.608} & 0.392 & 0.511 & \textbf{0.544} & \textbf{0.609} & \textbf{0.279} \\
KEEP~\citep{keep} & \underline{0.633} & \underline{0.575} & 0.495 & \underline{0.605} & 0.523 & 0.205 & \textbf{0.344} & \textbf{0.628} & \textbf{0.198} & \textbf{0.613} & 0.542 & \textbf{0.596} & 0.072 & 0.532 & 0.022 \\
\bottomrule
\end{tabular}}
\end{table*}

\begin{table*}[t]
\centering
\caption{Prompt ensembling gain (T3 Variant 3). $L_n$ denotes an ensemble of $n$ prompts drawn uniformly without replacement from the joint pool of nine clinically validated variants (five semantic-equivalence paraphrases and four length or reporting-style variants; Section~\ref{sec:prompt_curation}), with the selected text embeddings averaged and re-normalized before inference. For $n<9$ we report the mean over 100 random draws (seed 42); $n=9$ is deterministic. $L_1$ reports absolute accuracy (range 0--1); $L_2$--$L_9$ report the relative change (\%) with respect to $L_1$, with percentage signs
omitted for compactness. Positive and negative changes are highlighted in green and red, respectively. $L_1$ is the expected accuracy of a single randomly drawn prompt and therefore differs from the canonical-prompt accuracy in Table~\ref{tab:task1}; using it as the reference avoids attributing to ensembling any gain that merely reflects a favorable hand-chosen baseline. Sampling intervals and interquartile ranges across draws are reported in Table~\ref{tab:ci_iqr_all_datasets}. Results are reported under the upper-bound aggregation protocol (Section~\ref{sec:topk_method}); per-setting results and rank-stability analysis are in \ref{app:k-sensitive}.}
\label{tab:ensemble}
\resizebox{\textwidth}{!}{
\begin{tabular}{l ccccccccc ccccccccc ccccccccc}
\toprule
\multirow{2}{*}{Models} 
& \multicolumn{9}{c}{CAMELYON16} 
& \multicolumn{9}{c}{CAMELYON17} 
& \multicolumn{9}{c}{PANDA} \\
\cmidrule(lr){2-10}
\cmidrule(lr){11-19}
\cmidrule(lr){20-28}
& L1 & L2 & L3 & L4 & L5 & L6 & L7 & L8 & L9
& L1 & L2 & L3 & L4 & L5 & L6 & L7 & L8 & L9
& L1 & L2 & L3 & L4 & L5 & L6 & L7 & L8 & L9 \\
\midrule
QuiltNet \citep{quilt1m}
& 0.472 & \dropval{-00.1} & \dropval{-07.6} & \dropval{-07.3} & \dropval{-10.2} & \dropval{-12.7} & \dropval{-14.5} & \dropval{-14.8} & \dropval{-14.8} 
& 0.165 & \dropval{-27.2} & \dropval{-32.0} & \dropval{-36.9} & \dropval{-42.4} & \dropval{-47.8} & \dropval{-48.6} & \dropval{-53.3} & \dropval{-56.2} 
& 0.185 & \gain{+03.4} & \gain{+00.7} & \gain{+06.2} & \gain{+05.2} & \gain{+13.2} & \gain{+17.6} & \gain{+19.3} & \gain{+24.9} \\

PLIP \citep{plip}
& 0.543 & \dropval{-05.0} & \dropval{-04.7} & \dropval{-08.6} & \dropval{-07.2} & \dropval{-08.6} & \dropval{-05.2} & \dropval{-07.7} & \dropval{-10.1} 
& 0.229 & \gain{+04.1} & \dropval{-02.4} & \dropval{-09.5} & \dropval{-14.9} & \dropval{-18.8} & \dropval{-24.7} & \dropval{-24.9} & \dropval{-28.3} 
& 0.265 & \gain{+03.7} & \gain{+06.2} & \gain{+07.1} & \gain{+06.3} & \gain{+05.9} & \gain{+05.2} & \gain{+04.1} & \gain{+03.5} \\

MI-Zero  \citep{mizero}
& 0.481 & \dropval{-00.1} & \dropval{-04.3} & \dropval{-04.9} & \dropval{-07.3} & \dropval{-09.3} & \dropval{-10.7} & \dropval{-11.7} & \dropval{-11.9} 
& 0.189 & \dropval{-23.0} & \dropval{-19.3} & \dropval{-31.9} & \dropval{-34.0} & \dropval{-36.4} & \dropval{-37.3} & \dropval{-37.8} & \dropval{-37.4} 
& 0.158 & \gain{+04.4} & \gain{+03.7} & \dropval{-03.7} & \dropval{-08.4} & \dropval{-08.9} & \dropval{-14.9} & \dropval{-12.7} & \dropval{-17.4} \\

CONCH \citep{conch}
& 0.431 & \dropval{-00.8} & \dropval{-01.0} & \dropval{-04.6} & \dropval{-06.5} & \dropval{-06.8} & \dropval{-06.8} & \dropval{-06.7} & \dropval{-06.7} 
& 0.323 & \dropval{-08.9} & \dropval{-25.1} & \dropval{-18.4} & \dropval{-22.6} & \dropval{-26.4} & \dropval{-27.7} & \dropval{-34.6} & \dropval{-60.3} 
& 0.179 & \dropval{-01.1} & \dropval{-07.5} & \dropval{-09.8} & \dropval{-08.2} & \dropval{-11.5} & \dropval{-18.9} & \dropval{-10.3} & \dropval{-10.4} \\

BiomedCLIP \citep{biomedclip}
& 0.520 & \gain{+04.1} & \gain{+03.5} & \gain{+03.8} & \gain{+07.4} & \gain{+11.3} & \gain{+14.4} & \gain{+15.1} & \gain{+15.1} 
& 0.363 & \gain{+15.7} & \gain{+31.9} & \gain{+34.9} & \gain{+57.7} & \gain{+63.2} & \gain{+72.5} & \gain{+75.0} & \gain{+75.0} 
& 0.162 & \dropval{-00.5} & \gain{+00.7} & \gain{+00.3} & \dropval{-03.9} & \dropval{-05.4} & \dropval{-06.0} & \dropval{-08.7} & \dropval{-14.2} \\

PathGen-CLIP \citep{pathgenclip}
& 0.564 & \gain{+01.5} & \gain{+02.2} & \gain{+04.3} & \gain{+04.1} & \gain{+03.7} & \gain{+03.4} & \gain{+03.4} & \gain{+03.7} 
& 0.298 & \dropval{-44.2} & \dropval{-53.7} & \dropval{-59.9} & \dropval{-66.8} & \dropval{-69.3} & \dropval{-70.9} & \dropval{-71.3} & \dropval{-71.8} 
& 0.259 & \gain{+06.9} & \gain{+12.3} & \gain{+13.0} & \gain{+13.1} & \gain{+13.3} & \gain{+13.7} & \gain{+12.6} & \gain{+11.8} \\

MUSK  \citep{musk}
& 0.571 & \gain{+03.0} & \gain{+04.9} & \gain{+05.5} & \gain{+05.4} & \gain{+05.9} & \gain{+05.9} & \gain{+05.7} & \gain{+07.1} 
& 0.272 & \gain{+56.6} & \gain{+65.9} & \gain{+85.2} & \gain{+93.4} & \gain{+98.2} & \gain{+102.5} & \gain{+112.7} & \gain{+120.4} 
& 0.223 & \gain{+02.1} & \gain{+07.0} & \gain{+08.5} & \gain{+10.4} & \gain{+11.1} & \gain{+11.8} & \gain{+12.3} & \gain{+12.6} \\

Patho-CLIP  \citep{pathoclip}
& 0.556 & \dropval{-00.9} & \gain{+03.0} & \gain{+05.8} & \gain{+07.0} & \gain{+10.1} & \gain{+11.1} & \gain{+11.8} & \gain{+11.8} 
& 0.210 & \dropval{-21.7} & \dropval{-16.2} & \dropval{-42.3} & \dropval{-51.9} & \dropval{-57.1} & \dropval{-63.9} & \dropval{-64.9} & \dropval{-65.7} 
& 0.234 & \gain{+09.7} & \gain{+12.4} & \gain{+15.2} & \gain{+16.1} & \gain{+16.2} & \gain{+16.4} & \gain{+16.1} & \gain{+16.0} \\

KEEP \citep{keep}
& 0.617 & \gain{+00.0} & \gain{+00.0} & \dropval{-00.3} & \dropval{-00.6} & \dropval{-00.7} & \dropval{-00.7} & \dropval{-00.7} & \dropval{-00.6} 
& 0.417 & \gain{+00.1} & \gain{+01.3} & \gain{+14.8} & \gain{+17.6} & \gain{+16.9} & \gain{+17.8} & \gain{+18.4} & \gain{+20.3} 
& 0.322 & \gain{+03.8} & \gain{+06.5} & \gain{+05.4} & \gain{+04.8} & \gain{+04.8} & \gain{+04.4} & \gain{+04.2} & \gain{+03.7} \\

\bottomrule
\end{tabular}
}
\label{tab:prompt_length_recall_change}
\vspace{2pt}

\resizebox{0.66\textwidth}{!}{
\begin{tabular}{l ccccccccc ccccccccc}
\toprule
\multirow{2}{*}{Models} 
& \multicolumn{9}{c}{TCGA-GBMLGG} 
& \multicolumn{9}{c}{UniToPatho} \\
\cmidrule(lr){2-10}
\cmidrule(lr){11-19}
& L1 & L2 & L3 & L4 & L5 & L6 & L7 & L8 & L9
& L1 & L2 & L3 & L4 & L5 & L6 & L7 & L8 & L9 \\
\midrule

QuiltNet \citep{quilt1m}
& 0.557 & \gain{+14.6} & \gain{+15.4} & \gain{+16.3} & \gain{+15.3} & \gain{+15.5} & \gain{+16.2} & \gain{+17.1} & \gain{+17.4}
& 0.217 & \gain{+11.3} & \gain{+10.4} & \gain{+04.4} & \gain{+01.4} & \dropval{-07.9} & \dropval{-16.1} & \dropval{-28.3} & \dropval{-37.9} \\

PLIP \citep{plip}
& 0.504 & \gain{+09.3} & \gain{+11.4} & \gain{+14.8} & \gain{+16.9} & \gain{+17.4} & \gain{+19.7} & \gain{+20.9} & \gain{+23.2}
& 0.360 & \dropval{-02.8} & \dropval{-01.1} & \gain{+05.3} & \gain{+14.3} & \gain{+17.6} & \gain{+18.4} & \gain{+27.3} & \gain{+32.6} \\

MI-Zero \citep{mizero}
& 0.477 & \gain{+08.5} & \gain{+04.6} & \gain{+05.8} & \gain{+03.9} & \gain{+04.5} & \gain{+05.1} & \gain{+02.9} & \gain{+02.3}
& 0.171 & \dropval{-38.6} & \dropval{-53.5} & \dropval{-47.0} & \dropval{-42.0} & \dropval{-36.3} & \dropval{-41.8} & \dropval{-36.1} & \dropval{-30.9} \\

CONCH \citep{conch}
& 0.509 & \gain{+00.0} & \gain{+02.4} & \gain{+02.7} & \gain{+02.1} & \gain{+03.0} & \gain{+04.5} & \gain{+03.5} & \gain{+03.3}
& 0.134 & \gain{+08.5} & \gain{+20.1} & \gain{+15.8} & \gain{+24.3} & \gain{+17.5} & \dropval{-02.8} & \dropval{-04.1} & \dropval{-14.9} \\

BiomedCLIP \citep{biomedclip}
& 0.516 & \gain{+01.9} & \gain{+04.7} & \gain{+05.9} & \gain{+07.8} & \gain{+08.3} & \gain{+08.7} & \gain{+08.7} & \gain{+08.7}
& 0.307 & \dropval{-08.4} & \dropval{-18.4} & \dropval{-12.7} & \dropval{-15.8} & \dropval{-22.6} & \dropval{-39.8} & \dropval{-40.0} & \dropval{-50.6} \\

PathGen-CLIP \citep{pathgenclip}
& 0.556 & \gain{+05.4} & \gain{+09.4} & \gain{+10.3} & \gain{+11.4} & \gain{+11.2} & \gain{+12.1} & \gain{+12.5} & \gain{+12.5}
& 0.339 & \dropval{-01.1} & \dropval{-10.7} & \dropval{-11.0} & \dropval{-17.0} & \dropval{-23.3} & \dropval{-23.7} & \dropval{-24.9} & \dropval{-26.5} \\

MUSK \citep{musk}
& 0.519 & \gain{+00.9} & \gain{+01.7} & \gain{+02.0} & \gain{+02.9} & \gain{+02.8} & \gain{+03.2} & \gain{+04.0} & \gain{+06.0}
& 0.104 & \dropval{-07.8} & \dropval{-08.8} & \dropval{-06.8} & \dropval{-06.8} & \dropval{-06.8} & \dropval{-07.9} & \dropval{-16.3} & \dropval{-19.1} \\

Patho-CLIP \citep{pathoclip}
& 0.541 & \gain{+02.2} & \gain{+00.3} & \dropval{-02.1} & \dropval{-05.1} & \dropval{-05.8} & \dropval{-07.1} & \dropval{-08.4} & \dropval{-09.5}
& 0.320 & \dropval{-11.0} & \dropval{-19.3} & \dropval{-29.0} & \dropval{-34.1} & \dropval{-36.5} & \dropval{-41.0} & \dropval{-45.4} & \dropval{-48.5} \\

KEEP \citep{keep}
& 0.578 & \gain{+06.5} & \gain{+08.7} & \gain{+10.1} & \gain{+09.6} & \gain{+10.1} & \gain{+11.0} & \gain{+11.3} & \gain{+11.9}
& 0.350 & \gain{+12.4} & \gain{+10.6} & \gain{+03.3} & \gain{+03.5} & \gain{+01.0} & \dropval{-00.1} & \dropval{-01.8} & \gain{+00.1} \\

\bottomrule
\end{tabular}
}
\label{tab:language_levels_tcga_unitopatho}
\end{table*}

\subsection{T2 -- Cross-Modal Retrieval}\label{sec:experiment_task2}


T1 evaluates zero-shot classification using one canonical prompt per class, whereas T2 evaluates retrieval using a fixed bank of four positive descriptions per class. T3 examines sensitivity to diagnostic phrasing by systematically varying the prompt text while holding the underlying slides and labels fixed. We evaluate cross-modal alignment across three retrieval directions: I2I, I2T, and T2I (Fig.~\ref{fig:crossmodality}). 
I2I retrieval establishes a strong language-free visual baseline, with most models maintaining high Recall@1, MRR, and nDCG, particularly on TCGA-GBMLGG. Introducing the language modality reveals a substantial gap on most datasets: I2T performance degrades across nearly all models, with pronounced drops on the multi-class datasets PANDA and UniToPatho, and T2I performance varies widely across datasets, with PathGen-CLIP and Patho-CLIP peaking on CAMELYON16 but not elsewhere. Because T2I uses only 8--24 text queries per dataset, single-dataset differences in this direction should be read with caution.


\noindent\textbf{Takeaway.} Averaged across datasets, cross-modal retrieval is weaker than image-to-image retrieval for all nine models, although the gap is dataset-dependent and reverses on CAMELYON16, where several models achieve higher text-to-image than image-to-image MRR. We report this asymmetry as a descriptive observation rather than as direct evidence of weak language grounding: the three directions are not on a common scale; image-to-image relevance is defined by class match, so a random ranking already succeeds at a rate close to the class prior, whereas in the cross-modal directions the correct text or image must be identified from a much larger candidate set, and a modality gap between image and text embeddings is a generic property of contrastive vision--language pretraining that is present at initialization and preserved during training~\citep{liang2022mind}. Evidence for language-side fragility comes instead from T3 and T4, which compare alternative diagnostic-language conditions against identical slides within a fixed image--text direction and therefore hold the modality gap constant. Consistent with this reading, image and text embeddings occupy separate regions of the joint space on CAMELYON17 for both KEEP and Patho-CLIP (Figure ~\ref{fig:cam17-joint-umap}); we treat this as a property of the embedding geometry rather than as further evidence of weak language grounding, for the reasons given there.

\subsection{T3 -- Paraphrase Robustness}\label{sec:experiment_task3}

Whereas T1 evaluates alignment under a single canonical prompt per class and T2 under a fixed bank of four positive descriptions per class, T3 tests robustness to clinically valid diagnostic-language variation by holding the underlying slide and label fixed and varying the diagnostic text through semantic-equivalence paraphrases and length or reporting-style changes (Fig.~\ref{fig:paraphrase_robustness}). As shown in Figure~\ref{fig:paraphrase_robustness}(a), paraphrase robustness varies substantially across both models and datasets: KEEP achieves the best or second best performance on all datasets, while other models show much wider fluctuations under semantic-equivalence paraphrases. Figure~\ref{fig:paraphrase_robustness}(b) further shows that robustness is prompt-length dependent. KEEP exhibits comparatively stable behavior, with relatively smooth accuracy changes across short, medium, long, and clinical prompts in several datasets, whereas QuiltNet and CONCH show stronger length sensitivity, with more pronounced accuracy shifts across prompt formulations. PathGen-CLIP is moderately stable in some datasets but remains dataset-dependent, indicating that no model achieves uniformly robust performance across both prompt length and diagnostic context. Detailed quantitative results for Accuracy, Prediction Consistency, and Prompt Stability Score are provided in \ref{app:appendix_paraphrase}, Tables~\ref{tab:robustness_synonyms} and \ref{tab:robustness_length}. Variant 3 tests whether ensembling multiple prompt variants recovers the stability lost under any single phrasing. We measure gains relative to the expected accuracy of a single randomly drawn prompt rather than the canonical prompt, so that the reference does not depend on which phrasing a benchmark designer happens to select. Table~\ref{tab:ensemble} reports accuracy as a
function of ensemble size $n$. Gains are inconsistent in sign and highly variable in magnitude: on CAMELYON17, MUSK improves by 120.4\% at $n=9$ while PathGen-CLIP degrades by 71.8\% under the same protocol, and on UniToPatho seven of nine models lose accuracy as more prompts are added, with only PLIP improving substantially (+32.6\%). Ensembling therefore averages over the linguistic manifold rather than correcting it: it helps models whose single-prompt performance sits below their pool average, and hurts those whose favorable phrasings are diluted by weaker ones. Consistent with this, the sampling intervals in Table~\ref{tab:ci_iqr_all_datasets} remain wide for most model--dataset pairs. To verify that this instability is not an artifact of the text-conditioned Hybrid Top-K aggregation, we additionally evaluate a query-independent mean-pooling baseline (\ref{app:query_independent}); instability persists under this baseline.



\noindent\textbf{Takeaway.} \emph{Clinically equivalent wording changes induce severe prediction instability.} Many models exhibit large performance fluctuations, near-zero prediction consistency, and negative Prompt Stability Scores under semantically equivalent paraphrases, indicating fragile lexical grounding rather than robust clinical language understanding. Prompt ensembling reduces but does not eliminate this fragility.

\subsection{T4 -- Open Vocabulary Retrieval}\label{sec:experiment_task4}

Building upon the prompt-robustness findings of T3, T4 evaluates whether pathology VLMs maintain semantic specificity under different forms of diagnostic competition. The four candidate pools differ in abstraction, organ scope, and construction: organ-agnostic abstract concepts (Set 1), label-specific mixed candidates (Set 2), cross-organ candidates (Set 3), and a global multi-organ candidate set (Set 4) (Fig.~\ref{fig:openvocab}). Because the pools differ in construction and size, we report a chance baseline for Set 1, whose pool is shared across all slides, and otherwise treat each pool as a separate condition rather than comparing raw scores across sets.
Performance does not decline monotonically with candidate-pool size. The organ-agnostic abstract pool of Set 1 is the hardest condition for most models: on CAMELYON16, chance Recall@1 is 0.050, yet MI-Zero, BiomedCLIP, and QuiltNet all fall at or below this level, and PLIP reaches only 0.005 --- an order of magnitude below chance. By contrast, the cross-organ pool of Set 3 contains half as many candidates and yields substantially higher recall for most models, with KEEP reaching Recall@1 of 0.593 and PLIP 0.421 on the same dataset. Model behavior is also heterogeneous: KEEP maintains high Recall@K trajectories under cross-organ and global settings, while BiomedCLIP remains near zero across most pool configurations.

\noindent\textbf{Takeaway.} Retrieval performance varies substantially across the four forms of semantic competition and does not track pool size: Set 1 contains twice as many candidates as Set 3 yet is the harder condition for most models. Abstraction away from organ-specific language, rather than the number of competing candidates, is the dominant source of difficulty for the evaluated models.

More detailed discussions can be found in \ref{sec:discussion}.


\section{Conclusion}

We propose \textbf{PathLang}, a language-centered, clinically grounded benchmark that isolates the language axis of pathology vision--language models by comparing alternative diagnostic-language conditions against identical slides and labels within a fixed image--text evaluation direction. PathLang evaluates four task families: zero-shot classification, cross-modal retrieval, paraphrase robustness, and open-vocabulary diagnosis retrieval, supported by a clinically validated prompt corpus and four candidate pools. Across nine VLMs and five public datasets spanning four organs, we show that the nine evaluated VLMs are highly unstable to clinically equivalent paraphrases, vary substantially across different forms of semantic competition, and exhibit a mismatch between alignment quality and inter-class separability. These findings suggest that conventional closed-set, canonical-prompt evaluation substantially overestimates pathology VLM language capability. PathLang provides a principled framework for measuring language robustness and highlights linguistic fragility as a key unresolved bottleneck for pathology foundation models.

\noindent\textbf{Limitations.} 
PathLang focuses on patch-level dual-encoder pathology VLMs and evaluates neither slide-level encoders such as TITAN~\citep{titan} nor generative pathology assistants and multimodal reasoning systems. Although prompts were clinically validated across multiple institutions and countries, the benchmark remains constrained to five datasets and a finite set of linguistic perturbations. PathLang evaluates an expert-curated set of clinically valid diagnostic-language variations. The prompt corpus is not intended to estimate the naturally occurring frequency or distribution of diagnostic phrasing across institutions, reporting systems, or clinical populations. Chance baselines are reported only for Set 1 of T4, whose pool is shared across all slides. Because the remaining pools are constructed per label and differ in size across classes, raw scores are not directly comparable across sets. We do not report chance baselines for the cross-modal retrieval directions of T2, which we interpret descriptively rather than as evidence of language-side behavior (Section~\ref{sec:experiment_task2}). Main-text results use an oracle aggregation setting selected on the evaluation data itself rather than on held-out validation data, which inflates absolute performance relative to any deployable configuration. Prompt-induced instability remains observable under the evaluated oracle-selected configurations and in the supplementary fixed-setting analyses, including query-independent mean pooling. These observations show that instability is present under multiple evaluated aggregation choices, but do not establish that it would be equally large or larger under every other configuration. Per-setting results in \ref{app:k-sensitive} show that model rankings are largely stable across settings.

\noindent\textbf{Broader Impact.} PathLang aims to improve evaluation of VLMs by exposing failure modes hidden under canonical closed-set prompts. The benchmark may help reduce overestimation of model capability in real-world deployment. However, pathology VLMs remain unsuitable for autonomous diagnostic use without expert oversight and external validation.

\section*{Acknowledgments}
This research was supported by the WCM Radiology AIMI Fellowship, WCM CTSC 2027 Pilot Award, NIH 1R01HL174863-01A1(Sabuncu), and 1U54DK144866-01(Sabuncu).

\bibliographystyle{model2-names.bst}\biboptions{authoryear}
\bibliography{refs}

\appendix

\makeatletter
\@addtoreset{table}{section}
\@addtoreset{figure}{section}
\makeatother

\section{Additional Experimental Details}

\subsection{Hybrid Top-$K$ Aggregation}\label{app:topk}
All evaluated VLMs produce tile-level embeddings.
To obtain a single slide-level representation we propose a \textbf{Hybrid Top-K} aggregation rule:
\begin{equation}\label{eq:topk}
    K = \max\left\{ K_{\min},\; \lceil r \cdot N \rceil \right\}
\end{equation}
where $N$ is the number of tiles in a slide and $r$ is a proportional ratio.
The top-$K$ tiles (ranked by their cosine similarity to the text query) are selected and their embeddings averaged. Throughout the benchmark we set $K_{\min} = 10$, so that slides with fewer than $10/r$ tiles fall back to a fixed minimum rather than selecting a
degenerate number of tiles.

\paragraph{Motivation}
Pathological findings exhibit extreme spatial heterogeneity.
Focal lesions such as micrometastases occupy a tiny fraction of the slide; a purely proportional rule ($K = \lceil r \cdot N \rceil$) would select too few tiles in small slides, potentially missing the signal entirely.
Conversely, diffuse patterns such as inflammatory infiltrates benefit from a proportional selection that scales with slide size; a fixed $K$ would either under-sample large slides or over-sample small ones.
The hybrid formulation adapts to both regimes: $K_{\min}$ guarantees a minimum number of informative tiles for focal signals, while the proportional term $\lceil r \cdot N \rceil$ scales naturally for diffuse patterns.

We evaluate $r \in \{0.01, 0.05, 0.1, 0.5\}$ as a sensitivity analysis
(\ref{app:k-sensitive}), together with a query-independent baseline that
averages all tile embeddings regardless of the text query
(\ref{app:query_independent}). Main-text results report, for each model, the setting that maximizes its own performance, following the upper-bound protocol described in Section~\ref{sec:topk_method}.

All slide-level embeddings and corresponding slide identifiers are precomputed and stored as NumPy arrays for reproducibility.

\subsubsection{Aggregation Sensitivity Analysis}\label{app:k-sensitive}
We repeat all task variants under $r \in \{0.01, 0.05, 0.1, 0.5\}$ together with query-independent mean pooling. Full per-dataset results are released in the supplementary data \textbf{(https://doi.org/10.5281/zenodo.22221382)}; this section reports dataset-averaged summaries and rank-stability analysis for each task.

\textbf{T1 -- Zero-shot Classification}
As shown in Table~\ref{tab:t1-k-sensitivity-mf1}, the overall zero-shot classification performance is broadly consistent across aggregation settings. Moreover, the corresponding model rankings are highly stable, with pairwise Spearman correlations ranging from 0.883 to 0.967 (Table~\ref{tab:t1-k-sensitivity-spearman}).

\begin{table*}[t]
\centering
\caption{Sensitivity of zero-shot classification performance to aggregation. Each entry is macro-F1 averaged over PANDA, UniToPatho, CAMELYON16, CAMELYON17, and TCGA-GBMLGG. The best aggregation within each model is bolded.}
\label{tab:t1-k-sensitivity-mf1}
\begin{tabular}{lccccc}
\toprule
Model & Top 1\% & Top 5\% & Top 10\% & Top 50\% & Mean \\
\midrule
BiomedCLIP & 0.214 & 0.220 & \textbf{0.252} & 0.176 & 0.218 \\
CONCH & \textbf{0.235} & 0.234 & 0.229 & 0.233 & 0.215 \\
KEEP & 0.277 & 0.300 & \textbf{0.314} & 0.270 & 0.268 \\
MI-Zero & \textbf{0.184} & 0.183 & 0.182 & 0.182 & 0.182 \\
MUSK & \textbf{0.285} & 0.281 & 0.277 & 0.261 & 0.284 \\
PathGen-CLIP & \textbf{0.248} & 0.241 & 0.242 & 0.245 & 0.245 \\
Patho-CLIP & 0.333 & 0.342 & \textbf{0.348} & 0.339 & 0.340 \\
PLIP & 0.278 & 0.282 & 0.279 & 0.282 & \textbf{0.308} \\
QuiltNet & \textbf{0.177} & 0.174 & 0.166 & 0.162 & 0.161 \\
\bottomrule
\end{tabular}
\end{table*}

\begin{table*}[t]
\centering
\caption{Spearman rank correlation ($\rho$) between model rankings under different aggregation settings, computed from cross-dataset macro-F1 in Table~\ref{tab:t1-k-sensitivity-mf1}.}
\label{tab:t1-k-sensitivity-spearman}
\begin{tabular}{lccccc}
\toprule
 & Top 1\% & Top 5\% & Top 10\% & Top 50\% & Mean \\
\midrule
Top 1\% & 1.000 & 0.933 & 0.883 & 0.933 & 0.967 \\
Top 5\% & 0.933 & 1.000 & 0.950 & 0.967 & 0.933 \\
Top 10\% & 0.883 & 0.950 & 1.000 & 0.883 & 0.933 \\
Top 50\% & 0.933 & 0.967 & 0.883 & 1.000 & 0.933 \\
Mean & 0.967 & 0.933 & 0.933 & 0.933 & 1.000 \\
\bottomrule
\end{tabular}
\end{table*}

\textbf{T2 -- Cross-Modal Retrieval} For each setting, we report nDCG averaged over CAMELYON16, CAMELYON17, PANDA,
TCGA-GBMLGG, and UniToPatho. We use nDCG because it provides a rank-sensitive measure that penalizes correct results appearing lower in the retrieved list. Recall@1, Recall@5, and MRR results are provided in the supplementary material.
Tables~\ref{tab:t2-i2t-k-sensitivity-ndcg},
\ref{tab:t2-t2i-k-sensitivity-ndcg}, and
\ref{tab:t2-i2i-k-sensitivity-ndcg} summarize the results for image-to-text,
text-to-image, and image-to-image retrieval, respectively. The results show that the main conclusions are not driven by a single aggregation setting. In particular, model rankings are highly consistent for image-to-text and text-to-image retrieval. Image-to-image retrieval exhibits somewhat greater sensitivity to aggregation, which we quantify below.
To quantify ranking stability, we calculate pairwise Spearman rank
correlations between the nine-model rankings induced by each aggregation
setting. As shown in Tables~\ref{tab:t2-i2t-k-sensitivity-spearman},
\ref{tab:t2-t2i-k-sensitivity-spearman}, and
\ref{tab:t2-i2i-k-sensitivity-spearman}, the correlations range from
$0.767$ to $0.983$ for image-to-text retrieval, from $0.867$ to $0.967$ for
text-to-image retrieval, and from $0.633$ to $0.950$ for image-to-image
retrieval. Thus, the relative ranking of models is robust to the aggregation
setting for the two cross-modal retrieval directions, while the image-to-image
setting warrants a more cautious interpretation.

\begin{table*}[t]
\centering
\caption{Sensitivity of Image-to-Text retrieval to aggregation. Each entry is nDCG averaged over CAMELYON16, CAMELYON17, PANDA, TCGA-GBMLGG, and UniToPatho. The best aggregation within each model is bolded.}
\label{tab:t2-i2t-k-sensitivity-ndcg}
\begin{tabular}{lccccc}
\toprule
Model & Top 1\% & Top 5\% & Top 10\% & Top 50\% & Mean \\
\midrule
BiomedCLIP & 0.665 & \textbf{0.675} & 0.661 & 0.657 & 0.664 \\
CONCH & 0.684 & 0.685 & \textbf{0.689} & 0.664 & 0.672 \\
KEEP & 0.683 & 0.683 & 0.685 & 0.687 & \textbf{0.688} \\
MI-Zero & \textbf{0.642} & 0.638 & 0.638 & 0.635 & 0.640 \\
MUSK & 0.649 & 0.645 & 0.642 & \textbf{0.652} & 0.648 \\
PathGen-CLIP & 0.671 & 0.664 & \textbf{0.673} & 0.666 & 0.671 \\
Patho-CLIP & 0.681 & 0.689 & \textbf{0.689} & 0.678 & 0.681 \\
PLIP & 0.640 & 0.646 & \textbf{0.647} & 0.645 & 0.645 \\
QuiltNet & 0.671 & 0.668 & 0.672 & \textbf{0.677} & 0.674 \\
\bottomrule
\end{tabular}
\end{table*}

\begin{table*}[t]
\centering
\caption{Sensitivity of Text-to-Image retrieval to aggregation. Each entry is nDCG averaged over CAMELYON16, CAMELYON17, PANDA, TCGA-GBMLGG, and UniToPatho. The best aggregation within each model is bolded.}
\label{tab:t2-t2i-k-sensitivity-ndcg}
\begin{tabular}{lccccc}
\toprule
Model & Top 1\% & Top 5\% & Top 10\% & Top 50\% & Mean \\
\midrule
BiomedCLIP & 0.460 & 0.462 & \textbf{0.463} & 0.416 & 0.422 \\
CONCH & \textbf{0.492} & 0.449 & 0.443 & 0.366 & 0.347 \\
KEEP & \textbf{0.633} & 0.613 & 0.589 & 0.599 & 0.579 \\
MI-Zero & \textbf{0.413} & 0.381 & 0.399 & 0.403 & 0.390 \\
MUSK & \textbf{0.518} & 0.481 & 0.462 & 0.458 & 0.505 \\
PathGen-CLIP & 0.602 & 0.583 & \textbf{0.612} & 0.595 & 0.608 \\
Patho-CLIP & 0.552 & \textbf{0.566} & 0.543 & 0.546 & 0.551 \\
PLIP & 0.527 & 0.535 & 0.548 & 0.540 & \textbf{0.550} \\
QuiltNet & 0.554 & 0.529 & 0.539 & \textbf{0.560} & 0.531 \\
\bottomrule
\end{tabular}
\end{table*}

\begin{table*}[t]
\centering
\caption{Sensitivity of Image-to-Image retrieval to aggregation. Each entry is nDCG averaged over CAMELYON16, CAMELYON17, PANDA, TCGA-GBMLGG, and UniToPatho. The best aggregation within each model is bolded.}
\label{tab:t2-i2i-k-sensitivity-ndcg}
\begin{tabular}{lccccc}
\toprule
Model & Top 1\% & Top 5\% & Top 10\% & Top 50\% & Mean \\
\midrule
BiomedCLIP & 0.674 & 0.687 & \textbf{0.693} & 0.685 & 0.692 \\
CONCH & 0.715 & 0.749 & 0.764 & 0.774 & \textbf{0.785} \\
KEEP & 0.729 & 0.749 & 0.759 & 0.785 & \textbf{0.787} \\
MI-Zero & 0.683 & 0.680 & 0.682 & \textbf{0.692} & 0.686 \\
MUSK & 0.671 & 0.705 & 0.727 & 0.769 & \textbf{0.778} \\
PathGen-CLIP & 0.713 & 0.741 & 0.748 & 0.770 & \textbf{0.778} \\
Patho-CLIP & 0.710 & 0.734 & 0.753 & 0.769 & \textbf{0.775} \\
PLIP & 0.698 & 0.726 & 0.729 & 0.757 & \textbf{0.769} \\
QuiltNet & 0.691 & 0.724 & 0.733 & 0.759 & \textbf{0.763} \\
\bottomrule
\end{tabular}
\end{table*}

\begin{table*}[t]
\centering
\caption{Spearman rank correlation ($\rho$) between model rankings under different aggregation settings for Image-to-Text retrieval, computed from cross-dataset nDCG.}
\label{tab:t2-i2t-k-sensitivity-spearman}
\begin{tabular}{lccccc}
\toprule
 & Top 1\% & Top 5\% & Top 10\% & Top 50\% & Mean \\
\midrule
Top 1\% & 1.000 & 0.833 & 0.933 & 0.800 & 0.850 \\
Top 5\% & 0.833 & 1.000 & 0.917 & 0.767 & 0.833 \\
Top 10\% & 0.933 & 0.917 & 1.000 & 0.783 & 0.833 \\
Top 50\% & 0.800 & 0.767 & 0.783 & 1.000 & 0.983 \\
Mean & 0.850 & 0.833 & 0.833 & 0.983 & 1.000 \\
\bottomrule
\end{tabular}
\end{table*}

\begin{table*}[t]
\centering
\caption{Spearman rank correlation ($\rho$) between model rankings under different aggregation settings for Text-to-Image retrieval, computed from cross-dataset nDCG.}
\label{tab:t2-t2i-k-sensitivity-spearman}
\begin{tabular}{lccccc}
\toprule
 & Top 1\% & Top 5\% & Top 10\% & Top 50\% & Mean \\
\midrule
Top 1\% & 1.000 & 0.933 & 0.867 & 0.950 & 0.883 \\
Top 5\% & 0.933 & 1.000 & 0.950 & 0.933 & 0.967 \\
Top 10\% & 0.867 & 0.950 & 1.000 & 0.883 & 0.950 \\
Top 50\% & 0.950 & 0.933 & 0.883 & 1.000 & 0.933 \\
Mean & 0.883 & 0.967 & 0.950 & 0.933 & 1.000 \\
\bottomrule
\end{tabular}
\end{table*}

\begin{table*}[t]
\centering
\caption{Spearman rank correlation ($\rho$) between model rankings under different aggregation settings for Image-to-Image retrieval, computed from cross-dataset nDCG.}
\label{tab:t2-i2i-k-sensitivity-spearman}
\begin{tabular}{lccccc}
\toprule
 & Top 1\% & Top 5\% & Top 10\% & Top 50\% & Mean \\
\midrule
Top 1\% & 1.000 & 0.933 & 0.883 & 0.817 & 0.633 \\
Top 5\% & 0.933 & 1.000 & 0.950 & 0.917 & 0.833 \\
Top 10\% & 0.883 & 0.950 & 1.000 & 0.900 & 0.783 \\
Top 50\% & 0.817 & 0.917 & 0.900 & 1.000 & 0.917 \\
Mean & 0.633 & 0.833 & 0.783 & 0.917 & 1.000 \\
\bottomrule
\end{tabular}
\end{table*}

\textbf{T3 -- Variant 1 (Semantic-Equivalence Paraphrases)} We report Mean Accuracy, which is first averaged over the five
paraphrase variants ($v_1$--$v_5$) within each dataset and then averaged over
PANDA, UniToPatho, CAMELYON16, CAMELYON17, and TCGA-GBMLGG. As shown in
Table~\ref{tab:t3-paraphrase-v1-k-sensitivity-mean-acc}, the overall results
are not driven by a single aggregation choice.
We further quantify ranking stability using pairwise Spearman rank correlations
between the nine-model rankings induced by each aggregation setting. The
correlations range from $0.817$ to $0.983$,
indicating that the relative model ranking is stable across aggregation
strategies (Table~\ref{tab:t3-paraphrase-v1-k-sensitivity-spearman}).

\begin{table*}[t]
\centering
\caption{Sensitivity of T3 Variant~1: semantic-equivalence paraphrases to aggregation. Each entry is Mean Accuracy averaged over PANDA, UniToPatho, CAMELYON16, CAMELYON17, and TCGA-GBMLGG. Mean Accuracy is first averaged across the five paraphrase variants ($v_1$--$v_5$) within each dataset. The best aggregation within each model is bolded.}
\label{tab:t3-paraphrase-v1-k-sensitivity-mean-acc}
\begin{tabular}{lccccc}
\toprule
Model & Top 1\% & Top 5\% & Top 10\% & Top 50\% & Mean \\
\midrule
BiomedCLIP & \textbf{0.338} & 0.332 & 0.335 & 0.334 & 0.334 \\
CONCH & 0.288 & \textbf{0.288} & 0.288 & 0.285 & 0.285 \\
KEEP & 0.480 & 0.482 & \textbf{0.484} & 0.484 & 0.482 \\
MI-Zero & \textbf{0.308} & 0.306 & 0.307 & 0.307 & 0.306 \\
MUSK & \textbf{0.347} & 0.335 & 0.332 & 0.321 & 0.300 \\
PathGen-CLIP & 0.394 & 0.407 & 0.416 & 0.422 & \textbf{0.424} \\
Patho-CLIP & 0.333 & 0.340 & \textbf{0.344} & 0.344 & 0.344 \\
PLIP & 0.383 & 0.390 & 0.390 & 0.392 & \textbf{0.393} \\
QuiltNet & 0.329 & \textbf{0.340} & 0.327 & 0.329 & 0.328 \\
\bottomrule
\end{tabular}
\end{table*}

\begin{table*}[t]
\centering
\caption{Spearman rank correlation ($\rho$) between model rankings under different aggregation settings for T3 Variant~1: semantic-equivalence paraphrases, computed from cross-dataset Mean Accuracy.}
\label{tab:t3-paraphrase-v1-k-sensitivity-spearman}
\begin{tabular}{lccccc}
\toprule
 & Top 1\% & Top 5\% & Top 10\% & Top 50\% & Mean \\
\midrule
Top 1\% & 1.000 & 0.850 & 0.933 & 0.883 & 0.817 \\
Top 5\% & 0.850 & 1.000 & 0.883 & 0.917 & 0.883 \\
Top 10\% & 0.933 & 0.883 & 1.000 & 0.983 & 0.950 \\
Top 50\% & 0.883 & 0.917 & 0.983 & 1.000 & 0.983 \\
Mean & 0.817 & 0.883 & 0.950 & 0.983 & 1.000 \\
\bottomrule
\end{tabular}
\end{table*}

\textbf{T3 -- Variant 2 (Length and Reporting-Style Variation)} We report Mean Accuracy, which is first averaged over short,
medium, long, and clinical paraphrases within each dataset and then averaged
over PANDA, UniToPatho, CAMELYON16, CAMELYON17, and TCGA-GBMLGG. As shown in
Table~\ref{tab:t3-paraphrase-v2-k-sensitivity-mean-acc}, the overall results
are not driven by a single aggregation choice.
We further quantify ranking stability using pairwise Spearman rank correlations
between the nine-model rankings induced by each aggregation setting. All
pairwise correlations equal $1.000$, indicating perfect preservation of the
relative model ranking across aggregation strategies
(Table~\ref{tab:t3-paraphrase-v2-k-sensitivity-spearman}).

\begin{table*}[t]
\centering
\caption{Sensitivity of paraphrase robustness T3 Variant~2: length and reporting-style variation to aggregation. Each entry is Mean Accuracy averaged over PANDA, UniToPatho, CAMELYON16, CAMELYON17, and TCGA-GBMLGG. Mean Accuracy is first averaged across short, medium, long, and clinical paraphrases within each dataset. The best aggregation within each model is bolded.}
\label{tab:t3-paraphrase-v2-k-sensitivity-mean-acc}
\begin{tabular}{lccccc}
\toprule
Model & Top 1\% & Top 5\% & Top 10\% & Top 50\% & Mean \\
\midrule
BiomedCLIP & 0.439 & 0.439 & 0.439 & 0.440 & \textbf{0.442} \\
CONCH & 0.338 & 0.344 & \textbf{0.345} & 0.340 & 0.332 \\
KEEP & 0.417 & 0.419 & 0.421 & \textbf{0.423} & 0.422 \\
MI-Zero & 0.295 & 0.297 & 0.296 & \textbf{0.299} & 0.298 \\
MUSK & \textbf{0.335} & 0.333 & 0.335 & 0.327 & 0.324 \\
PathGen-CLIP & 0.371 & 0.375 & 0.377 & 0.379 & \textbf{0.379} \\
Patho-CLIP & 0.387 & 0.405 & 0.410 & \textbf{0.412} & 0.411 \\
PLIP & 0.376 & 0.378 & 0.380 & 0.384 & \textbf{0.387} \\
QuiltNet & 0.280 & \textbf{0.290} & 0.278 & 0.275 & 0.274 \\
\bottomrule
\end{tabular}
\end{table*}

\begin{table*}[t]
\centering
\caption{Spearman rank correlation ($\rho$) between model rankings under different aggregation settings for T3 Variant~2: length and reporting-style variation, computed from cross-dataset Mean Accuracy.}
\label{tab:t3-paraphrase-v2-k-sensitivity-spearman}
\begin{tabular}{lccccc}
\toprule
 & Top 1\% & Top 5\% & Top 10\% & Top 50\% & Mean \\
\midrule
Top 1\% & 1.000 & 1.000 & 1.000 & 1.000 & 1.000 \\
Top 5\% & 1.000 & 1.000 & 1.000 & 1.000 & 1.000 \\
Top 10\% & 1.000 & 1.000 & 1.000 & 1.000 & 1.000 \\
Top 50\% & 1.000 & 1.000 & 1.000 & 1.000 & 1.000 \\
Mean & 1.000 & 1.000 & 1.000 & 1.000 & 1.000 \\
\bottomrule
\end{tabular}
\end{table*}

\textbf{T4 -- Open-Vocabulary} We report Recall@1, as every query has one ground-truth diagnosis in its
label-specific candidate pool. We retain the four sets as separate conditions,
rather than averaging across them, because they represent distinct types and
levels of diagnostic difficulty. Tables~\ref{tab:t4-sensitivity-set12} and
\ref{tab:t4-sensitivity-set34} show that the results are not driven by one
particular aggregation setting.
We quantify ranking stability by computing pairwise Spearman correlations
between the nine-model rankings induced by different aggregation settings.
Table~\ref{tab:t4-spearman-summary} reports the correlation ranges separately
for each candidate set. The complete pairwise matrices are provided in the supplementary material.

\begin{table*}[t]
\centering
\caption{Aggregation sensitivity for Task 4 open-vocabulary diagnosis retrieval (Set 1-2). Entries are Recall@1 averaged over five datasets; the best aggregation within each model is bolded. Candidate sets are shown separately because they represent distinct difficulty conditions.}
\label{tab:t4-sensitivity-set12}
\begin{minipage}[t]{0.49\textwidth}
\centering
\small
\textbf{Set 1: Open Vocabulary}\\[-2pt]
\resizebox{\linewidth}{!}{%
\begin{tabular}{lccccc}
\toprule
Model & Top 1\% & Top 5\% & Top 10\% & Top 50\% & Mean \\
\midrule
BiomedCLIP & 0.000 & \textbf{0.001} & 0.000 & 0.000 & 0.000 \\
CONCH & \textbf{0.075} & 0.069 & 0.070 & 0.048 & 0.024 \\
KEEP & \textbf{0.262} & 0.249 & 0.254 & 0.257 & 0.256 \\
MI-Zero & 0.089 & 0.098 & 0.095 & 0.099 & \textbf{0.099} \\
MUSK & 0.080 & \textbf{0.088} & 0.088 & 0.080 & 0.077 \\
PathGen-CLIP & 0.328 & 0.350 & 0.354 & 0.360 & \textbf{0.365} \\
Patho-CLIP & 0.221 & 0.235 & 0.229 & \textbf{0.241} & 0.240 \\
PLIP & 0.245 & 0.260 & 0.269 & \textbf{0.273} & 0.272 \\
QuiltNet & \textbf{0.015} & 0.010 & 0.007 & 0.007 & 0.006 \\
\bottomrule
\end{tabular}
}
\end{minipage}
\hfill
\begin{minipage}[t]{0.49\textwidth}
\centering
\small
\textbf{Set 2: Label-Specific Mixed Diagnostic Candidates}\\[-2pt]
\resizebox{\linewidth}{!}{%
\begin{tabular}{lccccc}
\toprule
Model & Top 1\% & Top 5\% & Top 10\% & Top 50\% & Mean \\
\midrule
BiomedCLIP & 0.038 & 0.041 & \textbf{0.042} & 0.040 & 0.040 \\
CONCH & 0.059 & 0.067 & 0.070 & 0.083 & \textbf{0.094} \\
KEEP & \textbf{0.277} & 0.258 & 0.262 & 0.242 & 0.242 \\
MI-Zero & 0.028 & 0.028 & \textbf{0.028} & 0.027 & 0.027 \\
MUSK & \textbf{0.067} & 0.064 & 0.064 & 0.052 & 0.050 \\
PathGen-CLIP & 0.245 & 0.274 & 0.272 & \textbf{0.287} & 0.286 \\
Patho-CLIP & 0.118 & 0.128 & 0.131 & \textbf{0.134} & 0.132 \\
PLIP & 0.289 & 0.297 & 0.298 & 0.302 & \textbf{0.304} \\
QuiltNet & 0.139 & \textbf{0.152} & 0.138 & 0.139 & 0.140 \\
\bottomrule
\end{tabular}
}
\end{minipage}
\end{table*}

\begin{table*}[t]
\centering
\caption{Aggregation sensitivity for Task 4 open-vocabulary diagnosis retrieval (Set 3-4). Entries are Recall@1 averaged over five datasets; the best aggregation within each model is bolded. Candidate sets are shown separately because they represent distinct difficulty conditions.}
\label{tab:t4-sensitivity-set34}
\begin{minipage}[t]{0.49\textwidth}
\centering
\small
\textbf{Set 3: Cross-Organ}\\[-2pt]
\resizebox{\linewidth}{!}{%
\begin{tabular}{lccccc}
\toprule
Model & Top 1\% & Top 5\% & Top 10\% & Top 50\% & Mean \\
\midrule
BiomedCLIP & \textbf{0.010} & 0.008 & 0.006 & 0.004 & 0.003 \\
CONCH & 0.283 & 0.286 & 0.290 & 0.279 & \textbf{0.295} \\
KEEP & \textbf{0.814} & 0.804 & 0.801 & 0.795 & 0.807 \\
MI-Zero & 0.028 & 0.028 & 0.029 & \textbf{0.029} & 0.029 \\
MUSK & 0.064 & 0.084 & 0.099 & 0.112 & \textbf{0.113} \\
PathGen-CLIP & 0.613 & 0.635 & 0.638 & \textbf{0.647} & 0.645 \\
Patho-CLIP & 0.368 & 0.401 & 0.403 & \textbf{0.410} & 0.409 \\
PLIP & 0.548 & 0.573 & 0.581 & 0.590 & \textbf{0.591} \\
QuiltNet & 0.357 & 0.387 & 0.388 & \textbf{0.397} & 0.392 \\
\bottomrule
\end{tabular}
}
\end{minipage}
\hfill
\begin{minipage}[t]{0.49\textwidth}
\centering
\small
\textbf{Set 4: All Datasets}\\[-2pt]
\resizebox{\linewidth}{!}{%
\begin{tabular}{lccccc}
\toprule
Model & Top 1\% & Top 5\% & Top 10\% & Top 50\% & Mean \\
\midrule
BiomedCLIP & \textbf{0.003} & 0.001 & 0.001 & 0.001 & 0.001 \\
CONCH & 0.060 & 0.065 & 0.066 & 0.078 & \textbf{0.089} \\
KEEP & \textbf{0.291} & 0.271 & 0.274 & 0.255 & 0.254 \\
MI-Zero & 0.059 & 0.061 & 0.062 & \textbf{0.062} & 0.062 \\
MUSK & 0.052 & 0.053 & 0.053 & 0.049 & \textbf{0.054} \\
PathGen-CLIP & 0.315 & 0.348 & 0.350 & 0.364 & \textbf{0.368} \\
Patho-CLIP & 0.132 & 0.141 & 0.143 & \textbf{0.145} & 0.140 \\
PLIP & 0.304 & 0.318 & 0.325 & 0.331 & \textbf{0.331} \\
QuiltNet & \textbf{0.047} & 0.045 & 0.043 & 0.045 & 0.043 \\
\bottomrule
\end{tabular}
}
\end{minipage}
\end{table*}

\begin{table*}[t]
\centering
\caption{Ranking stability across aggregation settings for Task 4, based on model rankings by cross-dataset Recall@1. The top-$r$ columns exclude mean aggregation.}
\label{tab:t4-spearman-summary}
\begin{tabular}{lcccc}
\toprule
Candidate set & Min. $\rho$ & Max. $\rho$ & Top-$K$ min. $\rho$ & Top-$r$ max. $\rho$ \\
\midrule
Set 1: Open Vocabulary & 0.983 & 1.000 & 0.983 & 1.000 \\
Set 2: Label-Specific Mixture & 0.967 & 1.000 & 0.967 & 1.000 \\
Set 3: Cross-Organ & 1.000 & 1.000 & 1.000 & 1.000 \\
Set 4: All Datasets & 1.000 & 1.000 & 1.000 & 1.000 \\
\bottomrule
\end{tabular}
\end{table*}

\subsubsection{Query-Independent Baseline}\label{app:query_independent}
To directly test whether the observed instability is an artifact of the text-conditioned Hybrid Top-K aggregation, we repeat T3 (Variants 1 and 2) using a query-independent mean-pooling baseline that averages all patch embeddings regardless of the text query (Table~\ref{tab:gate-v1-meanpool-vs-topk}, \ref{tab:gate-v2-meanpool-vs-topk}). Instability persists under this baseline: 35/45 (Variant 1) and 34/45 (Variant 2) model–dataset pairs exhibit negative Prompt Stability Scores under mean pooling, comparable to or slightly higher than under Hybrid Top-1\% (31/45 and 33/45, respectively). Per-model magnitudes are also closely matched across the two aggregation schemes (e.g., BiomedCLIP: PSS = -0.118 under Top-1\% vs. -0.124 under mean pooling). This confirms that language-induced prediction instability is an intrinsic property of the evaluated models rather than an artifact of query-dependent patch selection.

\begin{table*}[t]
\centering
\caption{Prediction Consistency (P.C.) and Prompt Stability Score (PSS) under Hybrid Top-1\% (query-dependent) versus mean pooling over all patches (query-independent), averaged across five datasets, for semantic-equivalence paraphrases (T3 Variant 1). Instability persists under the query-independent baseline, indicating that it is not an artifact of the text-conditioned aggregation mechanism.}
\label{tab:gate-v1-meanpool-vs-topk}
\begin{tabular}{lcccc}
\toprule
\multirow{2}{*}{Model} & \multicolumn{2}{c}{Hybrid Top-1\% (query-dependent)} & \multicolumn{2}{c}{Mean Pooling (query-independent)} \\
\cmidrule(lr){2-3} \cmidrule(lr){4-5}
 & P.C. & PSS & P.C. & PSS \\
\midrule
BiomedCLIP & 0.004 & -0.118 & 0.004 & -0.124 \\
CONCH & 0.147 & -0.092 & 0.153 & -0.108 \\
KEEP & 0.295 & 0.142 & 0.308 & 0.113 \\
MI-Zero & 0.065 & -0.018 & 0.070 & -0.028 \\
MUSK & 0.128 & 0.063 & 0.155 & -0.048 \\
PathGen-CLIP & 0.115 & 0.030 & 0.117 & -0.002 \\
Patho-CLIP & 0.070 & -0.047 & 0.071 & -0.066 \\
PLIP & 0.044 & -0.016 & 0.043 & -0.042 \\
QuiltNet & 0.010 & -0.090 & 0.002 & -0.134 \\
\bottomrule
\end{tabular}
\end{table*}
 
\begin{table*}[t]
\centering
\caption{Prediction Consistency (P.C.) and Prompt Stability Score (PSS) under Hybrid Top-1\% (query-dependent) versus mean pooling over all patches (query-independent), averaged across five datasets, for length and reporting-style variation (T3 Variant 2).}
\label{tab:gate-v2-meanpool-vs-topk}
\begin{tabular}{lcccc}
\toprule
\multirow{2}{*}{Model} & \multicolumn{2}{c}{Hybrid Top-1\% (query-dependent)} & \multicolumn{2}{c}{Mean Pooling (query-independent)} \\
\cmidrule(lr){2-3} \cmidrule(lr){4-5}
 & P.C. & PSS & P.C. & PSS \\
\midrule
BiomedCLIP & 0.170 & -0.185 & 0.179 & -0.227 \\
CONCH & 0.026 & -0.167 & 0.075 & -0.165 \\
KEEP & 0.261 & 0.050 & 0.259 & 0.029 \\
MI-Zero & 0.107 & -0.125 & 0.115 & -0.141 \\
MUSK & 0.198 & 0.013 & 0.042 & -0.100 \\
PathGen-CLIP & 0.283 & 0.055 & 0.262 & 0.029 \\
Patho-CLIP & 0.101 & -0.072 & 0.126 & -0.105 \\
PLIP & 0.160 & -0.024 & 0.184 & -0.033 \\
QuiltNet & 0.034 & -0.211 & 0.033 & -0.243 \\
\bottomrule
\end{tabular}
\end{table*}

\subsection{Detailed Implementation of Tasks}\label{app:implementation}
\textbf{Task 1.} For each dataset, we encode one canonical text prompt per class using the VLM’s text encoder. Slide-level image embeddings are obtained using Hybrid Top-K aggregation (Section~\ref{sec:topk_method}). We then compute the cosine similarity matrix and select the class with the highest similarity score as the classification result.
Task~1 evaluates each model's zero-shot generalization capacity, reflecting the quality of vision-language alignment acquired during pre-training without any task-specific adaptation or labelled data. This setting enables a fair, controlled comparison across models and provides an upper-bound reference for data-efficient learning scenarios.

\textbf{Task 2.} We evaluate three retrieval directions to comprehensively 
assess the quality of cross-modal alignment.

For each dataset, we construct a class-specific distractor pool containing 
both positive and negative text descriptions. Each class is associated with 
four positive descriptions (paraphrases of the ground-truth class) and a 
set of hard negatives drawn from other classes within the same dataset as 
well as semantically related classes from other datasets. This cross-dataset 
contamination of negatives is intentional: it stress-tests whether models 
can distinguish diagnostically relevant concepts from superficially similar 
but incorrect ones.

\begin{itemize}
    \item {\textbf{Image-to-Text Retrieval}}\label{sec:i2t}
Given a slide-level image embedding as the query, we rank all text 
candidates in the dataset-specific distractor pool by cosine similarity. 
The pool contains all positive and negative descriptions across all classes 
of the dataset. 
For each query, we identify the rank of the first text whose class matches the ground truth and report Recall@1, MRR, and nDCG. Although each class is represented by four positive descriptions in the distractor pool, we adopt a single-relevant-item formulation for evaluation, with the first class-matching result defining the relevant result for each query.
    \item {\textbf{Text-to-Image Retrieval}}\label{sec:t2i}
We reverse the direction: a text embedding serves as the query and all 
slide-level image embeddings form the database. For each class, we use all four positive descriptions as independent queries, each ranked against the full image embedding set by cosine similarity. Results are averaged across the four queries per class and evaluated using the same single-relevant-item metric formulation described above.
    \item {\textbf{Image-to-Image Retrieval}}\label{sec:i2i}
Each slide embedding is used as a query against all other slide embeddings 
(excluding itself) in the same dataset. We compute the full pairwise cosine 
similarity matrix, remove the diagonal (self-match), and sort each row in 
descending order.
\end{itemize}

Task 2 characterizes cross-modal alignment along three retrieval directions. Because the three directions differ in candidate-set size and in what counts as a relevant result, their absolute scores are not on a common scale and we do not interpret differences between directions as evidence about which encoder is responsible for misalignment (Section~\ref{sec:experiment_task2}). Direction-wise results are reported descriptively; evidence for language-side behavior comes from T3 and T4, which vary the diagnostic language within a fixed retrieval direction.

\textbf{Task 3.} 
We hold the underlying slide and ground-truth label fixed while systematically varying the textual prompt. Slide-level representations are constructed with the text-conditioned Hybrid Top-$K$ rule of Section~\ref{sec:topk_method}; a query-independent mean-pooling variant is reported in \ref{app:query_independent}.

\paragraph{Prompt pool construction}
For each dataset we construct a clinically validated prompt pool, whose
variation types are summarized in Table~\ref{tab:t3_variation_types}. All prompt embeddings are L2-normalized before evaluation.

\begin{table*}[t]
\centering
\caption{Prompt variation types used in the T3 prompt pool.}
\label{tab:t3_variation_types}
\begin{tabular}{ll}
\toprule
\textbf{Variation Type} & \textbf{Description} \\
\midrule
Synonym/Paraphrase & Semantically equivalent rewrites of the canonical class description \\
Length: Short & Concise, keyword-style prompt \\
Length: Medium & Moderately detailed prompt \\
Length: Long & Extended prompt with additional clinical context \\
Clinical Style & Formal clinical reporting style \\
\bottomrule
\end{tabular}
\end{table*}

Prompts were authored by domain experts and reviewed for clinical 
validity.

\textbf{Variant~1: Semantic-equivalence paraphrases.}
Each of the five paraphrase prompts (v1-v5) is evaluated 
independently as a zero-shot classifier.
We report per-variant accuracy, balanced accuracy, standard deviation, 
\textbf{Prediction Consistency} (fraction of slides where all five 
variants produce identical predictions), and the \textbf{Prompt 
Stability Score (PSS)}, defined as Krippendorff's $\alpha$ computed 
over the matrix of per-variant predictions under a nominal measurement 
scale. A PSS of 1.0 indicates perfect inter-variant agreement; values 
near or below 0 indicate chance-level consistency.

\textbf{Variant~2: Length and reporting-style variation.}
The four length levels (short, medium, long, clinical; v6-v9) 
are each evaluated independently.
We report the same metrics as Variant~1.
This variant isolates the effect of prompt verbosity and style on model 
predictions.

\textbf{Variant~3: Prompt ensemble gain.}
We measure how accuracy changes as more prompt variants are ensembled.
For each ensemble size $n \in \{1, 2, \ldots, 9\}$:
\begin{itemize}[leftmargin=*,nosep]
    \item If $n < 9$: we draw $n$ prompt indices uniformly at random 
    (without replacement) from the full pool of all 9 variants 
    (paraphrases and length levels combined), reflecting a realistic 
    scenario where prompt selection is agnostic to variant type. The 
    selected embeddings are averaged and re-normalized before computing 
    zero-shot accuracy. This is repeated 100 times (seed = 42), and we 
    report the mean and standard deviation across draws.
    \item If $n = 9$: the result is deterministic (all prompts averaged).
\end{itemize}
This analysis reveals whether ensembling diverse prompts can compensate 
for individual prompt sensitivity, and at what ensemble size the 
marginal gain saturates.

Variant~1 and Variant~2 test whether models are robust to clinically equivalent language variation: if so, accuracy, prediction consistency, and PSS should remain high across all variants. Variant~3 tests whether a simple mitigation strategy (prompt ensembling) can recover robustness, and quantifies the cost of relying on a single prompt.

\textbf{Task 4.}
We evaluate whether VLMs can retrieve the correct diagnosis from label-specific candidate pools that differ in the type of semantic competition they impose.

\paragraph{Candidate pool construction}
Each slide is evaluated against its own label-specific candidate pool containing one ground-truth description and a set of distractors. Pools were assembled independently per label from expert-provided candidate descriptions; in a small number of pools the same string appears twice, so total and unique counts differ. Duplicate strings yield identical embeddings and therefore occupy adjacent ranks; we score the highest-ranked occurrence of the ground-truth description. Set 4 draws on the union of ground-truth labels across all five datasets, and because CAMELYON16 and CAMELYON17 share the label normal lymph node, this union contains 19 unique candidates.

\begin{enumerate}[leftmargin=*,nosep]
    \item \textbf{Set~1 --- Open Vocabulary}: candidates are 
    abstract, organ-agnostic diagnostic terms (e.g., 
    ``low-grade malignancy'', ``metastatic carcinoma'', 
    ``high-grade dysplasia''). The ground truth is mapped to 
    the closest such abstraction for each class. This tests 
    whether models can match slides to coarse semantic concepts 
    without organ-specific language.

    \item \textbf{Set~2 --- Label-Specific Mixture}: pools are specified separately for each target label and combine same-organ diagnostic alternatives (e.g.,``ISUP Grade 2 prostate cancer'',``tubular adenoma high-grade'') with cross-organ and broader pathological distractors. Their composition varies by target label, and same-organ candidates are not necessarily the majority. The correct answer therefore cannot be identified by organ recognition alone, but the task is not an isolated measure of within-organ discrimination.

    \item \textbf{Set~3 --- Cross-Organ}: candidates include 
    the correct organ-specific diagnosis alongside distractors 
    from other organ systems evaluated in the benchmark 
    (e.g., prostate, colon, brain, lymph node, breast). This 
    tests whether models can correctly localize the organ of 
    origin in addition to identifying the diagnosis.

    \item \textbf{Set~4 --- All Datasets}: candidates are the union of ground-truth labels across all five datasets. 
\end{enumerate}

\subsection{Detailed Metrics}\label{app:metrics}
For task 1, we use \begin{itemize}[leftmargin=*,nosep]
    \item \textbf{Accuracy}: fraction of correctly classified slides.
    \item \textbf{Macro F1}: unweighted mean of per-class F1 scores, 
    mitigating class-imbalance effects.
    \item \textbf{AUC}: for binary tasks (CAMELYON16, TCGA-GBMLGG) we 
    report the ROC-AUC using the positive-class similarity score as the 
    decision function; for multi-class tasks (PANDA, UniToPatho, 
    CAMELYON17) we report the macro-averaged one-vs-rest AUC over the 
    full similarity matrix.
    \item \textbf{Alignment Score}: mean cosine similarity between each 
    slide embedding and its ground-truth class text embedding, 
    $\frac{1}{N}\sum_{i=1}^{N} S_{i,y_i}$, measuring how well the 
    pre-trained vision-language features are aligned for the correct class.
    \item \textbf{Similarity Gap}: mean difference between the 
    ground-truth class similarity and the highest competing class 
    similarity, $\frac{1}{N}\sum_{i=1}^{N}\bigl(S_{i,y_i} - 
    \max_{c \neq y_i} S_{i,c}\bigr)$, quantifying the model's ability 
    to discriminate between classes in feature space. A positive gap 
    indicates correct separation, while values near zero or negative 
    suggest ambiguous or incorrect alignment.
\end{itemize}

For Task 2, we use the following metrics:
\begin{itemize}[leftmargin=*,nosep]
    \item \textbf{Recall@K}: fraction of queries for which the correct 
    candidate appears within the top-$K$ ranked results.
    \item \textbf{MRR} (Mean Reciprocal Rank): average of the reciprocal 
    rank of the first correct result across all queries, 
    $\frac{1}{N}\sum_i \frac{1}{\text{rank}_i}$, rewarding models that 
    rank the correct candidate higher.
    \item \textbf{nDCG} (Normalized Discounted Cumulative Gain): \(\frac{1}{N}\sum_i \frac{1}{\log_2(\mathrm{rank}_i+1)}\), where \(\mathrm{rank}_i\) denotes the rank of the first class-matching result for query \(i\). This single-relevant-item formulation applies a logarithmic penalty to correct results ranked lower.
\end{itemize}

For Task 3, we report the following metrics across all three variants:
\begin{itemize}[leftmargin=*,nosep]
    \item \textbf{Accuracy}: fraction of correctly classified slides for 
    each prompt variant or length level.
    \item \textbf{Balanced Accuracy}: mean per-class recall, mitigating 
    class-imbalance effects.
    \item \textbf{Accuracy Std}: standard deviation of accuracy across 
    variants, quantifying sensitivity to prompt wording.
    \item \textbf{Prediction Consistency}: fraction of slides for which 
    all variants within a group produce identical predictions, regardless 
    of correctness.
    \item \textbf{Prompt Stability Score (PSS)}: Krippendorff's $\alpha$ 
    computed over the matrix of per-variant predictions 
    (nominal scale), measuring inter-variant agreement. A value of 1.0 
    indicates perfect agreement; values near 0 indicate chance-level 
    consistency.
    \item \textbf{Mean Accuracy vs.\ Ensemble Size}: For Variant~3 (ensemble gain), we additionally report: for each ensemble size $n \in \{1, \ldots, 9\}$, we randomly sample (without 
    replacement) $n$ prompt indices from the full pool of 9, average the 
    selected text embeddings, re-normalize, and compute zero-shot 
    accuracy. For $n < 9$, this is repeated 100 times (seed = 42) and 
    we report the mean and standard deviation across draws; for $n = 9$ 
    the result is deterministic.
    \item 95\% prompt-sampling interval: the 2.5th and 97.5th percentiles of accuracy across the 100 random prompt draws for a given ensemble size $n$. This interval describes sensitivity to which prompts are drawn, not statistical uncertainty in the accuracy estimate.
    \item \textbf{Interquartile Range (IQR)}: the difference between the 
    75th and 25th percentiles of accuracy across draws, measuring the 
    spread of the middle 50\% of results.
    \item \textbf{Bootstrap confidence intervals}: for Prediction Consistency 
    and PSS we resample slides with replacement 1{,}000 times (seed 42) and 
    recompute both quantities on each resample, reporting the 2.5th and 97.5th 
    percentiles. Resampling is over slides rather than prompt variants, so the 
    intervals reflect uncertainty due to the finite slide set. In no 
    model--dataset combination did resampling produce a degenerate prediction 
    matrix for which Krippendorff's $\alpha$ is undefined.
\end{itemize}

For Task 4, we report the following metrics for each of the four 
candidate set configurations:
\begin{itemize}[leftmargin=*,nosep]
    \item \textbf{Recall@K}: fraction of slides for which the 
    ground-truth text description appears within the top-$K$ ranked 
    candidates, where $K \in \{1,\ldots,10\}$ for Set~1 and Set~4, 
    and $K \in \{1,\ldots,5\}$ for Set~2 and Set~3.
    \item \textbf{MRR} (Mean Reciprocal Rank): 
    $\frac{1}{N}\sum_i \frac{1}{\text{rank}_i}$, where $\text{rank}_i$ is the position of the ground-truth description in the ranked 
    candidate list for slide $i$.
    \item \textbf{nDCG} (Normalized Discounted Cumulative Gain): 
    $\frac{1}{N}\sum_i \frac{1}{\log_2(\text{rank}_i + 1)}$, assuming 
    a single relevant item per query.
    \item \textbf{Chance baselines (Set 1).} For Set 1, all slides are ranked against the same 20-candidate pool containing a single ground-truth description, so chance performance under a uniformly random ranking is Recall@k = k/20, identical across datasets. We do not report chance levels for Sets 2--4, whose pools are constructed per label and differ in size across classes; these sets are therefore interpreted as separate conditions rather than against a common baseline.
\end{itemize}

\subsection{Prompt Development and Clinical Review}\label{app:clinical_review}

The prompt corpus was developed iteratively through prompt generation,
distractor construction, expert review, and subsequent refinement; the
timeline is given in Table~\ref{tab:prompt_timeline}. Representative
term-level revisions from the post-review verification round are listed in
Table~\ref{tab:rev_a}, and those from the final revision round, in which
distractor candidates judged implausible or non-discriminative were replaced,
in Table~\ref{tab:rev_c}. Revisions primarily improved diagnostic
terminology, specificity, and clinical plausibility while preserving the
intended diagnostic concepts.

Clinical review and subsequent verification were performed at the level of individual prompt entries rather than through a corpus-wide list of prohibited terms. Table~\ref{tab:rev_a} provides representative examples of entry-specific revisions, including removal of primary-site or nodal-designation wording when judged insufficiently supported or unnecessarily specific for the entry under review. These changes do not imply that terms such as ``breast'', ``mammar'', or ``sentinel node'' were excluded from every prompt. The final wording of the prompt corpus is provided in \ref{app:promptlist}.

\begin{table*}[t]
\centering
\caption{Prompt development and clinical review timeline.}
\label{tab:prompt_timeline}
\begin{tabularx}{\textwidth}{p{0.20\textwidth} p{0.13\textwidth} X}
\toprule
\textbf{Stage} & \textbf{Date} & \textbf{Description} \\
\midrule

Initial prompt set
& Nov.~18, 2025
& Initial ground-truth/zero-shot prompts were generated and shared. \\

Distractor revision
& Dec.~8, 2025
& A distractor-augmented prompt set was received; issues with the initial distractors were identified and revised. \\

Expert review
& Apr.~9, 2026
& The submitted prompts were reviewed and revised by the clinical expert(s), and the revised prompt set was returned. \\

Post-review verification
& Apr.~29, 2026
& The revised prompts were applied and tested, leading to identification of additional issues for correction. \\

Final revision
& May~3, 2026
& A subsequent revised version incorporating the requested changes was received and used as the final prompt set. \\

\bottomrule
\end{tabularx}
\end{table*}

%

\begin{table*}[t]
\centering
\footnotesize
\caption{Representative entry-specific prompt revisions from the post-review verification round (Apr. 29, 2026). Categories group the listed revisions by rationale; review was conducted entry by entry.}
\label{tab:rev_a}
\begin{tabularx}{\textwidth}{p{0.20\textwidth} X X}
\toprule
\textbf{Revision type} & \textbf{Original} & \textbf{Revised} \\
\midrule

Dysplasia terminology
& low-grade cytologic atypia & low-grade cytologic dysplasia \\
& severe epithelial atypia & severe epithelial dysplasia \\
& marked cytologic atypia & marked cytologic dysplasia \\
& mild atypia (colorectal adenoma) & mild dysplasia \\
& high-grade intraepithelial neoplasia & high-grade epithelial dysplasia \\
& tubular colorectal adenoma with mild dysplasia & tubular colorectal adenoma with low-grade dysplasia \\
& tubulovillous colorectal adenoma with mild dysplasia & tubulovillous colorectal adenoma with low-grade dysplasia \\
& no cytologic or invasive malignant features & no cytologic atypia or invasive malignant features \\
\midrule

Unverifiable primary site
& metastatic breast carcinoma in lymph node & metastatic carcinoma involving lymph node \\
& sentinel node involved by breast carcinoma & sentinel node involved by carcinoma \\
& nodal metastasis from breast carcinoma & nodal metastatic carcinoma \\
& lymph node with metastatic mammary carcinoma & lymph node with metastatic carcinoma \\
& no metastatic breast carcinoma, tumor nests, or epithelial cell clusters & no metastatic carcinoma \\
& metastatic breast carcinoma deposits & metastatic carcinoma deposits \\
& malignant epithelial nests or clusters from breast origin & malignant epithelial nests or clusters \\
\midrule

Unverifiable nodal designation
& benign sentinel lymph node & benign lymph node \\
& uninvolved sentinel node & uninvolved lymph node \\
& negative sentinel lymph node & negative lymph node \\
& sentinel node containing only scattered single cells or tiny clusters & lymph node containing only scattered single cells or tiny clusters \\
& the sentinel node contains a limited metastatic epithelial focus & the lymph node contains a limited metastatic epithelial focus \\
\midrule

Explicit diagnostic criteria
& minimal nodal tumor burden & minimal nodal tumor burden less than 200 tumor cells \\
& serrated hyperplastic colorectal polyp & serrated hyperplastic colorectal polyp, with serrations limited to the upper half of the crypts \\
& low-grade acinar carcinoma of prostate & low-grade acinar carcinoma of prostate with individual discrete well-formed glands \\
& acinar prostate cancer with emerging fused glands & acinar prostate cancer with focal emerging fused glands \\
& prostate carcinoma with fused and cribriform glands & prostate carcinoma with predominantly fused and cribriform glands \\
& prostatic adenocarcinoma composed of fused and cribriform glands & prostatic adenocarcinoma composed of fused and cribriform glands only \\
& high-grade gland-forming prostate malignancy with cribriform architecture & high-grade gland-forming prostate malignancy with cribriform architecture only \\
& high-grade prostatic malignancy with solid growth & high-grade prostatic malignancy with solid growth only \\
& \textit{(no corresponding entry)} & high-grade prostate adenocarcinoma with marked necrosis \emph{(added)} \\
\midrule

Benign vs.\ normal descriptors
& benign prostate tissue & normal prostate tissue \\
& histologically benign prostate core & histologically normal prostate core \\
& normal benign prostatic tissue & prostatic tissue within normal limits \\
& benign colorectal mucosa & colonic mucosa with no significant histological changes \\
& histologically negative lymph node & histologically normal lymph node \\
\midrule

Nodal involvement phrasing
& lymph node metastasis & lymph node with metastatic carcinoma \\
& lymph node isolated tumor cells & lymph node with isolated tumor cells \\
& lymph node micrometastasis & lymph node with micrometastasis \\
& lymph node macrometastasis & lymph node with macrometastasis \\
& CAM16 normal lymph node & normal lymph node \emph{(dataset prefix removed)} \\
\midrule

Architectural descriptors
& preserved gland formation & preserved glandular structure \\
& mild nuclear atypia but recognizable glandular structures & mild nuclear atypia with recognizable glandular structures \\
& mainly composed of recognizable glands & mainly composed of well-formed glands \\
& a secondary component of poorly formed or fused small glands & a lesser component of poorly formed or fused small glands \\
& a smaller component of well-formed glands & a lesser component of well-formed glands \\
& single carcinoma cells within nodal tissue & rare carcinoma cells within nodal tissue \\

\bottomrule
\end{tabularx}
\end{table*}

\begin{table*}[t]
\centering
\footnotesize
\caption{Entry-specific prompt revisions from the final revision round (May 3, 2026). Distractor candidates judged implausible or non-discriminative were replaced with diagnostically plausible alternatives.}
\label{tab:rev_c}
\begin{tabularx}{\textwidth}{p{0.20\textwidth} X X}
\toprule
\textbf{Revision type} & \textbf{Original} & \textbf{Revised} \\
\midrule

Generic $\rightarrow$ specific entity
& reactive tissue & tubular adenoma \\
& normal tissue & sessile serrated adenoma \\
& normal tissue & Barrett esophagus with low-grade dysplasia \\
& dysplasia & squamocolumnar mucosa with focal intestinal metaplasia \\
& benign tissue & squamocolumnar mucosa with focal intestinal metaplasia \\
& low-grade dysplasia & inflammatory polyp \\
\midrule

Lesion-specific mimics
& well-differentiated carcinoma & high-grade prostatic intraepithelial neoplasia (HGPIN) \\
& moderately differentiated carcinoma & high-grade prostatic intraepithelial neoplasia (HGPIN) \\
& invasive carcinoma & high-grade prostatic intraepithelial neoplasia (HGPIN) \\
& serrated adenoma & sessile serrated adenoma \\
\midrule

Benign mimics of nodal metastasis
& lymph node metastasis & lymph node with necrotizing lymphadenitis \\
& CAM16 normal lymph node & lymph node with necrotizing lymphadenitis \\
& lymph node metastasis & normal thyroid tissue \\
\midrule

Organ-appropriate distractors
& well-differentiated carcinoma & reactive gliosis \\
& poorly differentiated carcinoma & reactive gliosis \\
& CAM16 normal lymph node & reactive gliosis \\
& poorly differentiated carcinoma & well-differentiated carcinoma \\
\midrule

PANDA ISUP grade descriptions
& predominantly well-formed prostate adenocarcinoma & \dots{} with minor component of poorly formed glands \\
& \dots{} with fused and cribriform glands & \dots{} with fused and cribriform glands and a minor pattern of well-formed glands \\
& \dots{} and no lower-grade component & \dots{} and no lower-grade glandular component \\
& this aggressive malignancy shows solid sheets, cords, single-cell infiltration, or comedonecrosis & high-grade prostate malignancy shows solid sheets, cords, single-cell infiltration, or comedonecrosis \\

\bottomrule
\end{tabularx}
\end{table*}

\section{Additional Discussion}\label{sec:discussion}

Across nine pathology and biomedical VLMs and five public datasets spanning four organs, PathLang reveals a consistent pattern: the nine evaluated vision--language models are \emph{visually strong but linguistically fragile}. They inherit competitive visual representations from large-scale image--text pretraining, yet their language alignment remains brittle, dataset-dependent, and unable to support clinically valid diagnostic-language variation. Collectively, our results suggest that closed-set canonical-prompt evaluation substantially overestimates the language robustness of the models evaluated here and does not predict behavior under realistic clinical phrasing. Below we synthesize how the four task families together expose this fragility, propose a mechanistic interpretation, and discuss what it implies for the field.

\noindent\textbf{Closed-set canonical-prompt evaluation overestimates language capability.} Three independent lines of evidence converge on this point. 
First, under canonical prompts (T1), model leadership shifts across both datasets and metrics (Table~\ref{tab:task1}): KEEP leads on PANDA and TCGA-GBMLGG while Patho-CLIP leads on UniToPatho and CAMELYON16, and on CAMELYON17 CONCH attains the highest accuracy.
Second, under clinically equivalent paraphrases (T3), several models that perform competitively on canonical prompts collapse to zero prediction consistency and negative Prompt Stability Scores, with BiomedCLIP and QuiltNet on CAMELYON16 yielding consistency of exactly 0.000 (Table~\ref{tab:robustness_synonyms}). Third, under open-vocabulary candidate pools (T4), several models fall to or below chance on the organ-agnostic abstract candidates of Set 1, despite performing substantially better when organ-specific language is available (Figure~\ref{fig:openvocab}). Canonical closed-set evaluation therefore measures only a narrow slice of pathology VLM behavior and fails to predict robustness under clinically realistic language variation. Main-text results grant each model its most favorable aggregation setting under oracle selection; per-setting results and the query-independent baseline in Appendix A.1 show that the reported instability is not specific to that choice.


\textbf{This overestimation is most visible where language is the only variable.} 
In T1, alignment quality and inter-class separability are decoupled: on CAMELYON17, KEEP attains a competitive image--text alignment score yet exhibits the most negative similarity gap among the evaluated models, indicating semantic confusion under clinically similar distractor prompts (Figure~\ref{fig:alignment}). 
In T3 and T4, the underlying slides and labels are held fixed while the diagnostic-language condition varies. Because every condition is evaluated in the same image--text direction against the same slides, differences between conditions are attributable to the diagnostic language rather than to the modality gap that confounds cross-direction comparisons in T2~\citep{liang2022mind}. The T2 asymmetry between image-only and cross-modal retrieval is consistent with weaker semantic than morphological organization, but it is not on its own diagnostic of it.

\noindent\textbf{Models operate within narrow memorized linguistic manifolds.} The instability patterns observed in T3 and T4 suggest that many of the evaluated models operate within narrow memorized linguistic manifolds rather than learning clinically invariant semantic representations. In T3, length sensitivity exhibits a clear ``prompt sweet spot'' pattern that varies by model: CONCH performs well on short prompts on CAMELYON16 but collapses under longer descriptions (PSS = $-0.291$), while KEEP shows the opposite trend, and PathGen-CLIP improves with length on some datasets but degrades on others (Table~\ref{tab:robustness_length}). In T4, performance depends more on whether organ-specific diagnostic language is available than on the number of competing candidates: the abstract, organ-agnostic pool of Set 1 yields near-chance performance for several models, while the smaller cross-organ pool of Set 3 does not (Figure~\ref{fig:openvocab}). One plausible explanation is that the pretraining corpora of these models are dominated by templated captions and short class-label-style descriptions rather than the diverse clinical reporting language pathologists actually use; performance appears to depend heavily on lexical proximity to training-time caption distributions rather than on invariant clinical semantics. Prompt ensembling reduces but does not eliminate this fragility, indicating that ensemble averaging smooths the linguistic manifold but does not change the underlying representation.

\noindent\textbf{Implications for evaluation, training, and deployment.} In real pathology workflows, diagnoses are rarely queried using a single canonical phrasing; clinicians routinely vary terminology, reporting style, specificity, and differential-diagnosis structure across institutions and reporting systems. Three implications follow. \emph{For evaluation}, closed-set canonical-prompt benchmarks are insufficient predictors of clinical readiness; multi-prompt, open-vocabulary, and clinically validated evaluation protocols such as PathLang should become a standard companion to existing visual-axis benchmarks. \emph{For pretraining}, future pathology VLMs should target diversity in clinical text, including paraphrases drawn from real pathology reports across institutions, multiple specificity levels, and varied reporting styles, to avoid overfitting narrow linguistic manifolds. \emph{For deployment}, prompt ensembling is a useful but partial mitigation; it masks rather than resolves underlying alignment failures, and architectural or pretraining-level interventions are likely needed for clinically reliable systems. Taken together, our findings suggest that language robustness remains a major unresolved bottleneck for the models evaluated here and should become a first-class target for future pathology foundation model development.

\section{Result Details}

\subsection{Zero-shot Classification}
\label{app:zeroshot_results}

Table~\ref{tab:alignment_appendix} summarizes the alignment scores and similarity gaps for zero-shot classification across all models and pathology datasets. Overall, PathGen-CLIP and KEEP achieve the strongest alignment performance on most datasets, while Patho-CLIP obtains the largest similarity gaps in several settings, indicating better separation between matched and unmatched image--text pairs.

\begin{figure*}[t]
    \centering
    \includegraphics[width=\textwidth]{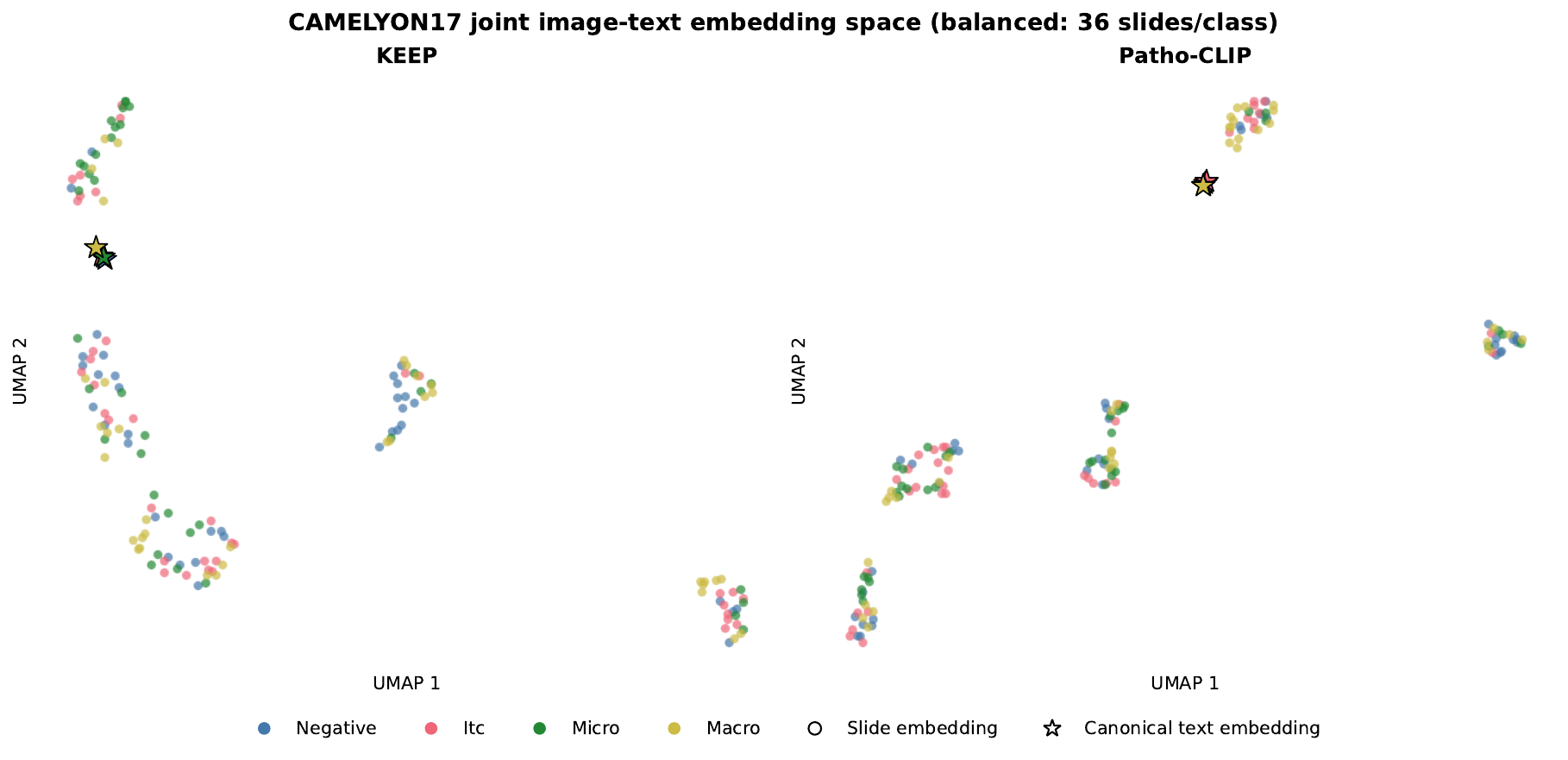}
    \caption{Qualitative UMAP visualization of the joint slide--text embedding
    space on CAMELYON17. Points are colored by ground-truth diagnostic class;
    circles denote slide embeddings and stars denote canonical text embeddings.
    For each model, 36 slides per class were sampled from the subset with
    verified labels. UMAP was fit independently for each model using
    L2-normalized embeddings, cosine distance, $n_{\mathrm{neighbors}}=15$,
    $\mathrm{min\_dist}=0.15$, and seed 42. Coordinates are not comparable across panels. This
    figure illustrates the modality gap between image and text embeddings~\citep{liang2022mind}; it is not evidence of class separability or of diagnostic-language grounding.}
    \label{fig:cam17-joint-umap}
\end{figure*}

Figure~\ref{fig:cam17-joint-umap} additionally visualizes the joint slide--text embedding space of KEEP and Patho-CLIP on CAMELYON17 using UMAP. In both models, canonical text embeddings occupy neighborhoods separated from most slide embeddings. We include this figure as a caveat rather than as supporting evidence. Separation between text and image embeddings is the expected signature of the modality gap, which arises from the cone effect at initialization and is preserved by the contrastive objective in vision--language models generally~\citep{liang2022mind}; it therefore cannot by itself distinguish weak diagnostic-language grounding from generic contrastive geometry. Because UMAP is additionally a nonlinear two-dimensional projection, we do not read distances or cluster shapes in this figure as measures of cross-modal alignment or class separability. Evidence for language-side fragility instead comes from T3 and T4, where alternative diagnostic-language conditions are compared against identical slides within a fixed image--text direction, holding the modality gap constant.

\begin{table*}[t]
\centering
\caption{Alignment score and similarity gap across five pathology datasets. Higher alignment scores indicate stronger image--text alignment, while similarity gap measures the separation between matched and unmatched pairs. The best result for each metric within each dataset is highlighted in bold.}
\label{tab:alignment_appendix}
\resizebox{\textwidth}{!}{
\begin{tabular}{l cc cc cc cc cc}
\toprule
\multirow{2}{*}{Models}
& \multicolumn{2}{c}{CAMELYON16}
& \multicolumn{2}{c}{CAMELYON17}
& \multicolumn{2}{c}{PANDA}
& \multicolumn{2}{c}{TCGA-GBMLGG}
& \multicolumn{2}{c}{UniToPatho} \\
\cmidrule(lr){2-3}
\cmidrule(lr){4-5}
\cmidrule(lr){6-7}
\cmidrule(lr){8-9}
\cmidrule(lr){10-11}
& Align. & Gap
& Align. & Gap
& Align. & Gap
& Align. & Gap
& Align. & Gap \\
\midrule

QuiltNet
& 0.2332 & -0.0028
& 0.2395 & -0.0167
& 0.2451 & -0.0014
& 0.2718 & 0.0002
& 0.2756 & 0.0068 \\

PLIP
& 0.2511 & 0.0011
& 0.2554 & 0.0058
& 0.2603 & 0.0043
& 0.2862 & -0.0003
& 0.2685 & 0.0046 \\

MI-Zero
& -0.0044 & -0.0022
& -0.0232 & -0.0055
& -0.0500 & 0.0002
& 0.0391 & -0.0022
& 0.0510 & 0.0011 \\

CONCH
& 0.0831 & -0.0074
& 0.0939 & -0.0066
& 0.0926 & -0.0058
& 0.0767 & -0.0067
& 0.0941 & -0.0069 \\

BiomedCLIP
& 0.0051 & 0.0014
& 0.0138 & 0.0071
& 0.0111 & 0.0000
& -0.0040 & 0.0000
& -0.0233 & -0.0055 \\

PathGen-CLIP
& \textbf{0.2891} & 0.0047
& \textbf{0.2653} & -0.0030
& 0.3023 & 0.0047
& 0.3145 & -0.0003
& 0.2968 & 0.0048 \\

MUSK
& 0.0129 & 0.0033
& -0.0083 & -0.0013
& 0.0221 & 0.0010
& 0.0047 & 0.0027
& -0.0199 & -0.0037 \\

Patho-CLIP
& 0.1662 & \textbf{0.0076}
& 0.2074 & \textbf{0.0127}
& 0.1609 & 0.0169
& 0.1692 & \textbf{0.0182}
& \textbf{0.3950} & \textbf{0.0456} \\

KEEP
& 0.2274 & 0.0065
& 0.2154 & -0.0284
& \textbf{0.3259} & \textbf{0.0543}
& \textbf{0.4855} & 0.0128
& 0.2892 & 0.0161 \\

\bottomrule
\end{tabular}
}
\end{table*}

\subsection{Cross-Modal Retrieval}
\label{app:cross_modal_results}

Tables~\ref{tab:crossmodal_I2I}-\ref{tab:crossmodal_T2I} present the full cross-modal retrieval results across five datasets and three retrieval directions. For Image-to-Image (I2I) retrieval, KEEP and Patho-CLIP consistently achieve strong performance, with KEEP leading on PANDA and UniToPatho, and Patho-CLIP performing best on CAMELYON16 and TCGA-GBMLGG. For Image-to-Text (I2T) retrieval, CONCH shows the strongest overall performance, particularly on PANDA and TCGA-GBMLGG, while Patho-CLIP leads on CAMELYON16. For Text-to-Image (T2I) retrieval, results are more variable across datasets, with PathGen-CLIP achieving perfect Recall@1 on CAMELYON16, and KEEP performing best on PANDA and UniToPatho. Across all three retrieval directions, no single model dominates consistently, suggesting that different models capture complementary aspects of pathology image-text alignment. Comparing across directions, the image-to-image advantage over cross-modal retrieval holds on average but not uniformly: on CAMELYON16 several models achieve higher text-to-image than image-to-image MRR, reflecting the small number of text queries in that direction (8--24 per dataset) rather than stronger language grounding.

\begin{table*}[t]
\centering
\caption{Image-to-Image Cross-modal retrieval performance across five pathology datasets. We report Recall@1 (R@1), Recall@5 (R@5), mean reciprocal rank (MRR), and nDCG. The best result for each metric within each dataset is highlighted in bold.}
\label{tab:crossmodal_I2I}
\resizebox{\textwidth}{!}{
\begin{tabular}{l cccc cccc cccc cccc cccc}
\toprule
\multirow{2}{*}{Models}
& \multicolumn{4}{c}{CAMELYON16}
& \multicolumn{4}{c}{CAMELYON17}
& \multicolumn{4}{c}{PANDA}
& \multicolumn{4}{c}{TCGA-GBMLGG}
& \multicolumn{4}{c}{UniToPatho} \\
\cmidrule(lr){2-5}
\cmidrule(lr){6-9}
\cmidrule(lr){10-13}
\cmidrule(lr){14-17}
\cmidrule(lr){18-21}
& R@1 & R@5 & MRR & nDCG
& R@1 & R@5 & MRR & nDCG
& R@1 & R@5 & MRR & nDCG
& R@1 & R@5 & MRR & nDCG
& R@1 & R@5 & MRR & nDCG \\
\midrule

QuiltNet
& 0.538 & 0.874 & 0.680 & 0.718
& 0.595 & 0.872 & 0.712 & 0.742
& 0.661 & 0.893 & 0.761 & 0.786
& 0.917 & 0.973 & 0.943 & 0.949
& 0.468 & 0.806 & 0.612 & 0.640 \\

PLIP
& 0.544 & 0.862 & 0.676 & 0.704
& 0.593 & 0.860 & 0.712 & 0.739
& 0.659 & 0.900 & 0.763 & 0.790
& 0.929 & \textbf{0.976} & 0.949 & 0.955
& 0.485 & 0.814 & 0.626 & 0.658 \\

MI-Zero
& 0.475 & 0.879 & 0.640 & 0.682
& 0.511 & 0.868 & 0.656 & 0.694
& 0.319 & 0.729 & 0.495 & 0.533
& 0.867 & 0.950 & 0.906 & 0.914
& 0.494 & 0.819 & 0.636 & 0.666 \\

CONCH
& 0.508 & 0.879 & 0.665 & 0.710
& \textbf{0.633} & 0.862 & \textbf{0.737} & \textbf{0.759}
& 0.707 & 0.913 & 0.796 & 0.819
& 0.921 & 0.973 & 0.943 & 0.949
& 0.544 & 0.844 & 0.668 & 0.696 \\

BiomedCLIP
& 0.491 & 0.882 & 0.653 & 0.696
& 0.505 & 0.856 & 0.662 & 0.701
& 0.332 & 0.737 & 0.508 & 0.545
& 0.846 & 0.952 & 0.894 & 0.905
& 0.447 & 0.823 & 0.607 & 0.648 \\

PathGen-CLIP
& 0.541 & 0.887 & 0.687 & 0.728
& 0.591 & \textbf{0.876} & 0.719 & 0.750
& 0.701 & 0.915 & 0.793 & 0.818
& 0.921 & 0.974 & 0.945 & 0.951
& 0.515 & 0.835 & 0.645 & 0.673 \\

MUSK
& 0.516 & 0.882 & 0.674 & 0.714
& 0.595 & 0.860 & 0.716 & 0.742
& 0.700 & \textbf{0.917} & 0.793 & 0.819
& 0.927 & 0.970 & 0.948 & 0.950
& 0.481 & 0.823 & 0.636 & 0.668 \\
Patho-CLIP
& \textbf{0.559} & 0.887 & \textbf{0.700} & \textbf{0.738}
& 0.581 & 0.856 & 0.702 & 0.731
& 0.689 & 0.906 & 0.783 & 0.807
& \textbf{0.930} & 0.974 & \textbf{0.951} & \textbf{0.955}
& 0.498 & 0.814 & 0.636 & 0.663 \\
KEEP
& 0.555 & \textbf{0.902} & 0.693 & 0.737
& 0.599 & 0.854 & 0.713 & 0.737
& \textbf{0.718} & \textbf{0.918} & \textbf{0.805} & \textbf{0.827}
& 0.919 & 0.973 & 0.943 & 0.948
& \textbf{0.557} & \textbf{0.861} & \textbf{0.681} & \textbf{0.716} \\

\bottomrule
\end{tabular}
}
\end{table*}

\begin{table*}[t]
\centering
\caption{Image-to-Text Cross-modal retrieval performance across five pathology datasets. We report Recall@1 (R@1), mean reciprocal rank (MRR), and nDCG. The best result for each metric within each dataset is highlighted in bold.}
\label{tab:crossmodal_I2T}
\resizebox{\textwidth}{!}{
\begin{tabular}{l ccc ccc ccc ccc ccc}
\toprule
\multirow{2}{*}{Models}
& \multicolumn{3}{c}{CAMELYON16}
& \multicolumn{3}{c}{CAMELYON17}
& \multicolumn{3}{c}{PANDA}
& \multicolumn{3}{c}{TCGA-GBMLGG}
& \multicolumn{3}{c}{UniToPatho} \\
\cmidrule(lr){2-4}
\cmidrule(lr){5-7}
\cmidrule(lr){8-10}
\cmidrule(lr){11-13}
\cmidrule(lr){14-16}
& R@1 & MRR & nDCG
& R@1 & MRR & nDCG
& R@1 & MRR & nDCG
& R@1 & MRR & nDCG
& R@1 & MRR & nDCG \\
\midrule

QuiltNet
& 0.500 & 0.728 & 0.798
& \textbf{0.397} & \textbf{0.618} & \textbf{0.713}
& 0.186 & 0.421 & 0.559
& 0.467 & 0.699 & 0.775
& 0.194 & 0.430 & 0.567 \\

PLIP
& 0.519 & 0.743 & 0.810
& 0.196 & 0.456 & 0.590
& 0.157 & 0.380 & 0.527
& 0.492 & 0.745 & 0.812
& 0.148 & 0.384 & 0.530 \\

MI-Zero
& 0.563 & 0.781 & 0.838
& 0.192 & 0.385 & 0.527
& 0.194 & 0.404 & 0.543
& 0.501 & 0.731 & 0.801
& 0.122 & 0.354 & 0.506 \\

CONCH
& 0.608 & 0.752 & 0.814
& 0.285 & 0.532 & 0.648
& \textbf{0.260} & \textbf{0.446} & \textbf{0.576}
& \textbf{0.563} & \textbf{0.775} & \textbf{0.834}
& 0.228 & 0.459 & 0.587 \\

BiomedCLIP
& 0.509 & 0.752 & 0.817
& 0.303 & 0.568 & 0.676
& 0.114 & 0.322 & 0.479
& 0.524 & 0.762 & 0.824
& 0.198 & 0.448 & 0.582 \\

PathGen-CLIP
& 0.406 & 0.680 & 0.762
& 0.393 & 0.612 & 0.709
& 0.151 & 0.389 & 0.535
& 0.507 & 0.752 & 0.817
& 0.194 & 0.427 & 0.565 \\

MUSK
& 0.546 & 0.773 & 0.833
& 0.160 & 0.413 & 0.555
& 0.204 & 0.421 & 0.558
& 0.484 & 0.729 & 0.799
& 0.236 & 0.463 & 0.590 \\

Patho-CLIP
& \textbf{0.609} & \textbf{0.804} & \textbf{0.855}
& 0.393 & 0.612 & 0.709
& 0.165 & 0.395 & 0.538
& 0.516 & 0.756 & 0.820
& 0.177 & 0.402 & 0.545 \\

KEEP
& 0.601 & 0.772 & 0.830
& 0.293 & 0.524 & 0.641
& 0.173 & 0.432 & 0.570
& 0.508 & 0.751 & 0.816
& \textbf{0.262} & \textbf{0.478} & \textbf{0.604} \\

\bottomrule
\end{tabular}
}
\end{table*}

\begin{table*}[t]
\centering
\caption{Text-to-Image Cross-modal Retrieval performance across five pathology datasets. We report Recall@1 (R@1), Recall@5 (R@5), mean reciprocal rank (MRR), and nDCG. The best result for each metric within each dataset is highlighted in bold.}
\label{tab:crossmodal_T2I}
\resizebox{\textwidth}{!}{
\begin{tabular}{l cccc cccc cccc cccc cccc}
\toprule
\multirow{2}{*}{Models}
& \multicolumn{4}{c}{CAMELYON16}
& \multicolumn{4}{c}{CAMELYON17}
& \multicolumn{4}{c}{PANDA}
& \multicolumn{4}{c}{TCGA-GBMLGG}
& \multicolumn{4}{c}{UniToPatho} \\
\cmidrule(lr){2-5}
\cmidrule(lr){6-9}
\cmidrule(lr){10-13}
\cmidrule(lr){14-17}
\cmidrule(lr){18-21}
& R@1 & R@5 & MRR & nDCG
& R@1 & R@5 & MRR & nDCG
& R@1 & R@5 & MRR & nDCG
& R@1 & R@5 & MRR & nDCG
& R@1 & R@5 & MRR & nDCG \\
\midrule

QuiltNet
& 0.625 & \textbf{1.000} & 0.781 & 0.803
& \textbf{0.438} & \textbf{0.813} & \textbf{0.533} & \textbf{0.601}
& 0.250 & 0.542 & 0.326 & 0.384
& \textbf{0.625} & 0.750 & 0.688 & 0.695
& 0.250 & 0.667 & 0.388 & 0.460 \\

PLIP
& 0.625 & 0.875 & 0.719 & 0.752
& 0.250 & 0.625 & 0.358 & 0.429
& 0.292 & 0.625 & 0.422 & 0.456
& 0.500 & 0.750 & 0.625 & 0.676
& 0.333 & 0.750 & 0.481 & 0.532 \\

MI-Zero
& 0.500 & 0.500 & 0.500 & 0.500
& 0.250 & 0.625 & 0.385 & 0.438
& 0.208 & 0.500 & 0.291 & 0.347
& 0.500 & 0.500 & 0.500 & 0.500
& 0.208 & 0.458 & 0.300 & 0.330 \\

CONCH
& 0.125 & 0.875 & 0.406 & 0.530
& 0.250 & 0.563 & 0.328 & 0.391
& 0.250 & 0.708 & 0.376 & 0.462
& \textbf{0.625} & \textbf{1.000} & \textbf{0.760} & \textbf{0.813}
& 0.167 & 0.542 & 0.283 & 0.328 \\

BiomedCLIP
& 0.750 & \textbf{1.000} & 0.800 & 0.828
& 0.188 & 0.688 & 0.359 & 0.443
& 0.292 & 0.542 & 0.375 & 0.386
& 0.500 & 0.875 & 0.598 & 0.665
& 0.208 & 0.250 & 0.208 & 0.205 \\

PathGen-CLIP
& \textbf{1.000} & \textbf{1.000} & \textbf{1.000} & \textbf{0.976}
& 0.313 & 0.750 & 0.495 & 0.566
& 0.417 & 0.750 & 0.501 & 0.533
& 0.500 & 0.625 & 0.542 & 0.558
& 0.333 & \textbf{0.917} & 0.517 & 0.593 \\

MUSK
& 0.500 & \textbf{1.000} & 0.729 & 0.809
& 0.313 & 0.625 & 0.438 & 0.472
& 0.292 & 0.708 & 0.404 & 0.451
& \textbf{0.625} & 0.750 & 0.650 & 0.668
& 0.292 & 0.542 & 0.345 & 0.379 \\
Patho-CLIP
& 0.875 & \textbf{1.000} & 0.938 & 0.907
& 0.250 & 0.750 & 0.396 & 0.475
& 0.333 & 0.708 & 0.442 & 0.489
& 0.500 & 0.750 & 0.604 & 0.641
& 0.458 & 0.708 & 0.547 & 0.572 \\
KEEP
& 0.625 & \textbf{1.000} & 0.813 & 0.865
& 0.375 & \textbf{0.813} & \textbf{0.533} & 0.591
& \textbf{0.500} & \textbf{0.833} & \textbf{0.540} & \textbf{0.552}
& 0.500 & 0.750 & 0.563 & 0.608
& \textbf{0.500} & 0.833 & \textbf{0.579} & \textbf{0.634} \\

\bottomrule
\end{tabular}
}
\end{table*}

\subsection{Paraphrase Robustness}
\label{app:appendix_paraphrase}
\begin{table*}[t]
\centering
\caption{Accuracy (Acc), Prediction Consistency (P.C.) and Prompt Stability Score (PSS) under semantic-equivalence paraphrases (T3 Variant~1). Results are reported under the upper-bound aggregation protocol (Section~\ref{sec:topk_method}); per-setting results and rank-stability analysis are in \ref{app:k-sensitive}.}
\label{tab:robustness_synonyms}
\resizebox{\textwidth}{!}{%
\begin{tabular}{l *{15}{c}} 
\toprule
\multirow{2}{*}{Model} 
& \multicolumn{3}{c}{\textbf{CAMELYON16}} & \multicolumn{3}{c}{\textbf{CAMELYON17}} & \multicolumn{3}{c}{\textbf{PANDA}} & \multicolumn{3}{c}{\textbf{TCGA-GBMLGG}} & \multicolumn{3}{c}{\textbf{UniToPatho}} \\
\cmidrule(lr){2-4} \cmidrule(lr){5-7} \cmidrule(lr){8-10} \cmidrule(lr){11-13} \cmidrule(lr){14-16}
& Acc. & P.C. & PSS 
& Acc. & P.C. & PSS 
& Acc. & P.C. & PSS 
& Acc. & P.C. & PSS 
& Acc. & P.C. & PSS \\
\midrule
QuiltNet \citep{quilt1m}
& 0.511 & 0.000 & -0.140 
& \underline{0.275} & 0.000 & -0.126 
& 0.179 & 0.013 & -0.073 
& 0.499 & 0.012 & -0.181 
& 0.236 & 0.000 & -0.048 \\
PLIP \citep{plip}
& 0.513 & 0.010 & -0.073 
& 0.236 & 0.002 & -0.068 
& 0.198 & 0.000 & -0.055 
& 0.536 & 0.196 & 0.080 
& 0.196 & 0.000 & -0.087 \\
MI-Zero \citep{mizero}
& 0.497 & 0.000 & -0.111 
& 0.255 & \textbf{0.341} & \textbf{0.381} 
& 0.166 & 0.000 & -0.144 
& 0.513 & 0.000 & -0.142 
& 0.178 & 0.000 & -0.105 \\
CONCH \citep{conch}
& 0.498 & \underline{0.724} & -0.026 
& 0.263 & 0.000 & -0.139 
& \underline{0.198} & 0.000 & -0.082 
& 0.513 & 0.011 & -0.149 
& 0.160 & 0.017 & -0.072 \\
BiomedCLIP \citep{biomedclip}
& 0.495 & 0.000 & -0.184 
& 0.246 & 0.000 & -0.211 
& 0.165 & 0.000 & -0.137 
& 0.492 & 0.025 & -0.115 
& 0.174 & 0.000 & 0.058 \\

PathGen-CLIP \citep{pathgenclip}
& \underline{0.514} & 0.108 & -0.139 
& 0.269 & 0.000 & -0.117 
& 0.197 & \underline{0.026} & -0.066 
& 0.564 & \textbf{0.293} & \textbf{0.171} 
& \textbf{0.310} & \underline{0.173} & \underline{0.164} \\
MUSK \citep{musk}
& 0.510 & 0.381 & \underline{0.055} 
& 0.250 & 0.000 & -0.124 
& 0.195 & 0.003 & \underline{0.067} 
& 0.507 & 0.016 & -0.110 
& 0.186 & \textbf{0.435} & \textbf{0.403} \\

Patho-CLIP \citep{pathoclip}
& 0.512 & 0.005 & -0.170 
& 0.271 & \underline{0.044} & -0.050 
& 0.183 & 0.019 & -0.072 
& \underline{0.567} & \underline{0.247} & 0.004 
& 0.275 & 0.059 & -0.004 \\

KEEP \citep{keep}
& \textbf{0.533} & \textbf{0.781} & \textbf{0.225} 
& \textbf{0.306} & 0.000 & \underline{-0.038} 
& \textbf{0.234} & \textbf{0.481} & \textbf{0.263} 
& \textbf{0.612} & 0.077 & \underline{0.114} 
& \underline{0.283} & 0.122 & 0.109 \\
\bottomrule
\end{tabular}
}
\end{table*}


\begin{table*}[t]
\centering
\small
\caption{Accuracy (Acc), Prediction Consistency (P.C.) and Prompt Stability Score (PSS) under length and reporting-style variation (T3 Variant~2). Results are reported under the upper-bound aggregation protocol (Section~\ref{sec:topk_method}); per-setting results and rank-stability analysis are in \ref{app:k-sensitive}.}
\label{tab:robustness_length}
\setlength{\tabcolsep}{3pt}
\begin{adjustbox}{width=\textwidth,center}
\begin{tabular}{l *{15}{c}}
\toprule
\multirow{2}{*}{Models} 
& \multicolumn{3}{c}{\textbf{CAMELYON16}} & \multicolumn{3}{c}{\textbf{CAMELYON17}} & \multicolumn{3}{c}{\textbf{PANDA}} & \multicolumn{3}{c}{\textbf{TCGA-GBMLGG}} & \multicolumn{3}{c}{\textbf{UniToPatho}} \\
\cmidrule(lr){2-4} \cmidrule(lr){5-7} \cmidrule(lr){8-10} \cmidrule(lr){11-13} \cmidrule(lr){14-16}
& Acc. & P.C. & PSS 
& Acc. & P.C. & PSS 
& Acc. & P.C. & PSS
& Acc. & P.C. & PSS
& Acc. & P.C. & PSS \\
\midrule
QuiltNet \citep{quilt1m}
& 0.500 & 0.000 & -0.333 
& 0.252 & 0.000 & -0.298 
& 0.175 & 0.000 & -0.202 
& 0.490 & 0.180 & -0.079 
& 0.180 & 0.013 & -0.099 \\
PLIP \citep{plip}
& \underline{0.518} & 0.456 & -0.022 
& \textbf{0.274} & \underline{0.170} & \underline{0.101} 
& 0.182 & \underline{0.223} & -0.064 
& 0.497 & 0.043 & -0.100 
& 0.239 & 0.000 & -0.100 \\
MI-Zero \citep{mizero}
& 0.500 & 0.025 & -0.275 
& 0.251 & \textbf{0.353} & \textbf{0.112} 
& 0.160 & 0.000 & -0.124 
& 0.460 & 0.166 & -0.122 
& 0.202 & 0.000 & -0.270 \\
CONCH \citep{conch}
& 0.495 & 0.003 & -0.291 
& 0.270 & 0.000 & -0.256 
& 0.190 & 0.001 & -0.111 
& 0.563 & \textbf{0.368} & \underline{0.042} 
& 0.167 & 0.000 & -0.094 \\

BiomedCLIP \citep{biomedclip}
& 0.507 & 0.000 & -0.259 
& 0.247 & 0.000 & -0.312 
& 0.169 & 0.000 & -0.261 
& 0.505 & 0.000 & -0.266 
& 0.177 & \textbf{0.852} & \underline{0.143} \\
PathGen-CLIP \citep{pathgenclip}
& 0.515 & \textbf{0.967} & \underline{0.273} 
& 0.239 & 0.004 & -0.178 
& 0.172 & 0.068 & -0.112 
& \underline{0.578} & \underline{0.210} & \textbf{0.115} 
& \textbf{0.325} & 0.055 & 0.011 \\
MUSK \citep{musk}
& 0.513 & 0.318 & -0.115 
& 0.254 & 0.008 & -0.161 
& \underline{0.192} & 0.172 & \underline{0.052} 
& 0.529 & 0.121 & 0.034 
& 0.191 & \underline{0.477} & \textbf{0.144} \\
Patho-CLIP \citep{pathoclip}
& 0.507 & 0.023 & -0.308 
& \underline{0.272} & 0.038 & -0.124 
& 0.171 & 0.140 & -0.034 
& 0.569 & 0.100 & -0.100 
& 0.257 & 0.317 & 0.070 \\

KEEP \citep{keep}
& \textbf{0.536} & \underline{0.925} & \textbf{0.366} 
& 0.266 & 0.000 & -0.186 
& \textbf{0.247} & \textbf{0.269} & \textbf{0.132} 
& \textbf{0.592} & 0.118 & 0.012 
& \underline{0.275} & 0.004 & -0.116 \\
\bottomrule
\end{tabular}
\end{adjustbox}
\end{table*}

Table~\ref{tab:robustness_synonyms} reports per-dataset accuracy, prediction consistency, and PSS under semantic-equivalence paraphrases, and Table~\ref{tab:robustness_length} reports the same quantities under length and reporting-style variation. Tables~\ref{tab:appendix_paraphrase_synonyms} and \ref{tab:appendix_prompt_length} give the full per-variant accuracies. 
The patterns discussed in Section~\ref{sec:experiment_task3} hold across all five datasets: no model achieves stable predictions on every dataset, and stability is dataset-dependent rather than model-intrinsic.

Length sensitivity exhibits a model-specific dependence on prompt length: CONCH performs well on short prompts but degrades sharply with verbosity (PSS = $-0.291$ on CAMELYON16), KEEP shows the opposite trend, and PathGen-CLIP improves with length on some datasets but degrades on others. Prompt ensembling partially smooths prediction instability and improves average accuracy as ensemble size grows toward nine, but it fails to restore stable semantic agreement across prompt variants for most models.

Slide-level bootstrap confidence intervals for Prediction Consistency and the Prompt Stability Score are reported in Table~\ref{tab:t3_bootstrap_ci_v1} (Variant~1) and Table~\ref{tab:t3_bootstrap_ci_v2} (Variant~2). Intervals are 2.5th--97.5th percentiles over 1{,}000 bootstrap resamples of slides with replacement (seed 42), recomputing both quantities on each resample. Across the 90 model--dataset--variant combinations, 85 PSS intervals exclude zero, indicating that the observed departures from chance-level inter-variant agreement are not attributable to slide sampling variability. The five intervals that contain zero (Patho-CLIP on TCGA-GBMLGG and UniToPatho under Variant~1; PLIP on CAMELYON16, PathGen-CLIP on UniToPatho, and KEEP on TCGA-GBMLGG under 
Variant~2) correspond to cases where predictions are statistically indistinguishable from chance-level agreement rather than to reliably stable behavior.


\begin{table*}[t]
\centering
\caption{Per-variant accuracy under semantic-equivalence paraphrases (T3 Variant~1). We report accuracy under five semantically equivalent prompt variants, Acc-v1 to Acc-v5. The best result for each prompt variant within each dataset is highlighted in bold.}
\label{tab:appendix_paraphrase_synonyms}
\resizebox{\textwidth}{!}{
\begin{tabular}{l ccccc ccccc ccccc ccccc ccccc}
\toprule
\multirow{2}{*}{Models}
& \multicolumn{5}{c}{CAMELYON16}
& \multicolumn{5}{c}{CAMELYON17}
& \multicolumn{5}{c}{PANDA}
& \multicolumn{5}{c}{TCGA-GBMLGG}
& \multicolumn{5}{c}{UniToPatho} \\
\cmidrule(lr){2-6}
\cmidrule(lr){7-11}
\cmidrule(lr){12-16}
\cmidrule(lr){17-21}
\cmidrule(lr){22-26}
& v1 & v2 & v3 & v4 & v5
& v1 & v2 & v3 & v4 & v5
& v1 & v2 & v3 & v4 & v5
& v1 & v2 & v3 & v4 & v5
& v1 & v2 & v3 & v4 & v5 \\
\midrule

QuiltNet
& 0.402 & 0.440 & 0.598 & 0.585 & 0.402
& 0.110 & 0.148 & 0.072 & 0.497 & 0.122
& 0.199 & 0.180 & 0.276 & 0.272 & 0.141
& 0.664 & \textbf{0.664} & 0.325 & \textbf{0.658} & 0.492
& 0.135 & 0.384 & 0.169 & 0.110 & 0.498 \\

PLIP
& 0.619 & 0.406 & 0.521 & 0.579 & 0.486
& 0.345 & 0.188 & 0.186 & 0.383 & 0.150
& 0.279 & 0.254 & \textbf{0.301} & 0.295 & 0.267
& 0.445 & 0.521 & 0.525 & 0.542 & 0.532
& 0.430 & 0.494 & \textbf{0.502} & \textbf{0.477} & 0.144 \\

MI-Zero
& 0.450 & 0.546 & 0.558 & 0.402 & 0.518
& 0.116 & 0.313 & 0.194 & 0.399 & 0.116
& 0.255 & 0.125 & 0.124 & 0.115 & 0.194
& 0.561 & 0.437 & 0.621 & 0.412 & 0.515
& 0.144 & 0.397 & 0.173 & 0.008 & 0.046 \\

CONCH
& 0.420 & 0.402 & 0.397 & 0.440 & 0.402
& 0.142 & 0.090 & 0.122 & 0.485 & 0.505
& 0.132 & 0.157 & 0.278 & 0.117 & 0.135
& 0.547 & 0.454 & 0.476 & 0.509 & 0.535
& 0.084 & 0.131 & 0.089 & 0.190 & 0.089 \\

BiomedCLIP
& 0.426 & 0.599 & 0.514 & 0.414 & 0.599
& 0.118 & \textbf{0.635} & 0.635 & 0.064 & 0.134
& 0.135 & 0.105 & 0.273 & 0.112 & 0.251
& 0.561 & 0.477 & 0.444 & 0.561 & 0.502
& 0.131 & 0.169 & 0.317 & 0.118 & 0.181 \\

PathGen-CLIP
& 0.607 & 0.421 & 0.594 & \textbf{0.609} & 0.602
& 0.265 & 0.585 & 0.106 & 0.140 & \textbf{0.579}
& 0.268 & \textbf{0.274} & 0.287 & \textbf{0.328} & 0.169
& 0.514 & 0.552 & 0.549 & 0.583 & 0.514
& 0.274 & 0.414 & 0.477 & 0.430 & \textbf{0.523} \\

MUSK
& 0.607 & 0.589 & 0.556 & 0.574 & 0.539
& 0.311 & 0.154 & 0.635 & 0.487 & 0.100
& 0.242 & 0.236 & 0.171 & 0.241 & 0.215
& 0.578 & 0.528 & 0.545 & 0.521 & 0.448
& 0.097 & 0.084 & 0.131 & 0.106 & 0.106 \\

Patho-CLIP
& 0.573 & 0.399 & 0.397 & 0.606 & \textbf{0.616}
& 0.198 & 0.271 & 0.076 & 0.289 & 0.130
& 0.282 & 0.272 & 0.235 & 0.300 & 0.217
& \textbf{0.695} & 0.481 & 0.483 & 0.470 & 0.548
& 0.076 & \textbf{0.506} & 0.160 & 0.144 & 0.245 \\

KEEP
& \textbf{0.628} & \textbf{0.626} & \textbf{0.598} & 0.608 & 0.613
& \textbf{0.627} & 0.607 & \textbf{0.671} & \textbf{0.571} & 0.120
& \textbf{0.352} & 0.306 & 0.318 & 0.355 & \textbf{0.359}
& 0.660 & 0.509 & 0.608 & 0.604 & \textbf{0.585}
& \textbf{0.532} & 0.502 & 0.435 & 0.274 & 0.165 \\

\bottomrule
\end{tabular}
}
\end{table*}

\begin{table*}[t]
\centering
\caption{Per-variant accuracy under length and reporting-style variation (T3 Variant~2) across five pathology datasets. We report accuracy under four prompt styles: Short, Medium, Long, and Clinical. The best result for each prompt style within each dataset is highlighted in bold.}
\label{tab:appendix_prompt_length}
\resizebox{\textwidth}{!}{
\begin{tabular}{l cccc cccc cccc cccc cccc}
\toprule
\multirow{2}{*}{Models}
& \multicolumn{4}{c}{CAMELYON16}
& \multicolumn{4}{c}{CAMELYON17}
& \multicolumn{4}{c}{PANDA}
& \multicolumn{4}{c}{TCGA-GBMLGG}
& \multicolumn{4}{c}{UniToPatho} \\
\cmidrule(lr){2-5}
\cmidrule(lr){6-9}
\cmidrule(lr){10-13}
\cmidrule(lr){14-17}
\cmidrule(lr){18-21}
& Short & Medium & Long & Clinical
& Short & Medium & Long & Clinical
& Short & Medium & Long & Clinical
& Short & Medium & Long & Clinical
& Short & Medium & Long & Clinical \\
\midrule

QuiltNet
& 0.402 & 0.402 & 0.598 & 0.402
& 0.074 & 0.116 & 0.170 & 0.072
& 0.115 & 0.164 & 0.190 & 0.119
& 0.665 & 0.459 & 0.577 & 0.577
& 0.152 & 0.110 & 0.283 & 0.160 \\

PLIP
& 0.609 & 0.599 & 0.516 & 0.602
& 0.375 & 0.184 & 0.086 & 0.323
& 0.272 & 0.268 & 0.251 & 0.258
& 0.538 & 0.534 & 0.436 & 0.475
& 0.490 & 0.122 & 0.485 & 0.346 \\

MI-Zero
& 0.452 & 0.548 & 0.414 & 0.407
& 0.419 & 0.072 & 0.118 & 0.118
& 0.242 & 0.124 & 0.160 & 0.155
& 0.568 & 0.440 & 0.345 & 0.570
& 0.000 & 0.173 & 0.152 & \textbf{0.549} \\

CONCH
& 0.598 & 0.405 & 0.400 & 0.558
& 0.118 & 0.172 & \textbf{0.673} & 0.455
& 0.229 & 0.260 & 0.118 & 0.231
& 0.463 & 0.569 & 0.504 & 0.591
& 0.089 & 0.101 & 0.144 & 0.409 \\

BiomedCLIP
& 0.599 & 0.579 & 0.399 & 0.599
& 0.619 & \textbf{0.635} & 0.072 & 0.619
& 0.272 & 0.126 & 0.148 & 0.120
& 0.561 & 0.561 & 0.439 & 0.556
& 0.511 & 0.485 & 0.473 & 0.485 \\

PathGen-CLIP
& 0.609 & \textbf{0.632} & 0.599 & 0.604
& 0.144 & 0.479 & 0.078 & 0.080
& 0.200 & \textbf{0.270} & 0.276 & 0.257
& \textbf{0.676} & 0.609 & 0.572 & 0.484
& 0.498 & 0.194 & 0.190 & 0.177 \\

MUSK
& 0.509 & 0.566 & \textbf{0.612} & 0.599
& \textbf{0.627} & 0.180 & 0.106 & 0.275
& 0.263 & 0.218 & 0.195 & 0.253
& 0.543 & 0.479 & 0.525 & 0.579
& 0.144 & 0.106 & 0.110 & 0.080 \\

Patho-CLIP
& 0.414 & 0.611 & 0.604 & 0.604
& 0.445 & 0.122 & 0.098 & 0.467
& 0.256 & 0.134 & 0.227 & 0.266
& 0.513 & \textbf{0.627} & 0.455 & 0.604
& \textbf{0.506} & 0.295 & 0.494 & 0.536 \\

KEEP
& \textbf{0.656} & 0.608 & 0.606 & \textbf{0.636}
& 0.164 & \textbf{0.635} & 0.239 & 0.076
& \textbf{0.304} & 0.262 & \textbf{0.325} & \textbf{0.329}
& 0.644 & 0.553 & 0.512 & \textbf{0.613}
& 0.498 & 0.274 & \textbf{0.536} & 0.076 \\

\bottomrule
\end{tabular}
}
\end{table*}

\begin{table*}[t]
\centering
\small
\setlength{\tabcolsep}{2.5pt}
\renewcommand{\arraystretch}{1.1}
\caption{Variability of mean accuracy under random prompt sampling. For each ensemble size $n \in \{1,\ldots,9\}$, $n$ prompts are drawn without replacement from the pool of nine variants and their embeddings averaged; this is repeated 100 times (seed 42). We report the 2.5th and 97.5th percentiles (CI95-L, CI95-U) and the interquartile range of accuracy across these draws. These intervals describe sensitivity to which prompts are sampled, rather than statistical uncertainty in the underlying accuracy estimate. For each dataset, the smallest interval width (CI95-L $-$ CI95-U) is shown in bold.}
\label{tab:ci_iqr_all_datasets}
\resizebox{\textwidth}{!}{
\begin{tabular}{lccccccccccccccc}
\toprule
Model 
& \multicolumn{3}{c}{CAMELYON16} 
& \multicolumn{3}{c}{CAMELYON17} 
& \multicolumn{3}{c}{PANDA} 
& \multicolumn{3}{c}{TCGA-GBMLGG} 
& \multicolumn{3}{c}{UniToPatho} \\
\cmidrule(lr){2-4} 
\cmidrule(lr){5-7} 
\cmidrule(lr){8-10} 
\cmidrule(lr){11-13} 
\cmidrule(lr){14-16}
& CI95-L & CI95-U & IQR 
& CI95-L & CI95-U & IQR 
& CI95-L & CI95-U & IQR 
& CI95-L & CI95-U & IQR 
& CI95-L & CI95-U & IQR \\
\midrule

QuiltNet \citep{quilt1m} & 0.3999 & 0.5346 & 0.0413 & 0.0708 & 0.2013 & 0.0359 & 0.1363 & 0.2572 & 0.0616 & 0.5509 & 0.6814 & 0.0482 & 0.1097 & 0.4412 & 0.1298 \\

PLIP \citep{plip} & 0.3951 & 0.6096 & 0.1386 & 0.0913 & 0.3478 & 0.1288 & 0.2608 & 0.3084 & 0.0124 & 0.5274 & 0.6196 & 0.0382 & 0.2096 & 0.5038 & 0.1784 \\

MI-Zero \citep{mizero} & 0.4029 & 0.5349 & 0.0682 & \textbf{0.1047} & \textbf{0.2350} & \textbf{0.0253} & 0.1180 & 0.2219 & 0.0373 & 0.4108 & 0.6087 & 0.1056 & 0.0145 & 0.2152 & 0.0982 \\

CONCH \citep{conch} & 0.3959 & 0.4756 & 0.0102 & 0.1041 & 0.5812 & 0.2186 & 0.1168 & 0.2455 & 0.0782 & 0.4619 & 0.5767 & 0.0586 & 0.0574 & 0.3534 & 0.0982 \\

BiomedCLIP \citep{biomedclip} & 0.4593 & 0.6016 & 0.0823 & 0.2491 & 0.6353 & 0.2050 & 0.1165 & 0.2416 & 0.0586 & 0.4943 & 0.5614 & 0.0171 & 0.1305 & 0.4441 & 0.1924 \\

PathGen-CLIP \citep{pathgenclip} & 0.5512 & 0.6365 & 0.0098 & 0.0746 & 0.2600 & 0.0813 & 0.2558 & 0.3203 & 0.0177 & 0.5567 & 0.6651 & 0.0430 & 0.1718 & 0.4462 & 0.1511 \\

MUSK \citep{musk} & 0.5639 & 0.6186 & 0.0162 & 0.2469 & 0.6235 & 0.1798 & \textbf{0.2122} & \textbf{0.2566} & \textbf{0.0132} & \textbf{0.4996} & \textbf{0.5647} & \textbf{0.0296} & \textbf{0.0818} & \textbf{0.1163} & \textbf{0.0165} \\

Patho-CLIP \citep{pathoclip} & 0.5019 & 0.6308 & 0.0272 & 0.0723 & 0.2896 & 0.0577 & 0.2313 & 0.2841 & 0.0153 & 0.4532 & 0.6152 & 0.0851 & 0.0961 & 0.3729 & 0.1596 \\

KEEP \citep{keep} & \textbf{0.6072} & \textbf{0.6274} & \textbf{0.0061} & 0.2104 & 0.6322 & 0.2335 & 0.3126 & 0.3572 & 0.0198 & 0.5697 & 0.6663 & 0.0339 & 0.1966 & 0.4724 & 0.1221 \\
\bottomrule
\end{tabular}
}
\end{table*}

\begin{table*}[t]
\centering
\caption{Slide-level bootstrap confidence intervals for Prediction Consistency (P.C.) and Prompt Stability Score (PSS) under semantic-equivalence paraphrases (T3 Variant~1). Intervals are 2.5th--97.5th percentiles over 1{,}000 bootstrap resamples of slides drawn with replacement (seed 42). Point estimates are those reported in Table~\ref{tab:robustness_synonyms}. Intervals shown in \textit{italics} contain zero.}
\label{tab:t3_bootstrap_ci_v1}
\resizebox{\textwidth}{!}{%
\begin{tabular}{l cc cc cc cc cc}
\toprule
\multirow{2}{*}{Model}
& \multicolumn{2}{c}{\textbf{CAMELYON16}} & \multicolumn{2}{c}{\textbf{CAMELYON17}}
& \multicolumn{2}{c}{\textbf{PANDA}} & \multicolumn{2}{c}{\textbf{TCGA-GBMLGG}}
& \multicolumn{2}{c}{\textbf{UniToPatho}} \\
\cmidrule(lr){2-3}\cmidrule(lr){4-5}\cmidrule(lr){6-7}\cmidrule(lr){8-9}\cmidrule(lr){10-11}
& P.C. & PSS & P.C. & PSS & P.C. & PSS & P.C. & PSS & P.C. & PSS \\
\midrule
QuiltNet & [0.000, 0.000] & [-0.151, -0.131] & [0.000, 0.000] & [-0.134, -0.118] & [0.011, 0.015] & [-0.076, -0.069] & [0.006, 0.019] & [-0.187, -0.174] & [0.000, 0.000] & [-0.070, -0.029] \\
PLIP & [0.003, 0.020] & [-0.089, -0.056] & [0.000, 0.006] & [-0.079, -0.058] & [0.000, 0.001] & [-0.059, -0.052] & [0.172, 0.218] & [0.059, 0.102] & [0.000, 0.000] & [-0.100, -0.075] \\
MI-Zero & [0.000, 0.000] & [-0.127, -0.095] & [0.303, 0.379] & [0.348, 0.414] & [0.000, 0.000] & [-0.145, -0.143] & [0.000, 0.000] & [-0.150, -0.135] & [0.000, 0.000] & [-0.131, -0.078] \\
CONCH & [0.681, 0.764] & [-0.049, -0.005] & [0.000, 0.000] & [-0.146, -0.133] & [0.000, 0.000] & [-0.085, -0.080] & [0.005, 0.017] & [-0.158, -0.142] & [0.004, 0.038] & [-0.092, -0.050] \\
BiomedCLIP & [0.000, 0.000] & [-0.193, -0.176] & [0.000, 0.000] & [-0.216, -0.207] & [0.000, 0.000] & [-0.138, -0.136] & [0.017, 0.035] & [-0.122, -0.107] & [0.000, 0.000] & [0.025, 0.086] \\
PathGen-CLIP & [0.073, 0.128] & [-0.160, -0.118] & [0.000, 0.000] & [-0.128, -0.106] & [0.023, 0.029] & [-0.070, -0.061] & [0.271, 0.318] & [0.147, 0.196] & [0.127, 0.219] & [0.125, 0.202] \\
MUSK & [0.336, 0.429] & [0.007, 0.109] & [0.000, 0.000] & [-0.134, -0.114] & [0.002, 0.003] & [0.064, 0.070] & [0.010, 0.023] & [-0.120, -0.099] & [0.367, 0.498] & [0.346, 0.456] \\
Patho-CLIP & [0.000, 0.013] & [-0.185, -0.155] & [0.028, 0.062] & [-0.062, -0.038] & [0.016, 0.021] & [-0.074, -0.070] & [0.224, 0.268] & \textit{[-0.014, 0.023]} & [0.034, 0.093] & \textit{[-0.036, 0.026]} \\
KEEP & [0.741, 0.819] & [0.158, 0.291] & [0.000, 0.000] & [-0.064, -0.012] & [0.471, 0.489] & [0.257, 0.269] & [0.062, 0.090] & [0.098, 0.130] & [0.080, 0.165] & [0.054, 0.161] \\
\bottomrule
\end{tabular}
}
\end{table*}

\begin{table*}[t]
\centering
\caption{Slide-level bootstrap confidence intervals for Prediction Consistency (P.C.) and Prompt Stability Score (PSS) under length and reporting-style variation (T3 Variant~2). Intervals are 2.5th--97.5th percentiles over 1{,}000 bootstrap resamples of slides drawn with replacement (seed 42). Point estimates are those reported in Table~\ref{tab:robustness_length}. Intervals shown in \textit{italics} contain zero.}
\label{tab:t3_bootstrap_ci_v2}
\resizebox{\textwidth}{!}{%
\begin{tabular}{l cc cc cc cc cc}
\toprule
\multirow{2}{*}{Model}
& \multicolumn{2}{c}{\textbf{CAMELYON16}} & \multicolumn{2}{c}{\textbf{CAMELYON17}}
& \multicolumn{2}{c}{\textbf{PANDA}} & \multicolumn{2}{c}{\textbf{TCGA-GBMLGG}}
& \multicolumn{2}{c}{\textbf{UniToPatho}} \\
\cmidrule(lr){2-3}\cmidrule(lr){4-5}\cmidrule(lr){6-7}\cmidrule(lr){8-9}\cmidrule(lr){10-11}
& P.C. & PSS & P.C. & PSS & P.C. & PSS & P.C. & PSS & P.C. & PSS \\
\midrule
QuiltNet & [0.000, 0.000] & [-0.333, -0.333] & [0.000, 0.000] & [-0.305, -0.290] & [0.000, 0.000] & [-0.205, -0.200] & [0.158, 0.204] & [-0.105, -0.055] & [0.000, 0.029] & [-0.137, -0.057] \\
PLIP & [0.386, 0.489] & \textit{[-0.051, 0.006]} & [0.140, 0.202] & [0.076, 0.121] & [0.209, 0.225] & [-0.067, -0.060] & [0.034, 0.054] & [-0.116, -0.086] & [0.000, 0.000] & [-0.124, -0.077] \\
MI-Zero & [0.010, 0.040] & [-0.296, -0.249] & [0.313, 0.395] & [0.077, 0.145] & [0.000, 0.000] & [-0.127, -0.121] & [0.147, 0.186] & [-0.135, -0.109] & [0.000, 0.000] & [-0.282, -0.257] \\
CONCH & [0.000, 0.007] & [-0.301, -0.280] & [0.000, 0.000] & [-0.262, -0.250] & [0.000, 0.002] & [-0.114, -0.108] & [0.343, 0.394] & [0.019, 0.065] & [0.000, 0.000] & [-0.113, -0.076] \\
BiomedCLIP & [0.000, 0.000] & [-0.266, -0.253] & [0.000, 0.000] & [-0.322, -0.302] & [0.000, 0.000] & [-0.264, -0.259] & [0.000, 0.000] & [-0.269, -0.261] & [0.802, 0.894] & [0.046, 0.231] \\
PathGen-CLIP & [0.950, 0.985] & [0.044, 0.445] & [0.000, 0.010] & [-0.190, -0.167] & [0.063, 0.073] & [-0.115, -0.108] & [0.189, 0.233] & [0.093, 0.139] & [0.029, 0.089] & \textit{[-0.025, 0.049]} \\
MUSK & [0.276, 0.363] & [-0.144, -0.081] & [0.002, 0.016] & [-0.177, -0.147] & [0.165, 0.180] & [0.047, 0.059] & [0.104, 0.139] & [0.014, 0.053] & [0.418, 0.536] & [0.089, 0.196] \\
Patho-CLIP & [0.010, 0.038] & [-0.320, -0.293] & [0.020, 0.056] & [-0.143, -0.104] & [0.134, 0.147] & [-0.038, -0.029] & [0.085, 0.118] & [-0.116, -0.083] & [0.257, 0.371] & [0.005, 0.131] \\
KEEP & [0.897, 0.950] & [0.205, 0.511] & [0.000, 0.000] & [-0.200, -0.172] & [0.261, 0.277] & [0.125, 0.139] & [0.103, 0.135] & \textit{[-0.012, 0.038]} & [0.000, 0.013] & [-0.136, -0.096] \\
\bottomrule
\end{tabular}
}
\end{table*}

\subsection{Open Vocabulary Retrieval}
\label{app:appendix_open_vocab}

Tables~\ref{tab:recall_set1_openvocab}-\ref{tab:recall_set4_alldatasets} report the full open vocabulary retrieval results across four candidate sets. For Set 1 (Open Vocabulary), Patho-CLIP achieves the highest mean recall on CAMELYON16 (0.891), while KEEP leads on PANDA and CAMELYON17, and PathGen-CLIP on UniToPatho. Notably, MI-Zero and BiomedCLIP score near zero on certain datasets, suggesting limited generalization to open vocabulary descriptions. For Set 2 (Label-Specific Mixed Diagnostic Candidates), KEEP consistently ranks among the top performers, particularly on CAMELYON16, CAMELYON17, and PANDA, while PLIP leads on UniToPatho. For Set 3 (Cross-Organ), KEEP dominates across CAMELYON16, CAMELYON17, and PANDA, and PathGen-CLIP achieves near-perfect recall on TCGA-GBMLGG and UniToPatho, with QuiltNet, PathGen-CLIP, and KEEP all reaching perfect recall on UniToPatho. For Set 4 (All Dataset Labels), KEEP again shows strong overall performance, achieving the best mean recall on CAMELYON17, PANDA, and competitive results elsewhere, while PathGen-CLIP leads on TCGA-GBMLGG. Across all four sets, KEEP and PathGen-CLIP emerge as the most robust models, whereas BiomedCLIP and MI-Zero struggle consistently, highlighting significant variation in open vocabulary generalization across models.

\begin{table*}[t]
\centering
\caption{Recall performance for Set 1 -- Open Vocabulary. We report mean recall and Recall@K across available cutoffs. The best result for each metric within each dataset is highlighted in bold. \textit{Chance} denotes the expected performance of a uniformly random ranking over the same candidate pool.}
\label{tab:recall_set1_openvocab}
\resizebox{\textwidth}{!}{
\begin{tabular}{lccccccccccc}
\toprule
Models & Mean & R@1 & R@2 & R@3 & R@4 & R@5 & R@6 & R@7 & R@8 & R@9 & R@10 \\
\midrule
\rowcolor{gray!10}
\multicolumn{12}{l}{\textbf{CAMELYON16}} \\
\textit{Chance} & \textit{0.275} & \textit{0.050} & \textit{0.100} & \textit{0.150} & \textit{0.200} & \textit{0.250} & \textit{0.300} & \textit{0.350} & \textit{0.400} & \textit{0.450} & \textit{0.500} \\
\addlinespace[1pt]
QuiltNet & 0.602 & 0.000 & 0.191 & 0.352 & 0.538 & 0.666 & 0.756 & 0.842 & 0.862 & 0.892 & 0.922 \\
PLIP & 0.478 & 0.005 & 0.007 & 0.085 & 0.251 & 0.451 & 0.584 & 0.732 & 0.830 & 0.902 & 0.932 \\
MI-Zero & 0.000 & 0.000 & 0.000 & 0.000 & 0.000 & 0.000 & 0.000 & 0.000 & 0.000 & 0.000 & 0.000 \\
CONCH & 0.313 & 0.073 & 0.143 & 0.201 & 0.347 & 0.384 & 0.395 & 0.395 & 0.397 & 0.397 & 0.397 \\
BiomedCLIP & 0.132 & 0.000 & 0.000 & 0.003 & 0.010 & 0.028 & 0.095 & 0.148 & 0.251 & 0.386 & 0.401 \\
PathGen-CLIP & 0.690 & \textbf{0.526} & 0.564 & 0.587 & 0.622 & 0.652 & 0.714 & 0.764 & 0.794 & 0.815 & 0.860 \\
MUSK & 0.624 & 0.060 & 0.421 & 0.531 & 0.604 & 0.659 & 0.704 & 0.742 & 0.779 & 0.817 & 0.922 \\
Patho-CLIP & \textbf{0.891} & 0.524 & \textbf{0.657} & \textbf{0.812} & \textbf{0.955} & \textbf{0.980} & \textbf{0.993} & \textbf{0.998} & \textbf{0.998} & \textbf{0.998} & \textbf{0.998} \\
KEEP & 0.767 & 0.337 & 0.613 & 0.623 & 0.681 & 0.764 & 0.824 & 0.912 & 0.945 & 0.972 & 0.995 \\
\addlinespace[3pt]
\rowcolor{gray!10}
\multicolumn{12}{l}{\textbf{CAMELYON17}} \\
\textit{Chance} & \textit{0.275} & \textit{0.050} & \textit{0.100} & \textit{0.150} & \textit{0.200} & \textit{0.250} & \textit{0.300} & \textit{0.350} & \textit{0.400} & \textit{0.450} & \textit{0.500} \\
\addlinespace[1pt]
QuiltNet & 0.349 & 0.012 & 0.112 & 0.148 & 0.178 & 0.222 & 0.403 & 0.513 & 0.571 & 0.637 & 0.689 \\
PLIP & 0.672 & \textbf{0.511} & 0.523 & 0.567 & 0.629 & 0.671 & 0.711 & 0.742 & 0.765 & 0.788 & 0.810 \\
MI-Zero & 0.649 & 0.170 & 0.499 & 0.579 & 0.639 & 0.701 & 0.744 & 0.763 & 0.773 & 0.796 & 0.826 \\
CONCH & 0.150 & 0.054 & 0.102 & 0.118 & 0.148 & 0.158 & 0.164 & 0.178 & 0.184 & 0.188 & 0.204 \\
BiomedCLIP & 0.044 & 0.000 & 0.000 & 0.000 & 0.000 & 0.000 & 0.018 & 0.032 & 0.118 & 0.118 & 0.158 \\
PathGen-CLIP & 0.707 & 0.347 & \textbf{0.567} & 0.611 & 0.645 & 0.697 & 0.750 & 0.818 & 0.844 & 0.876 & 0.914 \\
MUSK & 0.700 & 0.064 & 0.377 & \textbf{0.675} & \textbf{0.709} & 0.753 & 0.788 & 0.842 & 0.880 & 0.926 & \textbf{0.982} \\
Patho-CLIP & 0.669 & 0.012 & 0.365 & 0.553 & 0.671 & 0.755 & 0.810 & 0.844 & 0.870 & 0.896 & 0.914 \\
KEEP & \textbf{0.709} & 0.138 & 0.275 & 0.617 & 0.679 & \textbf{0.780} & \textbf{0.844} & \textbf{0.900} & \textbf{0.932} & \textbf{0.958} & 0.966 \\
\addlinespace[3pt]
\rowcolor{gray!10}
\multicolumn{12}{l}{\textbf{PANDA}} \\
\textit{Chance} & \textit{0.275} & \textit{0.050} & \textit{0.100} & \textit{0.150} & \textit{0.200} & \textit{0.250} & \textit{0.300} & \textit{0.350} & \textit{0.400} & \textit{0.450} & \textit{0.500} \\
\addlinespace[1pt]
QuiltNet & 0.178 & 0.002 & 0.012 & 0.029 & 0.065 & 0.114 & 0.174 & 0.245 & 0.311 & 0.379 & 0.445 \\
PLIP & 0.461 & 0.218 & 0.262 & 0.295 & 0.339 & 0.390 & 0.461 & 0.540 & 0.621 & 0.703 & 0.778 \\
MI-Zero & 0.046 & 0.000 & 0.000 & 0.000 & 0.023 & 0.041 & 0.053 & 0.063 & 0.078 & 0.089 & 0.109 \\
CONCH & 0.106 & 0.025 & 0.114 & 0.115 & 0.115 & 0.115 & 0.115 & 0.115 & 0.115 & 0.115 & 0.119 \\
BiomedCLIP & 0.158 & 0.000 & 0.000 & 0.000 & 0.007 & 0.084 & 0.130 & 0.242 & 0.324 & 0.368 & 0.429 \\
PathGen-CLIP & 0.564 & 0.274 & 0.283 & 0.391 & 0.469 & 0.547 & 0.635 & 0.697 & 0.742 & 0.781 & 0.825 \\
MUSK & 0.622 & 0.216 & 0.341 & \textbf{0.427} & \textbf{0.515} & 0.600 & 0.689 & 0.778 & 0.852 & 0.890 & 0.916 \\
Patho-CLIP & 0.280 & 0.228 & 0.261 & 0.270 & 0.273 & 0.278 & 0.282 & 0.286 & 0.294 & 0.306 & 0.323 \\
KEEP & \textbf{0.640} & \textbf{0.304} & \textbf{0.357} & 0.406 & 0.477 & \textbf{0.635} & \textbf{0.736} & \textbf{0.812} & \textbf{0.861} & \textbf{0.896} & \textbf{0.917} \\
\addlinespace[3pt]
\rowcolor{gray!10}
\multicolumn{12}{l}{\textbf{TCGA-GBMLGG}} \\
\textit{Chance} & \textit{0.275} & \textit{0.050} & \textit{0.100} & \textit{0.150} & \textit{0.200} & \textit{0.250} & \textit{0.300} & \textit{0.350} & \textit{0.400} & \textit{0.450} & \textit{0.500} \\
\addlinespace[1pt]
QuiltNet & 0.210 & 0.000 & 0.002 & 0.013 & 0.059 & 0.108 & 0.179 & 0.267 & 0.361 & 0.499 & 0.616 \\
PLIP & 0.608 & 0.009 & 0.104 & 0.277 & 0.496 & 0.661 & 0.798 & 0.876 & 0.924 & 0.959 & 0.980 \\
MI-Zero & \textbf{0.856} & \textbf{0.331} & \textbf{0.618} & \textbf{0.753} & \textbf{0.893} & \textbf{0.980} & \textbf{0.996} & \textbf{0.997} & \textbf{0.998} & \textbf{0.999} & \textbf{1.000} \\
CONCH & 0.165 & 0.021 & 0.041 & 0.073 & 0.111 & 0.146 & 0.176 & 0.211 & 0.243 & 0.285 & 0.339 \\
BiomedCLIP & 0.000 & 0.000 & 0.000 & 0.000 & 0.000 & 0.000 & 0.000 & 0.000 & 0.000 & 0.000 & 0.000 \\
PathGen-CLIP & 0.703 & 0.119 & 0.267 & 0.428 & 0.634 & 0.752 & 0.894 & 0.965 & 0.984 & 0.993 & 0.995 \\
MUSK & 0.585 & 0.047 & 0.159 & 0.281 & 0.430 & 0.619 & 0.734 & 0.830 & 0.886 & 0.920 & 0.944 \\
Patho-CLIP & 0.116 & 0.005 & 0.016 & 0.030 & 0.058 & 0.084 & 0.117 & 0.153 & 0.188 & 0.235 & 0.277 \\
KEEP & 0.685 & 0.059 & 0.250 & 0.436 & 0.596 & 0.744 & 0.861 & 0.942 & 0.975 & 0.995 & 0.996 \\
\addlinespace[3pt]
\rowcolor{gray!10}
\multicolumn{12}{l}{\textbf{UniToPatho}} \\
\textit{Chance} & \textit{0.275} & \textit{0.050} & \textit{0.100} & \textit{0.150} & \textit{0.200} & \textit{0.250} & \textit{0.300} & \textit{0.350} & \textit{0.400} & \textit{0.450} & \textit{0.500} \\
\addlinespace[1pt]
QuiltNet & 0.774 & 0.017 & 0.595 & 0.823 & 0.835 & 0.835 & 0.861 & 0.916 & 0.924 & 0.962 & 0.971 \\
PLIP & 0.822 & \textbf{0.595} & 0.730 & 0.810 & 0.823 & 0.848 & 0.857 & 0.886 & 0.886 & 0.890 & 0.899 \\
MI-Zero & 0.476 & 0.000 & 0.017 & 0.139 & 0.262 & 0.544 & 0.679 & 0.743 & 0.764 & 0.797 & 0.819 \\
CONCH & 0.434 & 0.253 & 0.329 & 0.371 & 0.397 & 0.435 & 0.452 & 0.477 & 0.510 & 0.544 & 0.570 \\
BiomedCLIP & 0.083 & 0.000 & 0.008 & 0.059 & 0.089 & 0.110 & 0.110 & 0.114 & 0.114 & 0.114 & 0.114 \\
PathGen-CLIP & \textbf{0.916} & 0.557 & 0.747 & \textbf{0.907} & \textbf{0.954} & \textbf{0.992} & \textbf{1.000} & \textbf{1.000} & \textbf{1.000} & \textbf{1.000} & \textbf{1.000} \\
MUSK & 0.157 & 0.072 & 0.080 & 0.093 & 0.105 & 0.135 & 0.152 & 0.173 & 0.219 & 0.257 & 0.287 \\
Patho-CLIP & 0.861 & 0.422 & 0.675 & 0.793 & 0.840 & 0.903 & 0.979 & 0.996 & \textbf{1.000} & \textbf{1.000} & \textbf{1.000} \\
KEEP & 0.896 & 0.544 & \textbf{0.755} & 0.848 & 0.894 & 0.916 & \textbf{1.000} & \textbf{1.000} & \textbf{1.000} & \textbf{1.000} & \textbf{1.000} \\
\bottomrule
\end{tabular}
}
\end{table*}

\begin{table*}[t]
\centering
\caption{Recall performance for Set 2 -- Label-Specific Mixed Diagnostic Candidates. We report mean recall and Recall@K across available cutoffs. The best result for each metric within each dataset is highlighted in bold.}
\label{tab:recall_set2_finegrained}{
\begin{tabular}{lcccccc}
\toprule
Models & Mean & R@1 & R@2 & R@3 & R@4 & R@5 \\
\midrule
\rowcolor{gray!10}
\multicolumn{7}{l}{\textbf{CAMELYON16}} \\
\addlinespace[1pt]
QuiltNet & 0.253 & 0.000 & 0.010 & 0.395 & 0.402 & 0.460 \\
PLIP & 0.658 & \textbf{0.373} & 0.404 & 0.529 & 0.988 & 0.998 \\
MI-Zero & 0.328 & 0.000 & 0.179 & 0.222 & 0.502 & 0.735 \\
CONCH & 0.185 & 0.003 & 0.038 & 0.151 & 0.266 & 0.467 \\
BiomedCLIP & 0.384 & 0.000 & 0.123 & 0.599 & 0.599 & 0.599 \\
PathGen-CLIP & 0.560 & 0.276 & 0.514 & 0.569 & 0.632 & 0.809 \\
MUSK & 0.351 & 0.000 & 0.191 & 0.391 & 0.544 & 0.632 \\
Patho-CLIP & 0.344 & 0.130 & 0.376 & 0.396 & 0.404 & 0.413 \\
KEEP & \textbf{0.690} & 0.194 & \textbf{0.578} & \textbf{0.686} & \textbf{0.995} & \textbf{1.000} \\
\addlinespace[3pt]
\rowcolor{gray!10}
\multicolumn{7}{l}{\textbf{CAMELYON17}} \\
\addlinespace[1pt]
QuiltNet & 0.286 & 0.072 & 0.186 & 0.327 & 0.363 & 0.483 \\
PLIP & 0.684 & \textbf{0.421} & 0.559 & 0.647 & 0.870 & 0.924 \\
MI-Zero & 0.405 & 0.000 & 0.194 & 0.501 & 0.633 & 0.697 \\
CONCH & 0.228 & 0.002 & 0.108 & 0.156 & 0.311 & 0.563 \\
BiomedCLIP & 0.365 & 0.050 & 0.160 & 0.431 & 0.545 & 0.637 \\
PathGen-CLIP & 0.618 & 0.216 & 0.483 & 0.691 & 0.824 & 0.878 \\
MUSK & 0.413 & 0.000 & 0.166 & 0.567 & 0.625 & 0.705 \\
Patho-CLIP & 0.250 & 0.052 & 0.142 & 0.218 & 0.319 & 0.519 \\
KEEP & \textbf{0.729} & 0.311 & \textbf{0.563} & \textbf{0.898} & \textbf{0.928} & \textbf{0.946} \\
\addlinespace[3pt]
\rowcolor{gray!10}
\multicolumn{7}{l}{\textbf{PANDA}} \\
\addlinespace[1pt]
QuiltNet & 0.147 & 0.008 & 0.025 & 0.105 & 0.221 & 0.375 \\
PLIP & 0.471 & 0.112 & 0.292 & 0.502 & 0.676 & 0.772 \\
MI-Zero & 0.311 & 0.085 & 0.224 & 0.304 & 0.386 & 0.556 \\
CONCH & 0.437 & 0.190 & 0.254 & 0.440 & 0.590 & 0.711 \\
BiomedCLIP & 0.021 & 0.000 & 0.000 & 0.000 & 0.000 & 0.107 \\
PathGen-CLIP & 0.687 & \textbf{0.383} & 0.504 & 0.710 & 0.841 & \textbf{0.998} \\
MUSK & 0.724 & 0.306 & 0.578 & 0.794 & 0.960 & 0.985 \\
Patho-CLIP & 0.472 & 0.296 & 0.370 & 0.441 & 0.551 & 0.703 \\
KEEP & \textbf{0.810} & 0.341 & \textbf{0.755} & \textbf{0.967} & \textbf{0.993} & 0.997 \\
\addlinespace[3pt]
\rowcolor{gray!10}
\multicolumn{7}{l}{\textbf{TCGA-GBMLGG}} \\
\addlinespace[1pt]
QuiltNet & 0.739 & \textbf{0.518} & \textbf{0.607} & 0.735 & 0.895 & 0.936 \\
PLIP & 0.613 & 0.098 & 0.220 & 0.761 & 0.987 & \textbf{0.998} \\
MI-Zero & 0.136 & 0.051 & 0.082 & 0.139 & 0.204 & 0.205 \\
CONCH & 0.658 & 0.199 & 0.501 & 0.722 & 0.909 & 0.960 \\
BiomedCLIP & 0.120 & 0.000 & 0.000 & 0.000 & 0.053 & 0.548 \\
PathGen-CLIP & \textbf{0.743} & 0.315 & 0.491 & \textbf{0.917} & \textbf{0.994} & 0.997 \\
MUSK & 0.232 & 0.036 & 0.089 & 0.178 & 0.349 & 0.506 \\
Patho-CLIP & 0.556 & 0.086 & 0.239 & 0.602 & 0.906 & 0.947 \\
KEEP & 0.739 & 0.364 & 0.512 & 0.838 & 0.986 & 0.993 \\
\addlinespace[3pt]
\rowcolor{gray!10}
\multicolumn{7}{l}{\textbf{UniToPatho}} \\
\addlinespace[1pt]
QuiltNet & 0.544 & 0.169 & 0.409 & 0.633 & 0.730 & 0.781 \\
PLIP & \textbf{0.650} & \textbf{0.489} & \textbf{0.629} & 0.667 & 0.696 & 0.768 \\
MI-Zero & 0.138 & 0.008 & 0.013 & 0.101 & 0.219 & 0.350 \\
CONCH & 0.167 & 0.046 & 0.110 & 0.177 & 0.228 & 0.274 \\
BiomedCLIP & 0.221 & 0.148 & 0.219 & 0.219 & 0.249 & 0.270 \\
PathGen-CLIP & 0.635 & 0.266 & 0.578 & 0.662 & 0.797 & \textbf{0.873} \\
MUSK & 0.230 & 0.000 & 0.000 & 0.072 & 0.494 & 0.582 \\
Patho-CLIP & 0.322 & 0.089 & 0.156 & 0.266 & 0.430 & 0.671 \\
KEEP & 0.624 & 0.156 & 0.515 & \textbf{0.734} & \textbf{0.844} & 0.869 \\
\bottomrule
\end{tabular}
}
\end{table*}

\clearpage

\begin{table*}[t]
\centering
\caption{Recall performance for Set 3 -- Cross-Organ. We report mean recall and Recall@K across available cutoffs. The best result for each metric within each dataset is highlighted in bold.}
\label{tab:recall_set3_crossorgan}
{\begin{tabular}{lcccccc}
\toprule
Models & Mean & R@1 & R@2 & R@3 & R@4 & R@5 \\
\midrule
\rowcolor{gray!10}
\multicolumn{7}{l}{\textbf{CAMELYON16}} \\
\addlinespace[1pt]

QuiltNet & 0.630 & 0.191 & 0.374 & 0.668 & 0.915 & \textbf{1.000} \\
PLIP & 0.819 & 0.421 & 0.769 & 0.925 & 0.980 & 0.998 \\
MI-Zero & 0.335 & 0.000 & 0.253 & 0.384 & 0.432 & 0.606 \\
CONCH & 0.445 & 0.364 & 0.395 & 0.420 & 0.482 & 0.563 \\
BiomedCLIP & 0.029 & 0.000 & 0.000 & 0.003 & 0.028 & 0.115 \\
PathGen-CLIP & 0.619 & 0.454 & 0.584 & 0.647 & 0.689 & 0.722 \\
MUSK & 0.415 & 0.003 & 0.120 & 0.386 & 0.682 & 0.885 \\
Patho-CLIP & 0.413 & 0.143 & 0.336 & 0.429 & 0.541 & 0.614 \\
KEEP & \textbf{0.895} & \textbf{0.593} & \textbf{0.932} & \textbf{0.967} & \textbf{0.985} & 0.998 \\
\addlinespace[3pt]
\rowcolor{gray!10}
\multicolumn{7}{l}{\textbf{CAMELYON17}} \\
\addlinespace[1pt]
QuiltNet & 0.654 & 0.285 & 0.475 & 0.675 & 0.874 & 0.960 \\
PLIP & 0.788 & 0.499 & 0.763 & 0.862 & 0.890 & 0.924 \\
MI-Zero & 0.384 & 0.000 & 0.351 & 0.455 & 0.473 & 0.643 \\
CONCH & 0.564 & 0.218 & 0.473 & 0.587 & 0.695 & 0.846 \\
BiomedCLIP & 0.229 & 0.000 & 0.100 & 0.116 & 0.132 & 0.796 \\
PathGen-CLIP & 0.748 & 0.491 & 0.625 & 0.830 & 0.868 & 0.928 \\
MUSK & 0.476 & 0.012 & 0.154 & 0.467 & 0.852 & 0.896 \\
Patho-CLIP & 0.484 & 0.188 & 0.373 & 0.523 & 0.603 & 0.732 \\
KEEP & \textbf{0.987} & \textbf{0.950} & \textbf{0.992} & \textbf{0.994} & \textbf{1.000} & \textbf{1.000} \\
\addlinespace[3pt]
\rowcolor{gray!10}
\multicolumn{7}{l}{\textbf{PANDA}} \\
\addlinespace[1pt]
QuiltNet & 0.056 & 0.003 & 0.009 & 0.027 & 0.064 & 0.178 \\
PLIP & 0.401 & 0.177 & 0.288 & 0.399 & 0.510 & 0.632 \\
MI-Zero & 0.394 & 0.025 & 0.177 & 0.231 & 0.757 & 0.779 \\
CONCH & 0.693 & 0.285 & 0.722 & 0.776 & 0.821 & 0.863 \\
BiomedCLIP & 0.041 & 0.000 & 0.000 & 0.000 & 0.000 & 0.207 \\
PathGen-CLIP & 0.853 & 0.326 & 0.960 & 0.985 & 0.995 & 0.999 \\
MUSK & 0.895 & 0.545 & 0.940 & \textbf{0.990} & \textbf{0.999} & \textbf{1.000} \\
Patho-CLIP & 0.495 & 0.224 & 0.409 & 0.529 & 0.619 & 0.695 \\
KEEP & \textbf{0.913} & \textbf{0.634} & \textbf{0.972} & 0.983 & 0.988 & 0.991 \\
\addlinespace[3pt]
\rowcolor{gray!10}
\multicolumn{7}{l}{\textbf{TCGA-GBMLGG}} \\
\addlinespace[1pt]
QuiltNet & 0.844 & 0.573 & 0.845 & 0.894 & 0.940 & 0.965 \\
PLIP & 0.968 & 0.909 & 0.956 & 0.988 & 0.993 & 0.996 \\
MI-Zero & 0.322 & 0.119 & 0.309 & 0.375 & 0.395 & 0.414 \\
CONCH & 0.935 & 0.725 & 0.962 & 0.995 & 0.996 & 0.999 \\
BiomedCLIP & 0.327 & 0.000 & 0.000 & 0.451 & 0.540 & 0.646 \\
PathGen-CLIP & \textbf{0.996} & \textbf{0.991} & \textbf{0.995} & \textbf{0.997} & \textbf{0.999} & \textbf{1.000} \\
MUSK & 0.327 & 0.020 & 0.083 & 0.282 & 0.497 & 0.755 \\
Patho-CLIP & 0.885 & 0.754 & 0.840 & 0.915 & 0.951 & 0.967 \\
KEEP & 0.987 & 0.973 & 0.983 & 0.991 & 0.993 & 0.993 \\
\addlinespace[3pt]
\rowcolor{gray!10}
\multicolumn{7}{l}{\textbf{UniToPatho}} \\
\addlinespace[1pt]
QuiltNet & \textbf{1.000} & \textbf{1.000} & \textbf{1.000} & \textbf{1.000} & \textbf{1.000} & \textbf{1.000} \\
PLIP & 0.979 & 0.954 & 0.979 & 0.983 & 0.987 & 0.992 \\
MI-Zero & 0.118 & 0.004 & 0.025 & 0.063 & 0.148 & 0.350 \\
CONCH & 0.188 & 0.059 & 0.118 & 0.152 & 0.262 & 0.350 \\
BiomedCLIP & 0.197 & 0.021 & 0.156 & 0.232 & 0.236 & 0.338 \\
PathGen-CLIP & \textbf{1.000} & \textbf{1.000} & \textbf{1.000} & \textbf{1.000} & \textbf{1.000} & \textbf{1.000} \\
MUSK & 0.089 & 0.000 & 0.008 & 0.080 & 0.148 & 0.211 \\
Patho-CLIP & 0.940 & 0.781 & 0.941 & 0.979 & \textbf{1.000} & \textbf{1.000} \\
KEEP & \textbf{1.000} & \textbf{1.000} & \textbf{1.000} & \textbf{1.000} & \textbf{1.000} & \textbf{1.000} \\
\bottomrule
\end{tabular}
}
\end{table*}

\clearpage

\begin{table*}[t]
\centering
\caption{Recall performance for Set 4 -- All Dataset Labels. We report mean recall and Recall@K across available cutoffs. The best result for each metric within each dataset is highlighted in bold.}
\label{tab:recall_set4_alldatasets}
\resizebox{\textwidth}{!}{
\begin{tabular}{lccccccccccc}
\toprule
Models & Mean & R@1 & R@2 & R@3 & R@4 & R@5 & R@6 & R@7 & R@8 & R@9 & R@10 \\
\midrule
\rowcolor{gray!10}
\multicolumn{12}{l}{\textbf{CAMELYON16}} \\
\addlinespace[1pt]
QuiltNet & 0.604 & 0.000 & 0.018 & 0.384 & 0.422 & 0.530 & 0.686 & 0.995 & \textbf{1.000} & \textbf{1.000} & \textbf{1.000} \\
PLIP & \textbf{0.839} & 0.368 & 0.494 & \textbf{0.652} & 0.895 & 0.990 & 0.998 & 0.998 & 0.998 & 0.998 & \textbf{1.000} \\
MI-Zero & 0.477 & 0.000 & 0.091 & 0.114 & 0.217 & 0.543 & 0.644 & 0.689 & 0.760 & 0.828 & 0.881 \\
CONCH & 0.410 & 0.000 & 0.000 & 0.020 & 0.111 & 0.231 & 0.447 & 0.711 & 0.784 & 0.869 & 0.922 \\
BiomedCLIP & 0.123 & 0.000 & 0.000 & 0.000 & 0.005 & 0.015 & 0.033 & 0.045 & 0.118 & 0.499 & 0.519 \\
PathGen-CLIP & 0.691 & \textbf{0.461} & \textbf{0.526} & 0.571 & 0.624 & 0.722 & 0.759 & 0.782 & 0.809 & 0.822 & 0.830 \\
MUSK & 0.361 & 0.003 & 0.203 & 0.223 & 0.253 & 0.288 & 0.346 & 0.534 & 0.571 & 0.594 & 0.594 \\
Patho-CLIP & 0.505 & 0.075 & 0.266 & 0.356 & 0.474 & 0.541 & 0.602 & 0.627 & 0.659 & 0.704 & 0.742 \\
KEEP & 0.811 & 0.186 & 0.306 & 0.643 & \textbf{0.977} & \textbf{1.000} & \textbf{1.000} & \textbf{1.000} & \textbf{1.000} & \textbf{1.000} & \textbf{1.000} \\
\addlinespace[3pt]
\rowcolor{gray!10}
\multicolumn{12}{l}{\textbf{CAMELYON17}} \\
\addlinespace[1pt]
QuiltNet & 0.614 & 0.072 & 0.192 & 0.323 & 0.405 & 0.505 & 0.711 & 0.952 & 0.986 & 0.998 & 0.998 \\
PLIP & 0.802 & \textbf{0.473} & \textbf{0.605} & 0.647 & 0.744 & 0.812 & 0.900 & 0.934 & 0.952 & 0.970 & 0.982 \\
MI-Zero & 0.265 & 0.238 & 0.244 & 0.253 & 0.261 & 0.261 & 0.273 & 0.275 & 0.275 & 0.275 & 0.293 \\
CONCH & 0.306 & 0.002 & 0.002 & 0.022 & 0.072 & 0.244 & 0.347 & 0.489 & 0.551 & 0.629 & 0.705 \\
BiomedCLIP & 0.261 & 0.000 & 0.000 & 0.050 & 0.100 & 0.108 & 0.120 & 0.397 & 0.597 & 0.603 & 0.637 \\
PathGen-CLIP & 0.694 & 0.291 & 0.461 & 0.633 & 0.689 & 0.759 & 0.788 & 0.802 & 0.810 & 0.842 & 0.862 \\
MUSK & 0.375 & 0.004 & 0.176 & 0.190 & 0.259 & 0.311 & 0.397 & 0.545 & 0.591 & 0.631 & 0.643 \\
Patho-CLIP & 0.382 & 0.064 & 0.156 & 0.248 & 0.339 & 0.397 & 0.429 & 0.481 & 0.533 & 0.569 & 0.603 \\
KEEP & \textbf{0.865} & 0.335 & 0.581 & \textbf{0.822} & \textbf{0.938} & \textbf{0.992} & \textbf{0.994} & \textbf{0.994} & \textbf{0.998} & \textbf{1.000} & \textbf{1.000} \\
\addlinespace[3pt]
\rowcolor{gray!10}
\multicolumn{12}{l}{\textbf{PANDA}} \\
\addlinespace[1pt]
QuiltNet & 0.102 & 0.000 & 0.000 & 0.001 & 0.003 & 0.014 & 0.032 & 0.107 & 0.198 & 0.283 & 0.378 \\
PLIP & 0.468 & 0.187 & 0.256 & 0.304 & 0.365 & 0.433 & 0.508 & 0.565 & 0.624 & 0.689 & 0.750 \\
MI-Zero & 0.272 & 0.020 & 0.055 & 0.074 & 0.099 & 0.149 & 0.262 & 0.388 & 0.485 & 0.561 & 0.625 \\
CONCH & 0.662 & 0.235 & 0.427 & 0.530 & 0.601 & 0.671 & 0.738 & 0.801 & 0.841 & 0.874 & 0.900 \\
BiomedCLIP & 0.072 & 0.000 & 0.000 & 0.000 & 0.000 & 0.000 & 0.000 & 0.000 & 0.171 & 0.272 & 0.272 \\
PathGen-CLIP & 0.757 & 0.271 & 0.412 & 0.518 & 0.629 & 0.759 & 0.983 & 0.996 & 0.999 & 1.000 & \textbf{1.000} \\
MUSK & 0.787 & 0.253 & 0.446 & 0.604 & 0.729 & 0.852 & 0.990 & \textbf{0.998} & \textbf{0.999} & \textbf{1.000} & 1.000 \\
Patho-CLIP & 0.559 & 0.218 & 0.287 & 0.387 & 0.482 & 0.551 & 0.632 & 0.689 & 0.735 & 0.783 & 0.830 \\
KEEP & \textbf{0.841} & \textbf{0.344} & \textbf{0.589} & \textbf{0.718} & \textbf{0.833} & \textbf{0.948} & \textbf{0.992} & 0.994 & 0.995 & 0.997 & 0.998 \\
\addlinespace[3pt]
\rowcolor{gray!10}
\multicolumn{12}{l}{\textbf{TCGA-GBMLGG}} \\
\addlinespace[1pt]
QuiltNet & 0.722 & 0.018 & 0.509 & 0.599 & 0.669 & 0.759 & 0.845 & 0.904 & 0.937 & 0.982 & 0.995 \\
PLIP & 0.899 & 0.136 & 0.924 & 0.962 & 0.984 & 0.991 & \textbf{0.996} & \textbf{0.998} & \textbf{0.999} & \textbf{0.999} & \textbf{0.999} \\
MI-Zero & 0.131 & 0.050 & 0.066 & 0.073 & 0.079 & 0.100 & 0.127 & 0.143 & 0.184 & 0.232 & 0.261 \\
CONCH & 0.773 & 0.180 & 0.447 & 0.681 & 0.787 & 0.871 & 0.915 & 0.946 & 0.956 & 0.969 & 0.981 \\
BiomedCLIP & 0.441 & 0.000 & 0.008 & 0.052 & 0.344 & 0.595 & 0.623 & 0.632 & 0.672 & 0.697 & 0.783 \\
PathGen-CLIP & \textbf{0.931} & \textbf{0.402} & \textbf{0.960} & \textbf{0.989} & \textbf{0.991} & \textbf{0.993} & 0.993 & 0.996 & 0.996 & 0.997 & 0.999 \\
MUSK & 0.282 & 0.018 & 0.042 & 0.072 & 0.094 & 0.126 & 0.165 & 0.305 & 0.471 & 0.688 & 0.836 \\
Patho-CLIP & 0.837 & 0.163 & 0.785 & 0.857 & 0.894 & 0.915 & 0.933 & 0.939 & 0.956 & 0.960 & 0.966 \\
KEEP & 0.927 & 0.400 & \textbf{0.960} & 0.979 & 0.986 & 0.988 & 0.990 & 0.993 & 0.993 & 0.993 & 0.993 \\
\addlinespace[3pt]
\rowcolor{gray!10}
\multicolumn{12}{l}{\textbf{UniToPatho}} \\
\addlinespace[1pt]
QuiltNet & 0.818 & 0.139 & 0.565 & 0.679 & 0.840 & 0.962 & 0.996 & 0.996 & \textbf{1.000} & \textbf{1.000} & \textbf{1.000} \\
PLIP & 0.847 & \textbf{0.481} & 0.625 & 0.747 & 0.852 & 0.899 & 0.958 & 0.962 & 0.975 & 0.983 & 0.987 \\
MI-Zero & 0.203 & 0.000 & 0.008 & 0.017 & 0.038 & 0.084 & 0.156 & 0.228 & 0.380 & 0.464 & 0.650 \\
CONCH & 0.133 & 0.051 & 0.076 & 0.101 & 0.114 & 0.127 & 0.143 & 0.160 & 0.173 & 0.186 & 0.198 \\
BiomedCLIP & 0.193 & 0.017 & 0.076 & 0.127 & 0.215 & 0.232 & 0.245 & 0.253 & 0.253 & 0.257 & 0.257 \\
PathGen-CLIP & \textbf{0.858} & 0.447 & 0.612 & 0.755 & 0.835 & 0.928 & \textbf{1.000} & \textbf{1.000} & \textbf{1.000} & \textbf{1.000} & \textbf{1.000} \\
MUSK & 0.043 & 0.000 & 0.004 & 0.008 & 0.008 & 0.008 & 0.008 & 0.051 & 0.084 & 0.114 & 0.148 \\
Patho-CLIP & 0.717 & 0.186 & 0.325 & 0.473 & 0.591 & 0.802 & 0.869 & 0.937 & 0.987 & \textbf{1.000} & \textbf{1.000} \\
KEEP & 0.851 & 0.186 & \textbf{0.646} & \textbf{0.810} & \textbf{0.899} & \textbf{0.971} & 0.996 & \textbf{1.000} & \textbf{1.000} & \textbf{1.000} & \textbf{1.000} \\
\bottomrule
\end{tabular}
}
\end{table*}

\clearpage
\onecolumn

\section{Prompt Lists}
\label{app:promptlist}



\subsection{Zero-Shot Classification Prompts}
\label{app:appendix_zeroshot_prompt}

Table~\ref{tab:prompts_zeroshot} lists all text prompts used for zero-shot
classification. Each prompt describes the visual appearance of a whole-slide
image corresponding to a specific class label.

\begin{longtable}{p{0.15\textwidth} p{0.78\textwidth}}
\caption{Text prompts used for zero-shot classification across all datasets.}
\label{tab:prompts_zeroshot} \\
\toprule
\textbf{Label} & \textbf{Prompt} \\
\midrule
\endfirsthead

\multicolumn{2}{c}{\small\itshape Table~\ref{tab:prompts_zeroshot} continued} \\
\toprule
\textbf{Label} & \textbf{Prompt} \\
\midrule
\endhead

\midrule
\multicolumn{2}{r}{\small\itshape Continued on next page} \\
\endfoot

\bottomrule
\endlastfoot

\rowcolor{gray!10}
\multicolumn{2}{l}{\textbf{PANDA}} \\
\addlinespace[2pt]

Grade 0 &
Benign prostate tissue with normal gland architecture, uniform small glands,
smooth luminal contours, and no nuclear atypia. Stroma is non-reactive and
orderly. No evidence of carcinoma. \\
\addlinespace

Grade 1 &
Low-grade prostate cancer with well-formed glands. Tumor glands are still
round, separate, and organized. Minimal infiltration into surrounding tissue.
Mild nuclear atypia with recognizable glandular structures. No fused or
cribriform patterns. \\
\addlinespace

Grade 2 &
Predominantly well-formed glands (pattern 3) with a lesser component of
poorly formed or fused small glands (pattern 4). Early architectural disorder,
some poorly formed glands beginning to appear. \\
\addlinespace

Grade 3 &
Predominantly poorly formed, fused, or cribriform glands (pattern 4) with a
smaller component of well-formed glands (pattern 3). Architectural
disorganization more prominent. Increased nuclear atypia. \\
\addlinespace

Grade 4 &
High-grade prostate cancer composed almost entirely of fused glands, poorly
formed glands, or cribriform structures. Marked architectural disruption and
significant nuclear atypia. No well-formed glands remain. \\
\addlinespace

Grade 5 &
Highest-grade prostate cancer with solid sheets of tumor cells, single-cell
infiltration, cords, or comedonecrosis. Complete loss of glandular structure.
Highly aggressive morphology with marked atypia. \\

\addlinespace[3pt]
\rowcolor{gray!10}
\multicolumn{2}{l}{\textbf{UniToPatho}} \\
\addlinespace[2pt]

NORM &
Benign colorectal tissue with no dysplasia or malignancy. \\
\addlinespace

HP &
Hyperplastic polyps are those in which the epithelial cells, although benign,
begin to outgrow their available space. The glands have an increased number of
goblet cells and therefore look pale or cleared out next to normal epithelium,
and there tends to be a thickened or sclerotic-appearing basement membrane
under the neoplastic cells. Because the surface area is outgrowing the lamina
propria, hyperplastic polyps take on a distinctive frilly or lacy appearance.
Crypts cut in cross section have a distinctive star-shaped lumen, also called
serrated architecture. \\
\addlinespace

TA.LG &
Tubular adenoma of the colon with low-grade dysplasia and predominantly
tubular glands with relatively smooth, parallel crypt architecture. \\
\addlinespace

TA.HG &
Tubular adenoma of the colon with high-grade dysplasia, showing back-to-back
or cribriform glands and marked cytologic dysplasia. \\
\addlinespace

TVA.LG &
Tubulovillous adenoma of the colon with mixed tubular and villous architecture
and low-grade cytologic dysplasia. \\
\addlinespace

TVA.HG &
Tubulovillous adenoma with high-grade epithelial dysplasia and complex
fused glandular architecture. \\

\addlinespace[3pt]
\rowcolor{gray!10}
\multicolumn{2}{l}{\textbf{CAMELYON16}} \\
\addlinespace[2pt]

Normal &
A benign lymph node with intact architecture and no metastatic carcinoma. \\
\addlinespace

Metastasis &
A lymph node containing metastatic carcinoma that forms disruptive
nests or clusters effacing normal nodal architecture. \\

\addlinespace[3pt]
\rowcolor{gray!10}
\multicolumn{2}{l}{\textbf{CAMELYON17}} \\
\addlinespace[2pt]

Negative &
A benign lymph node with preserved architecture and no metastatic carcinoma,
including no isolated tumor cells, micrometastases, or macrometastases. \\
\addlinespace

ITC &
A lymph node containing only isolated tumor cells---single cells or tiny
clusters under 0.2\,mm---without architectural distortion. \\
\addlinespace

Micro &
A lymph node with micrometastatic carcinoma, showing small tumor
deposits between 0.2--2\,mm without major architectural effacement. \\
\addlinespace

Macro &
A lymph node with macrometastatic carcinoma, featuring tumor deposits
larger than 2\,mm that markedly distort or replace normal nodal architecture. \\

\addlinespace[3pt]
\rowcolor{gray!10}
\multicolumn{2}{l}{\textbf{TCGA-GBMLGG}} \\
\addlinespace[2pt]

LGG &
Glioma without features of glioblastoma. \\
\addlinespace

GBM &
Malignant astrocytoma demonstrating necrosis and/or microvascular
proliferation characteristic of glioblastoma. \\

\end{longtable}

\subsection{Cross-Modal Retrieval Prompts}
\label{app:appendix_crossmodal_prompt}

\begin{longtable}{p{0.15\textwidth} p{0.78\textwidth}}
\caption{Positive prompt bank used for each dataset label. Negatives are drawn from all other classes within the same dataset plus cross-organ distractors from the remaining datasets.}
\label{tab:positive_prompt_bank} \\
\toprule
\textbf{Label} & \textbf{Positive Prompts} \\
\midrule
\endfirsthead

\multicolumn{2}{c}{\small\itshape Table~\ref{tab:positive_prompt_bank} continued} \\
\toprule
\textbf{Label} & \textbf{Positive Prompts} \\
\midrule
\endhead

\midrule
\multicolumn{2}{r}{\small\itshape Continued on next page} \\
\endfoot

\bottomrule
\endlastfoot

\rowcolor{gray!10}
\multicolumn{2}{l}{\textbf{PANDA}} \\
\addlinespace[2pt]

Grade 0 &
\begin{enumerate}[leftmargin=*, noitemsep, topsep=0pt, label=p\arabic*.]
\item Benign prostate tissue with normal gland architecture, uniform small glands, smooth luminal contours, and no nuclear atypia. Stroma is non-reactive and orderly. No evidence of carcinoma
\item benign prostate tissue with no evidence of carcinoma
\item non-neoplastic prostate glands with preserved architecture and no atypia
\item normal prostate glands without dysplasia or malignancy
\end{enumerate} \\
\addlinespace

Grade 1 &
\begin{enumerate}[leftmargin=*, noitemsep, topsep=0pt, label=p\arabic*.]
\item Low-grade prostate cancer with well-formed glands. Tumor glands are still round, separate, and organized. Minimal infiltration into surrounding tissue. Mild nuclear atypia with recognizable glandular structures. No fused or cribriform patterns.
\item ISUP grade 1 prostate adenocarcinoma with well-formed, separate glands
\item Gleason score 3+3=6 prostate adenocarcinoma composed of round, discrete glands
\item low-grade prostate adenocarcinoma with minimal architectural distortion and mild atypia
\end{enumerate} \\
\addlinespace

Grade 2 &
\begin{enumerate}[leftmargin=*, noitemsep, topsep=0pt, label=p\arabic*.]
\item Predominantly well-formed glands (pattern 3) with a lesser component of poorly formed or fused small glands (pattern 4). Early architectural disorder, some poorly formed glands beginning to appear.
\item ISUP grade 2 prostate adenocarcinoma with primary Gleason pattern 3 and secondary pattern 4
\item Gleason score 3+4=7 prostate cancer with a minor component of poorly formed or fused glands
\item predominantly pattern 3 prostate adenocarcinoma with an admixed pattern 4 component
\end{enumerate} \\
\addlinespace

Grade 3 &
\begin{enumerate}[leftmargin=*, noitemsep, topsep=0pt, label=p\arabic*.]
\item Predominantly poorly formed, fused, or cribriform glands (pattern 4) with a lesser component of well-formed glands (pattern 3). Architectural disorganization more prominent. Increased nuclear atypia.
\item ISUP grade 3 prostate adenocarcinoma with primary Gleason pattern 4 and secondary pattern 3
\item Gleason score 4+3=7 prostate cancer dominated by poorly formed or fused glands
\item high-intermediate grade prostate adenocarcinoma with extensive pattern 4 architecture
\end{enumerate} \\
\addlinespace

Grade 4 &
\begin{enumerate}[leftmargin=*, noitemsep, topsep=0pt, label=p\arabic*.]
\item High-grade prostate cancer composed almost entirely of fused glands, poorly formed glands, or cribriform structures. Marked architectural disruption and significant nuclear atypia. No well-formed glands remain.
\item ISUP grade 4 prostate adenocarcinoma composed almost entirely of Gleason pattern 4
\item Gleason score 4+4=8 prostate cancer with fused and cribriform glands and no well-formed glands
\item high-grade prostate adenocarcinoma with diffuse cribriform and fused glandular growth
\end{enumerate} \\
\addlinespace

Grade 5 &
\begin{enumerate}[leftmargin=*, noitemsep, topsep=0pt, label=p\arabic*.]
\item Highest-grade prostate cancer with solid sheets of tumor cells, single-cell infiltration, cords, or comedonecrosis. Complete loss of glandular structure. Highly aggressive morphology with marked atypia
\item ISUP grade 5 prostate adenocarcinoma with solid sheets of tumor cells and single-cell growth
\item Gleason score 4+5=9 or 5+4=9 prostate cancer with extensive pattern 5 and comedonecrosis
\item poorly differentiated prostate carcinoma with pattern 5 morphology and complete loss of gland formation
\end{enumerate} \\
\addlinespace

\rowcolor{gray!10}
\multicolumn{2}{l}{\textbf{UniToPatho}} \\
\addlinespace[2pt]

NORM &
\begin{enumerate}[leftmargin=*, noitemsep, topsep=0pt, label=p\arabic*.]
\item benign colorectal tissue with no dysplasia or malignancy
\item normal colorectal mucosa without dysplasia or malignancy
\item non-neoplastic colonic mucosa with intact crypt architecture
\item benign colonic tissue with preserved glandular architecture
\end{enumerate} \\
\addlinespace

HP &
\begin{enumerate}[leftmargin=*, noitemsep, topsep=0pt, label=p\arabic*.]
\item Hyperplastic polyps are those in which the epithelial cells, although benign, begin to outgrow their available space. The glands have an increased number of goblet cells and therefore look pale or cleared out next to normal epithelium, and there tends to be a thickened or sclerotic-appearing basement membrane under the neoplastic cells. Because the surface area is outgrowing the lamina propria, hyperplastic polyps take on a distinctive frilly or lacy appearance. Crypts cut in cross section have a distinctive star-shaped lumen, also called serrated architecture.
\item hyperplastic polyp of the colon with serrated, frilly crypt architecture
\item benign hyperplastic colonic mucosa with increased goblet cells and serrated crypts
\item non-dysplastic serrated polyp consistent with hyperplastic polyp
\end{enumerate} \\
\addlinespace

TA.LG &
\begin{enumerate}[leftmargin=*, noitemsep, topsep=0pt, label=p\arabic*.]
\item Tubular adenoma shows low-grade dysplasia with parallel crypts and smooth architecture. Crypt-lining epithelial cells appear tall, dark, and pseudostratified with hyperchromatic nuclei and depleted mucin. Mitoses may be present but usually remain at the crypt base.
\item tubular adenoma of the colon with low-grade dysplasia
\item low-grade colonic adenoma composed of closely packed tubular glands
\item adenomatous colonic polyp with predominantly tubular architecture and low-grade dysplasia
\end{enumerate} \\
\addlinespace

TA.HG &
\begin{enumerate}[leftmargin=*, noitemsep, topsep=0pt, label=p\arabic*.]
\item tubular adenoma of the colon with high-grade dysplasia, showing back-to-back or cribriform glands and marked cytologic dysplasia
\item tubular adenoma with high-grade epithelial dysplasia and complex crowded tubular glands
\item advanced tubular adenoma with high-grade dysplasia, nuclear pleomorphism, and loss of polarity
\item colonic tubular adenoma with high-grade dysplasia concerning for progression to invasive adenocarcinoma
\end{enumerate} \\
\addlinespace

TVA.LG &
\begin{enumerate}[leftmargin=*, noitemsep, topsep=0pt, label=p\arabic*.]
\item Tubulovillous adenoma contains both tubular and villous components. Villous areas show fingerlike projections, while tubular areas have smooth and parallel crypts. Low-grade dysplasia results in darkened nuclei, pseudostratification, and mucin depletion, with mitoses usually restricted to the crypt base.
\item tubulovillous adenoma of the colon with mixed tubular and villous architecture and low-grade dysplasia
\item colonic adenoma showing both villous projections and tubular glands, with low-grade cytologic dysplasia
\item partially villous adenomatous polyp (tubulovillous adenoma) with low-grade dysplasia
\end{enumerate} \\
\addlinespace

TVA.HG &
\begin{enumerate}[leftmargin=*, noitemsep, topsep=0pt, label=p\arabic*.]
\item Tubulovillous adenoma with high-grade dysplasia. Glands become back-to-back, fused, or cribriform. Cytology demonstrates pleomorphism, loss of polarity, atypical mitoses, and large nucleoli. Considered pre-invasive but concerning for carcinoma.
\item tubulovillous adenoma with high-grade dysplasia and back-to-back or cribriform glands
\item advanced adenoma with high-grade dysplasia and complex fused glandular architecture
\item colonic adenoma with high-grade epithelial dysplasia concerning for early carcinoma
\end{enumerate} \\
\addlinespace

\rowcolor{gray!10}
\multicolumn{2}{l}{\textbf{CAMELYON16}} \\
\addlinespace[2pt]

Normal &
\begin{enumerate}[leftmargin=*, noitemsep, topsep=0pt, label=p\arabic*.]
\item benign lymph node without metastatic carcinoma; preserved nodal architecture and no tumor cell clusters identified
\item reactive lymphoid tissue with intact architecture and no metastatic tumor cells in the sinuses or parenchyma
\item benign lymphoid tissue showing follicles and sinuses without evidence of metastatic carcinoma
\item lymph node free of metastatic carcinoma, with preserved cortical and medullary structures
\end{enumerate} \\
\addlinespace

Tumor &
\begin{enumerate}[leftmargin=*, noitemsep, topsep=0pt, label=p\arabic*.]
\item lymph node with metastatic carcinoma forming cohesive nests of atypical epithelial cells in the subcapsular sinus
\item lymph node containing metastatic carcinoma deposits partially effacing the normal nodal architecture
\item metastatic breast adenocarcinoma involving lymph node sinuses and parenchyma with infiltrative tumor cell clusters
\item lymph node metastasis with tumor cell clusters disrupting the normal lymphoid architecture
\end{enumerate} \\
\addlinespace

\rowcolor{gray!10}
\multicolumn{2}{l}{\textbf{CAMELYON17}} \\
\addlinespace[2pt]

Negative &
\begin{enumerate}[leftmargin=*, noitemsep, topsep=0pt, label=p\arabic*.]
\item lymph node without metastatic carcinoma; nodal architecture intact and no isolated tumor cells, micrometastases, or macrometastases identified
\item benign lymph node showing reactive follicles and sinuses with no tumor deposits or isolated tumor cells
\item H\&E-stained lymph node free of metastatic carcinoma, with preserved lymphoid architecture
\item negative lymph node without evidence of metastatic deposits or subtle tumor cell clusters
\end{enumerate} \\
\addlinespace

ITC &
\begin{enumerate}[leftmargin=*, noitemsep, topsep=0pt, label=p\arabic*.]
\item lymph node with isolated tumor cells, scattered single cells or tiny clusters not exceeding 0.2 mm, without overt micrometastasis or architectural distortion
\item H\&E-stained lymph node showing very small foci of metastatic carcinoma limited to isolated cells or tiny clusters, consistent with isolated tumor cells
\item isolated tumor cells present within a lymph node, with rare single cells or minute clusters and no significant disruption of nodal architecture
\item lymph node containing only isolated tumor cells without formation of a micrometastatic or macrometastatic deposit
\end{enumerate} \\
\addlinespace

Micro &
\begin{enumerate}[leftmargin=*, noitemsep, topsep=0pt, label=p\arabic*.]
\item lymph node with micrometastatic carcinoma, small tumor deposits between 0.2 mm and 2 mm in the subcapsular sinus and parenchyma
\item H\&E-stained lymph node showing small foci of metastatic carcinoma consistent with micrometastases, not effacing overall nodal architecture
\item micrometastatic carcinoma in a lymph node, with scattered small tumor clusters limited in size and extent
\item lymph node containing small metastatic deposits compatible with micrometastasis, without complete architectural effacement
\end{enumerate} \\
\addlinespace

Macro &
\begin{enumerate}[leftmargin=*, noitemsep, topsep=0pt, label=p\arabic*.]
\item lymph node with macrometastatic carcinoma, large confluent tumor deposits exceeding 2 mm and effacing normal nodal architecture
\item macrometastasis of breast carcinoma within a lymph node, with bulky tumor nodules replacing much of the lymphoid tissue
\item lymph node extensively involved by metastatic carcinoma, with large tumor sheets and marked distortion of nodal structure
\item H\&E-stained lymph node demonstrating macrometastatic carcinoma with broad tumor aggregates replacing normal lymphoid parenchyma
\end{enumerate} \\
\addlinespace

\rowcolor{gray!10}
\multicolumn{2}{l}{\textbf{TCGA-GBMLGG}} \\
\addlinespace[2pt]

LGG &
\begin{enumerate}[leftmargin=*, noitemsep, topsep=0pt, label=p\arabic*.]
\item glioma without features of glioblastoma
\item diffuse lower-grade glioma lacking necrosis or microvascular proliferation
\item infiltrating glioma with low-grade features and absence of glioblastoma characteristics
\item astrocytic or oligodendroglial tumor consistent with WHO grade II or III glioma
\end{enumerate} \\
\addlinespace

GBM &
\begin{enumerate}[leftmargin=*, noitemsep, topsep=0pt, label=p\arabic*.]
\item malignant astrocytoma demonstrating necrosis and/or microvascular proliferation characteristic of glioblastoma
\item glioblastoma with pseudopalisading necrosis and microvascular proliferation
\item high-grade astrocytic tumor consistent with WHO grade IV glioblastoma
\item highly malignant glioma showing necrosis, endothelial proliferation, and marked nuclear atypia
\end{enumerate} \\
\addlinespace

\end{longtable}

\subsection{Paraphrase Robustness Prompts}
\label{app:appendix_paraphrase_prompt}

\textbf{Variant 1: Semantic-Equivalence Paraphrases}

Each label is represented by five semantically equivalent but lexically varied
prompts. Table~\ref{tab:prompts_v1} lists all prompts used.

\begin{longtable}{p{0.15\textwidth} p{0.78\textwidth}}
\caption{Variant 1 prompts: five semantic-equivalence paraphrases per label across all datasets.}
\label{tab:prompts_v1} \\
\toprule
\textbf{Label} & \textbf{Prompts} \\
\midrule
\endfirsthead

\multicolumn{2}{c}{\small\itshape Table~\ref{tab:prompts_v1} continued} \\
\toprule
\textbf{Label} & \textbf{Prompts} \\
\midrule
\endhead

\midrule
\multicolumn{2}{r}{\small\itshape Continued on next page} \\
\endfoot

\bottomrule
\endlastfoot

\rowcolor{gray!10}
\multicolumn{2}{l}{\textbf{PANDA}} \\
\addlinespace[2pt]

Grade 0 &
\begin{enumerate}[leftmargin=*, noitemsep, topsep=0pt, label=v\arabic*.]
\item Normal prostate tissue
\item Non-neoplastic prostatic glands
\item Histologically normal prostate core
\item Prostate tissue without carcinoma
\item Prostatic tissue within normal limits
\end{enumerate} \\

Grade 1 &
\begin{enumerate}[leftmargin=*, noitemsep, topsep=0pt, label=v\arabic*.]
\item Well-formed low-grade prostate adenocarcinoma
\item Low-grade acinar carcinoma of prostate with individual discrete well-formed glands
\item Prostate carcinoma with well-formed glands
\item Well-differentiated prostate cancer
\item ISUP Grade 1 prostatic adenocarcinoma
\end{enumerate} \\

Grade 2 &
\begin{enumerate}[leftmargin=*, noitemsep, topsep=0pt, label=v\arabic*.]
\item Predominantly well-formed prostate adenocarcinoma with focal poorly formed glands
\item Predominantly well-formed prostate adenocarcinoma with minor component of poorly formed glands
\item Acinar prostate cancer with focal emerging fused glands
\item ISUP Grade 2 prostatic adenocarcinoma
\item Prostate tumor with minor poorly formed glands
\end{enumerate} \\

Grade 3 &
\begin{enumerate}[leftmargin=*, noitemsep, topsep=0pt, label=v\arabic*.]
\item Poorly formed gland-predominant prostate adenocarcinoma
\item Prostate carcinoma with predominantly fused and cribriform glands
\item Acinar prostate cancer with predominant poorly formed glands
\item ISUP Grade 3 prostatic adenocarcinoma
\item Prostate tumor with dominant poorly formed gland architecture
\end{enumerate} \\

Grade 4 &
\begin{enumerate}[leftmargin=*, noitemsep, topsep=0pt, label=v\arabic*.]
\item High-grade prostate carcinoma with exclusively poorly formed glands
\item Prostatic adenocarcinoma composed of fused and cribriform glands only
\item Poorly formed malignant prostatic glands without lower-grade component
\item High-grade acinar carcinoma in prostate tissue without well-formed glands
\item High-grade gland-forming prostate malignancy with cribriform architecture only
\end{enumerate} \\

Grade 5 &
\begin{enumerate}[leftmargin=*, noitemsep, topsep=0pt, label=v\arabic*.]
\item Poorly differentiated prostate carcinoma without gland formation
\item High-grade prostatic malignancy with solid growth only
\item Prostate carcinoma with complete loss of glandular architecture
\item Aggressive prostatic tumor with cords and single cells
\item High grade prostate adenocarcinoma with marked necrosis
\end{enumerate} \\

\addlinespace[3pt]
\rowcolor{gray!10}
\multicolumn{2}{l}{\textbf{UniToPatho}} \\
\addlinespace[2pt]

NORM &
\begin{enumerate}[leftmargin=*, noitemsep, topsep=0pt, label=v\arabic*.]
\item Colonic mucosa with no significant histological changes
\item Non-neoplastic colorectal tissue
\item Histologically normal colonic mucosa
\item Colorectal tissue without dysplasia
\item Benign colon epithelium without malignancy
\end{enumerate} \\

HP &
\begin{enumerate}[leftmargin=*, noitemsep, topsep=0pt, label=v\arabic*.]
\item Serrated hyperplastic colorectal polyp, with serrations limited to the upper half of the crypts
\item Benign serrated polyp of colon
\item Colorectal hyperplastic lesion with serrated crypts
\item Goblet cell-rich serrated colonic polyp
\item Hyperplastic polyp with saw-toothed architecture
\end{enumerate} \\

TA.LG &
\begin{enumerate}[leftmargin=*, noitemsep, topsep=0pt, label=v\arabic*.]
\item Low-grade tubular adenoma of colon
\item Tubular colorectal adenoma with low-grade dysplasia
\item Colonic tubular adenoma with low-grade atypia
\item Low-grade dysplastic tubular colonic polyp
\item Tubular adenomatous colon polyp, low grade
\end{enumerate} \\

TA.HG &
\begin{enumerate}[leftmargin=*, noitemsep, topsep=0pt, label=v\arabic*.]
\item High-grade tubular adenoma of colon
\item Tubular colorectal adenoma with severe dysplasia
\item Colonic tubular adenoma with high-grade atypia
\item High-grade dysplastic tubular colonic polyp
\item Tubular adenomatous colon polyp, high grade
\end{enumerate} \\

TVA.LG &
\begin{enumerate}[leftmargin=*, noitemsep, topsep=0pt, label=v\arabic*.]
\item Low-grade tubulovillous adenoma of colon
\item Tubulovillous colorectal adenoma with low-grade dysplasia
\item Colonic tubulovillous adenoma with low-grade atypia
\item Low-grade dysplastic tubulovillous colonic polyp
\item Tubulovillous adenomatous colon polyp, low grade
\end{enumerate} \\

TVA.HG &
\begin{enumerate}[leftmargin=*, noitemsep, topsep=0pt, label=v\arabic*.]
\item High-grade tubulovillous adenoma of colon
\item Tubulovillous colorectal adenoma with severe dysplasia
\item Colonic tubulovillous adenoma with high-grade atypia
\item High-grade dysplastic tubulovillous colonic polyp
\item Tubulovillous adenomatous colon polyp, high grade
\end{enumerate} \\

\addlinespace[3pt]
\rowcolor{gray!10}
\multicolumn{2}{l}{\textbf{CAMELYON16}} \\
\addlinespace[2pt]

Normal &
\begin{enumerate}[leftmargin=*, noitemsep, topsep=0pt, label=v\arabic*.]
\item Benign normal lymph node
\item Histologically normal lymph node
\item Lymph node without metastatic carcinoma
\item Non-neoplastic nodal tissue
\item Uninvolved lymph node
\end{enumerate} \\

Metastasis &
\begin{enumerate}[leftmargin=*, noitemsep, topsep=0pt, label=v\arabic*.]
\item Metastatic carcinoma involving lymph node
\item Sentinel node involved by carcinoma
\item Nodal metastatic carcinoma
\item Lymph node with metastatic carcinoma
\item Metastatic epithelial tumor in sentinel node
\end{enumerate} \\

\addlinespace[3pt]
\rowcolor{gray!10}
\multicolumn{2}{l}{\textbf{CAMELYON17}} \\
\addlinespace[2pt]

Negative &
\begin{enumerate}[leftmargin=*, noitemsep, topsep=0pt, label=v\arabic*.]
\item Negative lymph node
\item Benign lymph node without metastasis
\item Histologically uninvolved nodal tissue
\item Lymph node free of carcinoma
\item No metastatic tumor in lymph node
\end{enumerate} \\

ITC &
\begin{enumerate}[leftmargin=*, noitemsep, topsep=0pt, label=v\arabic*.]
\item Isolated tumor cells in lymph node
\item Rare carcinoma cells within nodal tissue
\item Tiny epithelial cell clusters in sentinel node
\item Minimal nodal tumor burden less than 200 tumor cells
\item Scattered isolated malignant cells in lymph node
\end{enumerate} \\

Micro &
\begin{enumerate}[leftmargin=*, noitemsep, topsep=0pt, label=v\arabic*.]
\item Micrometastatic carcinoma in lymph node
\item Small metastatic deposit in sentinel node
\item Nodal micrometastasis
\item Limited metastatic carcinoma in lymph node
\item Small-volume nodal metastatic tumor
\end{enumerate} \\

Macro &
\begin{enumerate}[leftmargin=*, noitemsep, topsep=0pt, label=v\arabic*.]
\item Macrometastatic carcinoma in lymph node
\item Large metastatic deposit in sentinel node
\item Nodal macrometastasis
\item Extensive metastatic carcinoma in lymph node
\item Large-volume nodal metastatic tumor
\end{enumerate} \\

\addlinespace[3pt]
\rowcolor{gray!10}
\multicolumn{2}{l}{\textbf{TCGA-GBMLGG}} \\
\addlinespace[2pt]

LGG &
\begin{enumerate}[leftmargin=*, noitemsep, topsep=0pt, label=v\arabic*.]
\item Diffuse low-grade glioma
\item Lower-grade infiltrating glioma
\item Glioma lacking glioblastoma features
\item Low-grade glial neoplasm
\item Diffuse glioma without GBM morphology
\end{enumerate} \\

GBM &
\begin{enumerate}[leftmargin=*, noitemsep, topsep=0pt, label=v\arabic*.]
\item Glioblastoma with necrosis and microvascular proliferation
\item High-grade astrocytic tumor consistent with glioblastoma
\item Malignant glioma with glioblastoma morphology
\item Glioblastoma, IDH-wildtype-type morphology
\item Aggressive astrocytic neoplasm with necrosis
\end{enumerate} \\

\end{longtable}

\textbf{Variant 2: Length and Reporting-Style Variations}

Each label is represented by four prompts of increasing length and clinical
detail: \textit{short} (label name only), \textit{medium} (one sentence),
\textit{long} (two to three sentences), and \textit{clinical} (full
pathology description). Table~\ref{tab:prompts_v2} lists all prompts.

\begin{longtable}{p{0.14\textwidth} p{0.10\textwidth} p{0.69\textwidth}}
\caption{Variant 2 prompts: four length and reporting-style variations per label across all datasets.}
\label{tab:prompts_v2} \\
\toprule
\textbf{Label} & \textbf{Length} & \textbf{Prompt} \\
\midrule
\endfirsthead

\multicolumn{3}{c}{\small\itshape Table~\ref{tab:prompts_v2} continued} \\
\toprule
\textbf{Label} & \textbf{Length} & \textbf{Prompt} \\
\midrule
\endhead

\midrule
\multicolumn{3}{r}{\small\itshape Continued on next page} \\
\endfoot

\bottomrule
\endlastfoot

\rowcolor{gray!10}
\multicolumn{3}{l}{\textbf{PANDA}} \\
\addlinespace[2pt]

Grade 0 & short & ISUP Grade 0 (Benign) \\
& medium & Benign prostatic tissue with preserved glandular architecture and bland cytology. \\
& long & Prostate core shows orderly non-neoplastic glands with smooth lumina, uniform epithelium, and no infiltrative or atypical features. \\
& clinical & Benign prostate tissue with normal gland architecture, uniform small glands, smooth luminal contours, and no nuclear atypia. Stroma is non-reactive and orderly. No evidence of carcinoma. \\
\addlinespace

Grade 1 & short & ISUP Grade 1 \\
& medium & Well-differentiated prostatic adenocarcinoma with separate, well-formed infiltrative glands. \\
& long & The biopsy contains discrete malignant acini with preserved glandular structure, mild atypia, and no fused or cribriform architecture. \\
& clinical & Low-grade prostate cancer with well-formed glands. Tumor glands are still round, separate, and organized. Minimal infiltration into surrounding tissue. Mild nuclear atypia with recognizable glandular structures. No fused or cribriform patterns. \\
\addlinespace

Grade 2 & short & ISUP Grade 2 \\
& medium & Predominantly well-formed prostatic carcinoma with focal poorly formed glands. \\
& long & This acinar malignancy is mainly composed of well-formed glands, with a lesser component showing early fusion and poor formation. \\
& clinical & Predominantly well-formed glands with a lesser component of poorly formed or fused small glands. Early architectural disorder, some poorly formed glands beginning to appear. \\
\addlinespace

Grade 3 & short & ISUP Grade 3 \\
& medium & Poorly formed gland-predominant prostatic adenocarcinoma with fused and cribriform glands and a minor well-formed component. \\
& long & The tumor is dominated by poorly formed, fused, and cribriform glands, with only a lesser component of well-formed acini. \\
& clinical & Predominantly poorly formed, fused, or cribriform glands with a lesser component of well-formed glands. Architectural disorganization more prominent. Increased nuclear atypia. \\
\addlinespace

Grade 4 & short & ISUP Grade 4 \\
& medium & High-grade prostatic adenocarcinoma lacking residual well-formed glandular structures. \\
& long & The biopsy shows extensively fused, poorly formed, and cribriform malignant glands with marked architectural disorganization and no lower-grade glandular component. \\
& clinical & High-grade prostate cancer composed almost entirely of fused glands, poorly formed glands, or cribriform structures. Marked architectural disruption and significant nuclear atypia. No well-formed glands remain. \\
\addlinespace

Grade 5 & short & ISUP Grade 5 \\
& medium & Poorly differentiated prostatic carcinoma with complete loss of gland formation. \\
& long & High-grade prostate malignancy shows solid sheets, cords, single-cell infiltration, or comedonecrosis, with absent glandular differentiation throughout. \\
& clinical & Highest-grade prostate cancer with solid sheets of tumor cells, single-cell infiltration, cords, or comedonecrosis. Complete loss of glandular structure. Highly aggressive morphology with marked atypia. \\

\addlinespace[3pt]
\rowcolor{gray!10}
\multicolumn{3}{l}{\textbf{UniToPatho}} \\
\addlinespace[2pt]

NORM & short & Normal \\
& medium & Benign colorectal mucosa without dysplasia, adenomatous change, or malignancy. \\
& long & The tissue shows non-neoplastic colorectal epithelium with preserved architecture and no cytologic atypia or invasive malignant features. \\
& clinical & Benign colorectal tissue with no dysplasia or malignancy. \\
\addlinespace

HP & short & Hyperplastic Polyp \\
& medium & Benign serrated colorectal polyp with goblet cell-rich crypt architecture. \\
& long & The lesion shows pale serrated crypts with stellate lumina, frilly surface contour, and no dysplastic epithelial atypia. \\
& clinical & Hyperplastic polyps in which epithelial cells begin to outgrow available space, with increased goblet cells, frilly surface, and star-shaped crypt lumina (serrated architecture). \\
\addlinespace

TA.LG & short & Tubular Adenoma, Low-Grade Dysplasia \\
& medium & Colorectal adenoma with predominantly tubular architecture and mild dysplasia. \\
& long & This neoplastic polyp shows mostly tubular glands with relatively orderly crypt architecture and low-grade cytologic dysplasia. \\
& clinical & Tubular adenoma of the colon with low-grade dysplasia and predominantly tubular glands with relatively smooth, parallel crypt architecture. \\
\addlinespace

TA.HG & short & Tubular Adenoma, High-Grade Dysplasia \\
& medium & Colorectal adenoma with tubular architecture and severe epithelial dysplasia. \\
& long & The lesion contains crowded back-to-back or cribriform glands accompanied by marked cytologic dysplasia consistent with advanced dysplasia. \\
& clinical & Tubular adenoma of the colon with high-grade dysplasia, showing back-to-back or cribriform glands and marked cytologic dysplasia. \\
\addlinespace

TVA.LG & short & Tubulovillous Adenoma, Low-Grade Dysplasia \\
& medium & Colorectal adenoma with mixed tubular-villous architecture and mild atypia. \\
& long & This adenomatous polyp shows combined villous and tubular growth patterns with low-grade epithelial dysplasia and limited cytologic abnormality. \\
& clinical & Tubulovillous adenoma of the colon with mixed tubular and villous architecture and low-grade cytologic dysplasia. \\
\addlinespace

TVA.HG & short & Tubulovillous Adenoma, High-Grade Dysplasia \\
& medium & Mixed-architecture colorectal adenoma with high-grade epithelial dysplasia. \\
& long & The lesion shows tubular and villous components with complex fused glands and marked epithelial atypia consistent with advanced dysplasia. \\
& clinical & Tubulovillous adenoma with high-grade epithelial dysplasia and complex fused glandular architecture. \\

\addlinespace[3pt]
\rowcolor{gray!10}
\multicolumn{3}{l}{\textbf{CAMELYON16}} \\
\addlinespace[2pt]

Normal & short & Normal \\
& medium & Benign lymph node without metastatic epithelial tumor deposits. \\
& long & The node retains intact lymphoid architecture and shows no metastatic carcinoma. \\
& clinical & A benign lymph node with intact architecture and no metastatic carcinoma. \\
\addlinespace

Metastasis & short & Metastasis \\
& medium & Sentinel lymph node involved by metastatic carcinoma deposits. \\
& long & The node contains malignant epithelial nests or clusters, disrupting and partially replacing normal nodal architecture. \\
& clinical & A lymph node containing metastatic carcinoma that forms disruptive nests or clusters effacing normal nodal architecture. \\

\addlinespace[3pt]
\rowcolor{gray!10}
\multicolumn{3}{l}{\textbf{CAMELYON17}} \\
\addlinespace[2pt]

Negative & short & Negative \\
& medium & Benign lymph node without metastatic carcinoma of any size. \\
& long & The node is free of isolated tumor cells, micrometastatic deposits, and larger metastatic foci, with preserved architecture. \\
& clinical & A benign lymph node with preserved architecture and no metastatic carcinoma, including no isolated tumor cells, micrometastases, or macrometastases. \\
\addlinespace

ITC & short & Isolated Tumor Cells (ITC) \\
& medium & Lymph node containing only scattered single cells or tiny clusters. \\
& long & The nodal tissue harbors only minute epithelial deposits below 0.2\,mm, without meaningful architectural distortion or effacement. \\
& clinical & A lymph node containing only isolated tumor cells---single cells or tiny clusters under 0.2\,mm---without architectural distortion. \\
\addlinespace

Micro & short & Micrometastasis \\
& medium & Small nodal metastatic carcinoma deposit measuring 0.2 to 2\,mm. \\
& long & The lymph node contains a limited metastatic epithelial focus of intermediate size, with only minor disturbance of architecture. \\
& clinical & A lymph node with micrometastatic carcinoma, showing small tumor deposits between 0.2--2\,mm without major architectural effacement. \\
\addlinespace

Macro & short & Macrometastasis \\
& medium & Large nodal metastatic carcinoma deposit with architectural effacement. \\
& long & The node shows a metastatic epithelial tumor focus exceeding 2\,mm, markedly distorting or replacing normal lymphoid tissue. \\
& clinical & A lymph node with macrometastatic carcinoma, featuring tumor deposits larger than 2\,mm that markedly distort or replace normal nodal architecture. \\

\addlinespace[3pt]
\rowcolor{gray!10}
\multicolumn{3}{l}{\textbf{TCGA-GBMLGG}} \\
\addlinespace[2pt]

LGG & short & Low-Grade Glioma (LGG) \\
& medium & Diffuse glioma lacking necrosis and microvascular proliferation. \\
& long & The infiltrating glial neoplasm does not exhibit the necrotic change or vascular proliferation characteristic of glioblastoma. \\
& clinical & Glioma without features of glioblastoma. \\
\addlinespace

GBM & short & Glioblastoma (GBM) \\
& medium & High-grade astrocytic tumor with necrosis and microvascular proliferation. \\
& long & This malignant glial neoplasm shows necrosis and/or prominent microvascular proliferation, supporting a diagnosis of glioblastoma. \\
& clinical & Malignant astrocytoma demonstrating necrosis and/or microvascular proliferation characteristic of glioblastoma. \\

\end{longtable}

\subsection{Open Vocabulary Retrieval Prompts}
\label{app:appendix_openvocab_prompt}


We evaluate open vocabulary retrieval under four candidate pool settings that impose distinct forms of semantic competition. For each setting, a slide image is ranked against a label-specific candidate pool; the ground truth text must be retrieved from among the distractors. Tables~\ref{tab:ov_set1}-\ref{tab:ov_set4} list the ground truth and candidate texts for every dataset and label.

\subsubsection*{Set 1: Open Vocabulary}

Candidates are generic, organ-agnostic pathology terms (e.g., \textit{benign
tissue}, \textit{high-grade malignancy}). The pool size is 20 per label.

\begin{longtable}{p{0.15\textwidth} p{0.78\textwidth}}
\caption{Set 1 (Open Vocabulary): shared 20-candidate pool and dataset-specific ground-truth mapping.}
\label{tab:ov_set1} \\
\toprule
\multicolumn{2}{l}{\textbf{Part I: Shared Candidate Pool}} \\
\midrule
\endfirsthead

\multicolumn{2}{c}{\small\itshape Table~\ref{tab:ov_set1} continued} \\
\toprule
\textbf{Label} & \textbf{Ground Truth} \\
\midrule
\endhead

\midrule
\multicolumn{2}{r}{\small\itshape Continued on next page} \\
\endfoot

\bottomrule
\endlastfoot

\addlinespace[2pt]

\multicolumn{2}{p{0.95\textwidth}}{
benign tissue; low-grade malignancy; intermediate-grade malignancy;
high-grade malignancy; poorly differentiated carcinoma; well-differentiated
carcinoma; moderately differentiated carcinoma; adenocarcinoma; metastatic
carcinoma; in situ carcinoma; dysplasia; low-grade dysplasia; high-grade
dysplasia; invasive carcinoma; neuroendocrine carcinoma; squamous cell
carcinoma; small cell carcinoma; undifferentiated carcinoma; tubular adenoma;
sessile serrated adenoma.
} \\

\addlinespace[8pt]
\toprule
\multicolumn{2}{l}{\textbf{Part II: Dataset-specific Ground Truth Mapping}} \\
\midrule
\textbf{Dataset Label} & \textbf{Ground Truth Candidate} \\
\midrule

\rowcolor{gray!10}
\multicolumn{2}{l}{\textbf{PANDA}} \\
\addlinespace[2pt]

Grade 0 & benign tissue \\
Grade 1 & low-grade malignancy \\
Grade 2 & intermediate-grade malignancy \\
Grade 3 & intermediate-grade malignancy \\
Grade 4 & high-grade malignancy \\
Grade 5 & poorly differentiated carcinoma \\

\addlinespace[3pt]
\rowcolor{gray!10}
\multicolumn{2}{l}{\textbf{UniToPatho}} \\
\addlinespace[2pt]

NORM & benign tissue \\
HP & benign tissue \\
TA.LG & low-grade dysplasia \\
TA.HG & high-grade dysplasia \\
TVA.LG & low-grade dysplasia \\
TVA.HG & high-grade dysplasia \\

\addlinespace[3pt]
\rowcolor{gray!10}
\multicolumn{2}{l}{\textbf{CAMELYON16}} \\
\addlinespace[2pt]

Normal & benign tissue \\
Tumor & metastatic carcinoma \\

\addlinespace[3pt]
\rowcolor{gray!10}
\multicolumn{2}{l}{\textbf{CAMELYON17}} \\
\addlinespace[2pt]

Negative & benign tissue \\
ITC & metastatic carcinoma \\
Micro & metastatic carcinoma \\
Macro & metastatic carcinoma \\

\addlinespace[3pt]
\rowcolor{gray!10}
\multicolumn{2}{l}{\textbf{TCGA-GBMLGG}} \\
\addlinespace[2pt]

LGG & low-grade malignancy \\
GBM & high-grade malignancy \\

\end{longtable}

\subsubsection*{Set 2: Label-Specific Mixed Diagnostic Candidates}

Pools are specified separately for each target label and combine same-organ diagnostic alternatives with cross-organ and broader pathological distractors. Composition varies by target label, and same-organ candidates are not necessarily the majority. The pool size is up to 10 candidates per label.

\begin{longtable}{p{0.13\textwidth} p{0.20\textwidth} p{0.60\textwidth}}
\caption{Set 2 (Label-Specific Mixed Diagnostic Candidates): label-specific mixtures of same-organ diagnostic alternatives, cross-organ candidates, and broader pathological distractors. Each pool contains up to 10 entries and its composition varies by target label. A small number of pools contain a repeated string, so total and unique counts differ.}
\label{tab:ov_set2} \\
\toprule
\textbf{Label} & \textbf{Ground Truth} & \textbf{Candidate Pool} \\
\midrule
\endfirsthead

\multicolumn{3}{c}{\small\itshape Table~\ref{tab:ov_set2} continued} \\
\toprule
\textbf{Label} & \textbf{Ground Truth} & \textbf{Candidate Pool} \\
\midrule
\endhead

\midrule
\multicolumn{3}{r}{\small\itshape Continued on next page} \\
\endfoot

\bottomrule
\endlastfoot

\rowcolor{gray!10}
\multicolumn{3}{l}{\textbf{PANDA}} \\
\addlinespace[2pt]

Grade 0 & benign prostate tissue &
benign prostate tissue; ISUP Grade 1 prostate cancer; ISUP Grade 2 prostate cancer; normal colon tissue; benign colorectal mucosa; normal lymph node; benign breast tissue; normal liver; normal lung tissue; benign thyroid tissue \\
\addlinespace

Grade 1 & ISUP Grade 1 prostate cancer &
ISUP Grade 1 prostate cancer; benign prostate tissue; ISUP Grade 2 prostate cancer; ISUP Grade 3 prostate cancer; low-grade tubular adenoma; low-grade glioma; tubular adenoma low-grade; lymph node micrometastasis; low-grade dysplasia; high grade prostatic intraepithelial neoplasia (HGPIN) \\
\addlinespace

Grade 2 & ISUP Grade 2 prostate cancer &
ISUP Grade 2 prostate cancer; ISUP Grade 1 prostate cancer; ISUP Grade 3 prostate cancer; ISUP Grade 4 prostate cancer; tubular adenoma high-grade; tubular adenoma low-grade; moderately differentiated carcinoma; intermediate grade dysplasia; lymph node micrometastasis; lymph node macrometastasis \\
\addlinespace

Grade 3 & ISUP Grade 3 prostate cancer &
ISUP Grade 3 prostate cancer; ISUP Grade 2 prostate cancer; ISUP Grade 4 prostate cancer; ISUP Grade 5 prostate cancer; tubular adenoma high-grade; tubulovillous adenoma high-grade; high-grade dysplasia; HGPIN; poorly differentiated carcinoma; lymph node macrometastasis \\
\addlinespace

Grade 4 & ISUP Grade 4 prostate cancer &
ISUP Grade 4 prostate cancer; ISUP Grade 3 prostate cancer; ISUP Grade 5 prostate cancer; ISUP Grade 2 prostate cancer; tubulovillous adenoma high-grade; glioblastoma; high-grade dysplasia; well differentiated carcinoma; lymph node macrometastasis; HGPIN \\
\addlinespace

Grade 5 & ISUP Grade 5 prostate cancer &
ISUP Grade 5 prostate cancer; ISUP Grade 4 prostate cancer; ISUP Grade 3 prostate cancer; glioblastoma; well differentiated carcinoma; high-grade serous ovarian carcinoma; invasive carcinoma; tubulovillous adenoma high-grade; lymph node macrometastasis; small cell carcinoma \\

\addlinespace[4pt]
\rowcolor{gray!10}
\multicolumn{3}{l}{\textbf{UniToPatho}} \\
\addlinespace[2pt]

NORM & normal colon tissue &
normal colon tissue; hyperplastic polyp; benign prostate tissue; normal lymph node; normal liver; normal lung tissue; benign breast tissue; benign thyroid tissue; normal skin; benign gastric mucosa \\
\addlinespace

HP & hyperplastic polyp &
hyperplastic polyp; normal colon tissue; tubular adenoma low-grade; tubulovillous adenoma low-grade; serrated adenoma; benign colorectal mucosa; tubular adenoma high-grade; inflammatory polyp; juvenile polyp; lymphoid polyp \\
\addlinespace

TA.LG & tubular adenoma low-grade &
tubular adenoma low-grade; tubular adenoma high-grade; tubulovillous adenoma low-grade; hyperplastic polyp; normal colon tissue; ISUP Grade 1 prostate cancer; low-grade glioma; serrated adenoma; tubulovillous adenoma high-grade; low-grade dysplasia \\
\addlinespace

TA.HG & tubular adenoma high-grade &
tubular adenoma high-grade; tubular adenoma low-grade; tubulovillous adenoma high-grade; tubulovillous adenoma low-grade; ISUP Grade 3 prostate cancer; high-grade dysplasia; invasive colorectal carcinoma; glioblastoma; poorly differentiated carcinoma; ISUP Grade 4 prostate cancer \\
\addlinespace

TVA.LG & tubulovillous adenoma low-grade &
tubulovillous adenoma low-grade; tubulovillous adenoma high-grade; tubular adenoma low-grade; tubular adenoma high-grade; hyperplastic polyp; normal colon tissue; sessile serrated adenoma; inflammatory polyp; ISUP Grade 1 prostate cancer; inflammatory polyp \\
\addlinespace

TVA.HG & tubulovillous adenoma high-grade &
tubulovillous adenoma high-grade; tubulovillous adenoma low-grade; tubular adenoma high-grade; tubular adenoma low-grade; invasive colorectal carcinoma; high-grade dysplasia; ISUP Grade 4 prostate cancer; glioblastoma; poorly differentiated carcinoma; ISUP Grade 3 prostate cancer \\

\addlinespace[4pt]
\rowcolor{gray!10}
\multicolumn{3}{l}{\textbf{CAMELYON16}} \\
\addlinespace[2pt]

Normal & normal lymph node &
normal lymph node; lymph node metastasis; lymph node micrometastasis; lymph node isolated tumor cells; benign prostate tissue; normal colon tissue; normal liver; reactive lymph node; normal spleen; normal thymus \\
\addlinespace

Tumor & lymph node metastasis &
lymph node metastasis; normal lymph node; lymph node micrometastasis; lymph node macrometastasis; lymph node isolated tumor cells; ISUP Grade 4 prostate cancer; invasive colorectal carcinoma; glioblastoma; invasive ductal carcinoma; poorly differentiated carcinoma \\

\addlinespace[4pt]
\rowcolor{gray!10}
\multicolumn{3}{l}{\textbf{CAMELYON17}} \\
\addlinespace[2pt]

Negative & normal lymph node &
normal lymph node; lymph node isolated tumor cells; lymph node micrometastasis; lymph node macrometastasis; reactive lymph node; normal spleen; normal thymus; benign prostate tissue; normal colon tissue; normal liver \\
\addlinespace

ITC & lymph node isolated tumor cells &
lymph node isolated tumor cells; normal lymph node; lymph node micrometastasis; lymph node macrometastasis; lymph node metastasis; reactive lymph node; ISUP Grade 1 prostate cancer; tubular adenoma low-grade; low-grade glioma; normal spleen \\
\addlinespace

Micro & lymph node micrometastasis &
lymph node micrometastasis; lymph node isolated tumor cells; lymph node macrometastasis; normal lymph node; lymph node metastasis; ISUP Grade 2 prostate cancer; ISUP Grade 3 prostate cancer; tubular adenoma high-grade; reactive lymph node; low-grade glioma \\
\addlinespace

Macro & lymph node macrometastasis &
lymph node macrometastasis; lymph node micrometastasis; lymph node with necrotizing lymphadenitis; lymph node isolated tumor cells; normal lymph node; ISUP Grade 4 prostate cancer; ISUP Grade 5 prostate cancer; glioblastoma; invasive colorectal carcinoma; poorly differentiated carcinoma \\

\addlinespace[4pt]
\rowcolor{gray!10}
\multicolumn{3}{l}{\textbf{TCGA-GBMLGG}} \\
\addlinespace[2pt]

LGG & low-grade glioma &
low-grade glioma; glioblastoma; ISUP Grade 1 prostate cancer; ISUP Grade 2 prostate cancer; tubular adenoma low-grade; low-grade dysplasia; lymph node micrometastasis; reactive gliosis; normal brain tissue; reactive gliosis \\
\addlinespace

GBM & glioblastoma &
glioblastoma; low-grade glioma; ISUP Grade 4 prostate cancer; ISUP Grade 5 prostate cancer; tubulovillous adenoma high-grade; lymph node macrometastasis; reactive gliosis; high-grade serous ovarian carcinoma; small cell carcinoma; invasive ductal carcinoma \\

\end{longtable}

\subsubsection*{Set 3: Cross-Organ}

Candidates include terms from different organ systems, testing whether models can distinguish the correct organ and diagnosis. Each pool contains 10 entries, comprising a dataset-level shared cross-organ pool plus the label-specific ground-truth candidate. Where the shared pool already contains the ground-truth string, the pool contains 9 unique candidates.

\begin{longtable}{p{0.13\textwidth} p{0.22\textwidth} p{0.58\textwidth}}
\caption{Set 3 (Cross-Organ): dataset-level shared cross-organ candidate pools with label-specific candidates.}
\label{tab:ov_set3} \\
\toprule
\textbf{Label} & \textbf{Ground Truth} & \textbf{Label-specific Candidate(s)} \\
\midrule
\endfirsthead

\multicolumn{3}{c}{\small\itshape Table~\ref{tab:ov_set3} continued} \\
\toprule
\textbf{Label} & \textbf{Ground Truth} & \textbf{Label-specific Candidate(s)} \\
\midrule
\endhead

\midrule
\multicolumn{3}{r}{\small\itshape Continued on next page} \\
\endfoot

\bottomrule
\endlastfoot

\rowcolor{gray!10}
\multicolumn{3}{l}{\textbf{PANDA}} \\
\addlinespace[2pt]

\multicolumn{3}{p{0.95\textwidth}}{
\textbf{Shared candidate pool:}
glioblastoma; low-grade glioma; lymph node metastasis; normal lymph node;
colon adenocarcinoma; normal colon tissue; breast carcinoma; normal lung tissue;
benign prostate tissue
} \\
\addlinespace[2pt]

Grade 0 & benign prostate tissue & benign prostate tissue \\
Grade 1 & ISUP Grade 1 prostate cancer & ISUP Grade 1 \\
Grade 2 & ISUP Grade 2 prostate cancer & ISUP Grade 2 \\
Grade 3 & ISUP Grade 3 prostate cancer & ISUP Grade 3 \\
Grade 4 & ISUP Grade 4 prostate cancer & ISUP Grade 4 \\
Grade 5 & ISUP Grade 5 prostate cancer & ISUP Grade 5 \\

\addlinespace[4pt]

\rowcolor{gray!10}
\multicolumn{3}{l}{\textbf{UniToPatho}} \\
\addlinespace[2pt]

\multicolumn{3}{p{0.95\textwidth}}{
\textbf{Shared candidate pool:}
benign prostate tissue; ISUP Grade 4 prostate cancer; glioblastoma;
low-grade glioma; lymph node metastasis; normal lymph node; breast carcinoma;
normal lung tissue; hepatocellular carcinoma
} \\
\addlinespace[2pt]

NORM & normal colon tissue & normal colon tissue \\
HP & hyperplastic polyp & hyperplastic polyp \\
TA.LG & tubular adenoma low-grade & tubular adenoma low-grade \\
TA.HG & tubular adenoma high-grade & tubular adenoma high-grade \\
TVA.LG & tubulovillous adenoma low-grade & tubulovillous adenoma low-grade \\
TVA.HG & tubulovillous adenoma high-grade & tubulovillous adenoma high-grade \\

\addlinespace[4pt]

\rowcolor{gray!10}
\multicolumn{3}{l}{\textbf{CAMELYON16}} \\
\addlinespace[2pt]

\multicolumn{3}{p{0.95\textwidth}}{
\textbf{Shared candidate pool:}
benign prostate tissue; ISUP Grade 4 prostate cancer; glioblastoma;
low-grade glioma; colon adenocarcinoma; normal colon tissue; breast carcinoma;
normal lung tissue; normal lymph node; lymph node with metastatic carcinoma
} \\
\addlinespace[2pt]

Normal & normal lymph node & -- \\
Tumor & lymph node with metastatic carcinoma & -- \\

\addlinespace[4pt]

\rowcolor{gray!10}
\multicolumn{3}{l}{\textbf{CAMELYON17}} \\
\addlinespace[2pt]

\multicolumn{3}{p{0.95\textwidth}}{
\textbf{Shared candidate pool:}
benign prostate tissue; ISUP Grade 4 prostate cancer; glioblastoma;
low-grade glioma; colon adenocarcinoma; normal colon tissue; breast carcinoma;
normal lung tissue; hepatocellular carcinoma
} \\
\addlinespace[2pt]

Negative & normal lymph node & normal lymph node \\
ITC & lymph node with isolated tumor cells & lymph node with isolated tumor cells \\
Micro & lymph node with micrometastasis & lymph node with micrometastasis \\
Macro & lymph node with macrometastasis & lymph node with macrometastasis \\

\addlinespace[4pt]

\rowcolor{gray!10}
\multicolumn{3}{l}{\textbf{TCGA-GBMLGG}} \\
\addlinespace[2pt]

\multicolumn{3}{p{0.95\textwidth}}{
\textbf{Shared candidate pool:}
benign prostate tissue; ISUP Grade 4 prostate cancer; lymph node metastasis;
normal lymph node; colon adenocarcinoma; normal colon tissue; breast carcinoma;
normal lung tissue; hepatocellular carcinoma} \\
\addlinespace[2pt]

LGG & low-grade glioma & -- \\
GBM & glioblastoma & -- \\

\end{longtable}

\subsubsection*{Set 4: All Dataset Labels}

The candidate pool contains all class labels from all five datasets. Each pool contains 20 total entries, of which 19 are unique: CAMELYON16 and CAMELYON17 share the label \emph{normal lymph node}.

\begin{longtable}{p{0.20\textwidth} p{0.73\textwidth}}
\caption{Set 4 (All Dataset Labels): shared pool of all class labels across the benchmark: 20 entries, of which 19 are unique (CAMELYON16 and CAMELYON17 share the label \emph{normal lymph node}).}
\label{tab:ov_set4} \\
\midrule
\endfirsthead

\multicolumn{2}{c}{\small\itshape Table~\ref{tab:ov_set4} continued} \\
\toprule
\textbf{Label} & \textbf{Ground Truth} \\
\midrule
\endhead

\midrule
\multicolumn{2}{r}{\small\itshape Continued on next page} \\
\endfoot

\bottomrule
\endlastfoot

\multicolumn{2}{l}{\textbf{Part I: Shared Candidate Pool}} \\
\toprule
\addlinespace[2pt]

\multicolumn{2}{p{0.95\textwidth}}{
benign prostate tissue; ISUP Grade 1 prostate cancer; ISUP Grade 2 prostate cancer;
ISUP Grade 3 prostate cancer; ISUP Grade 4 prostate cancer; ISUP Grade 5 prostate cancer;
normal colon tissue; hyperplastic polyp; tubular adenoma low-grade; tubular adenoma high-grade;
tubulovillous adenoma low-grade; tubulovillous adenoma high-grade; normal lymph node;
lymph node metastasis; lymph node isolated tumor cells; lymph node micrometastasis;
lymph node macrometastasis; low-grade glioma; glioblastoma.
} \\

\addlinespace[8pt]
\toprule
\multicolumn{2}{l}{\textbf{Part II: Dataset-specific Ground Truth Mapping}} \\
\midrule
\textbf{Label} & \textbf{Ground Truth} \\
\midrule

\rowcolor{gray!10}
\multicolumn{2}{l}{\textbf{PANDA}} \\
\addlinespace[2pt]

Grade 0 & benign prostate tissue \\
Grade 1 & ISUP Grade 1 prostate cancer \\
Grade 2 & ISUP Grade 2 prostate cancer \\
Grade 3 & ISUP Grade 3 prostate cancer \\
Grade 4 & ISUP Grade 4 prostate cancer \\
Grade 5 & ISUP Grade 5 prostate cancer \\

\addlinespace[3pt]
\rowcolor{gray!10}
\multicolumn{2}{l}{\textbf{UniToPatho}} \\
\addlinespace[2pt]

NORM    & normal colon tissue \\
HP      & hyperplastic polyp \\
TA.LG   & tubular adenoma low-grade \\
TA.HG   & tubular adenoma high-grade \\
TVA.LG  & tubulovillous adenoma low-grade \\
TVA.HG  & tubulovillous adenoma high-grade \\

\addlinespace[3pt]
\rowcolor{gray!10}
\multicolumn{2}{l}{\textbf{CAMELYON16}} \\
\addlinespace[2pt]

Normal & normal lymph node \\
Tumor  & lymph node metastasis \\

\addlinespace[3pt]
\rowcolor{gray!10}
\multicolumn{2}{l}{\textbf{CAMELYON17}} \\
\addlinespace[2pt]

Negative & normal lymph node \\
ITC      & lymph node isolated tumor cells \\
Micro    & lymph node micrometastasis \\
Macro    & lymph node macrometastasis \\

\addlinespace[3pt]
\rowcolor{gray!10}
\multicolumn{2}{l}{\textbf{TCGA-GBMLGG}} \\
\addlinespace[2pt]

LGG & low-grade glioma \\
GBM & glioblastoma \\

\end{longtable}

\newpage

\end{document}